\documentclass[10pt,twocolumn,letterpaper]{article}

\usepackage[pagenumbers]{wacv}        %

\usepackage{multirow}
\usepackage{colortbl}   %
\usepackage{placeins}   %
\usepackage{float}      %
\usepackage{eccvabbrv}  %
\usepackage[accsupp]{axessibility}  %

\renewcommand{\dbltopfraction}{0.95}
\renewcommand{\dblfloatpagefraction}{0.9}
\definecolor{wacvblue}{rgb}{0.21,0.49,0.74}
\definecolor{bestgreen}{rgb}{0.80,0.93,0.80}  %
\definecolor{hltint}{rgb}{0.86,0.92,1.00}     %
\usepackage[pagebackref,breaklinks,colorlinks,allcolors=wacvblue]{hyperref}

\newcommand{\suptab}[1]{Table~\ref{S-tab:#1}}
\newcommand{\supfig}[1]{Fig.~\ref{S-fig:#1}}
\newcommand{\supsec}[1]{Sec.~\ref{S-sec:#1}}

\usepackage{pdflscape}
\makeatletter
\newcommand{\cleanclearpage}{\clearpage\global\@colroom\@colht\global\vsize\@colroom}
\makeatother

\def\wacvPaperID{1944}
\def\confName{WACV}
\def\confYear{2027}

\title{BhopalMLS: A LiDAR Dataset for Semantic Scene Understanding in
Heterogeneous Urban Environments}

\author{Pyare Lal Chauhan$^{1}$\qquad Vaibhav Kumar$^{1}$\thanks{Corresponding author.}\qquad Bharat Lohani$^{2}$\\
$^{1}$GeoAI4Cities Lab, Department of Data Science and Engineering, IISER Bhopal, India\\
$^{2}$Department of Civil Engineering, IIT Kanpur, India\\
{\tt\small \{pyare22, vaibhav\}@iiserb.ac.in, blohani@iitk.ac.in}
}

\AddToShipoutPicture*{\AtPageUpperLeft{\put(\LenToUnit{\dimexpr(\paperwidth-\textwidth)/2\relax},\LenToUnit{-46pt}){\parbox[t]{\textwidth}{\centering\fontsize{7}{8.4}\selectfont
\textcopyright~2027 IEEE. Personal use of this material is permitted. Permission from IEEE must be obtained for all other uses, in any current or future media, including reprinting/republishing this material for advertising or promotional purposes, creating new collective works, for resale or redistribution to servers or lists, or reuse of any copyrighted component of this work in other works. This is the accepted version of a paper to appear in the 2027 IEEE/CVF Winter Conference on Applications of Computer Vision (WACV).}}}}
\begin{document}
\maketitle

\begin{abstract}
We introduce BhopalMLS, a large-scale Mobile Laser Scanning (MLS) dataset from a dense, heterogeneous urban environment in Bhopal, India, with mixed traffic, informal structures, and narrow unplanned streets. It spans approximately 10\,km of road corridors with over 1.2 billion points and 19{,}518 annotated instances across 62 fine-grained static categories, grouped into 12-class and 21-class benchmark taxonomies; 6 optional moving-class subclasses support dynamic-object analysis. Existing 3D datasets, collected in well-planned cities, are structurally more uniform than such unplanned settings, and lack their subtype diversity. On semantic and instance segmentation at both granularities, fine-tuned foundation models surpass supervised baselines: Utonia(L) reaches 71.28\% mIoU (vs.\ PTv3's 68.04\% from scratch) and 44.07\% mAP50 on the 12-class benchmark. Cross-region experiments against four established datasets (Ke-MLS, Toronto-3D, Paris-Lille-3D, SemanticKITTI) reveal a consistent \emph{asymmetric transferability}: models trained on BhopalMLS transfer better than the reverse on every pair, for both Utonia(L) and Concerto(L) (mean 21.5, up to 36 mIoU points), holding even when the target is denser. Factor-controlled retraining at a fixed label budget, repeated on both models, attributes this to the \emph{coverage} of BhopalMLS's subtype inventory rather than to dataset size or corridor geometry: collapsing each class to a single subtype costs 13.0 and 13.3 mIoU points for Utonia(L) and Concerto(L). BhopalMLS therefore offers a testbed for 3D perception beyond well-planned cities. Data and an interactive viewer are at \url{https://geoai4cities.github.io/bhopalmls}.
\end{abstract}

\section{Introduction}
\label{sec:intro}

Accurate semantic segmentation of 3D point clouds is fundamental to autonomous navigation \cite{gu2024clft}, urban modeling \cite{batty2021new}, and digital twin development for smart cities \cite{mazzetto2024review}. Real-world urban scenes comprise geometrically and semantically diverse elements, including buildings, dense vegetation, vehicles, road furniture, and overhead wiring. Their reliable recognition demands high-density, semantically labeled LiDAR data capturing this diversity. However, deep learning performance is constrained by the limited geographic and structural variety of available training benchmarks.

Significant progress in 3D perception has been driven by benchmark datasets such as SemanticKITTI \cite{behley2019semantickitti}, Semantic3D \cite{hackel2017semantic3d}, nuScenes \cite{caesar2020nuscenes}, Toronto-3D \cite{tan2020toronto}, and KITTI-360 \cite{liao2022kitti}. While invaluable, these datasets are collected in well-planned, regulation-compliant urban environments with orderly road layouts and clearly separated object classes. Cities in developing regions such as India present a fundamentally different setting: unstructured road networks, narrow congested corridors, extreme geometric clutter, and diverse object categories such as auto-rickshaws, hand-carts, stray animals, and informal roadside structures that are absent from existing benchmarks.

This data gap has direct consequences: models trained on structured benchmarks suffer severe performance degradation when tested on such dense, unstructured scenes. Across four established benchmarks (Ke-MLS, Toronto-3D, Paris-Lille-3D, and SemanticKITTI), our cross-region experiments (Table~\ref{tab:cross_region_1cm}) reveal a consistent \emph{asymmetric transferability}, with a benchmark-pretrained model collapsing to just 3.89\% mIoU on Indian urban data in the most severe case (\suptab{cross_city_full}). Despite this critical need, no large-scale MLS dataset currently offers dense point clouds with fine-grained semantic and instance-level annotations from such environments.

To bridge this gap, we introduce \textbf{BhopalMLS}, a large-scale, high-density MLS point cloud dataset collected from the city of Bhopal, India. It spans approximately 10\,km of urban road corridors with over 1.2 billion points across 62 fine-grained static classes (19{,}518 instances; 68 classes with 6 optional moving subclasses). Under a fair voxel-normalized protocol (5\,m XY tiles, 25\,cm voxel), BhopalMLS exhibits substantially higher density heterogeneity (Gini=0.53, CoV=1.15) and vertical extent (10.1\,m) than Western MLS datasets (Toronto-3D~\cite{tan2020toronto}, Paris-Lille-3D~\cite{roynard2018paris}) and SemanticKITTI~\cite{behley2019semantickitti}; the same holds for another Indian MLS dataset, Ke-MLS~\cite{kumar2026ke} (vehicle Riegl VQ-450), indicating these properties reflect the urban environment rather than the acquisition sensor (Table~\ref{tab:dataset_complexity}). This makes BhopalMLS a challenging benchmark for 3D perception in complex, unstructured urban environments.

The principal contributions of this work are as follows:
\begin{enumerate}
\item \textbf{BhopalMLS}, to our knowledge the first large-scale, densely annotated MLS dataset of a dense, unplanned commercial urban environment, with fine-grained semantic and instance-level labels.
\item Comprehensive semantic segmentation benchmarks using state-of-the-art supervised models (PTv3, LitePT-L, KpConvX-L) and pretrained foundation models (Sonata, Concerto, Utonia) under multiple adaptation strategies (linear probe, decoder probe, LoRA, fine-tuning). Results demonstrate that foundation models can surpass fully supervised baselines on complex urban data.
\item A cross-region study identifying a consistent \emph{asymmetric transferability}: models trained on BhopalMLS generalize to four structured benchmarks (Ke-MLS, Toronto-3D, Paris-Lille-3D, SemanticKITTI) better than the reverse, on every pair and across two architectures (by up to 36 mIoU points). The pattern holds even where the target is denser and across sensors and regions. Factor-controlled retraining localises the cause to subtype \emph{coverage}, not size or geometry; a domain-adaptation analysis confirms the gap is structural, not a second-order covariance shift.
\end{enumerate}

\begin{table}[!htbp]
\centering
\setlength{\tabcolsep}{2.5pt}
\renewcommand{\arraystretch}{1.1}
\newcommand{\tyes}{$\checkmark$}\newcommand{\tno}{\textcolor[gray]{0.6}{$\times$}}%
\caption{Comparison of major LiDAR point cloud datasets with \textbf{BhopalMLS}. Point totals reflect coverage \emph{and} density: ours arises from dense handheld capture (4.9\,cm spacing) of a compact ${\sim}$10\,km corridor, not large extent. Classes: used in the benchmark (total annotated), where they differ.}
\label{tab:dataset_comparison}
\resizebox{\columnwidth}{!}{
\begin{tabular}{lcrcccccc}
\toprule
\textbf{Name} & \textbf{Year} & \textbf{Points} & \textbf{Classes} & \textbf{RGB} & \textbf{Sensor} & \begin{tabular}[c]{@{}c@{}}\textbf{Inten-}\\\textbf{sity}\end{tabular} & \begin{tabular}[c]{@{}c@{}}\textbf{Inst-}\\\textbf{ance}\end{tabular} & \textbf{Country} \\
\midrule
Oakland \cite{munoz2009oakland} & 2009 & 1.6M & 5 (44) & \tno & MLS & \tno & \tno & USA \\
Paris-rue-Madame \cite{serna2014paris} & 2014 & 20M & 17 & \tno & MLS & \tyes &  \tyes & France \\
TUM-MLS \cite{zhu2020tum} & 2016 & 40M & 8 & \tno & MLS & \tyes &  \tyes & Germany \\
Semantic3D \cite{hackel2017semantic3d} & 2017 & 4{,}000M & 8 & \tyes & TLS & \tyes &  \tno & Europe \\
Paris-Lille-3D \cite{roynard2018paris} & 2018 & 143M & 9 (50) & \tno & MLS & \tyes &  \tyes & France \\
SemanticKITTI \cite{behley2019semantickitti} & 2019 & 4{,}549M & 19 (28) & \tno & MLS & \tyes &  \tyes & Germany \\
Campus3D \cite{li2020campus3d} & 2020 & 937M & 24 & \tyes & Photogr. & \tno &  \tyes & Singapore \\
DALES \cite{varney2020dales} & 2020 & 505M & 8 & \tno & ALS & \tno & \tno & Canada \\
LASDU \cite{ye2020lasdu} & 2020 & 3.12M & 5 & \tno & ALS & \tyes &  \tno & China \\
Toronto-3D \cite{tan2020toronto} & 2020 & 78.3M & 8 & \tyes & MLS & \tyes &  \tno & Canada \\
KITTI-360 \cite{liao2022kitti} & 2022 & 1{,}000M & 19 (37) & \tyes & MLS & \tno &  \tyes & Germany \\
WHU-Urban3D \cite{han2024whu} & 2024 & 393M & 18 (35) & \tno & MLS & \tyes &  \tyes & China \\
Ke-MLS \cite{kumar2026ke} & 2026 & 1{,}054M & 48 & \tno & MLS & \tyes &  \tyes & India \\
\rowcolor{hltint}\textbf{BhopalMLS} [Ours] & 2026 & 1{,}205.2M & 12/21 (62) & ${\sim}$40\% & MLS & \tyes &  \tyes & India \\
\bottomrule
\end{tabular}
}
\end{table}

\begin{figure*}[!htbp]
  \centering
  \includegraphics[width=\textwidth]{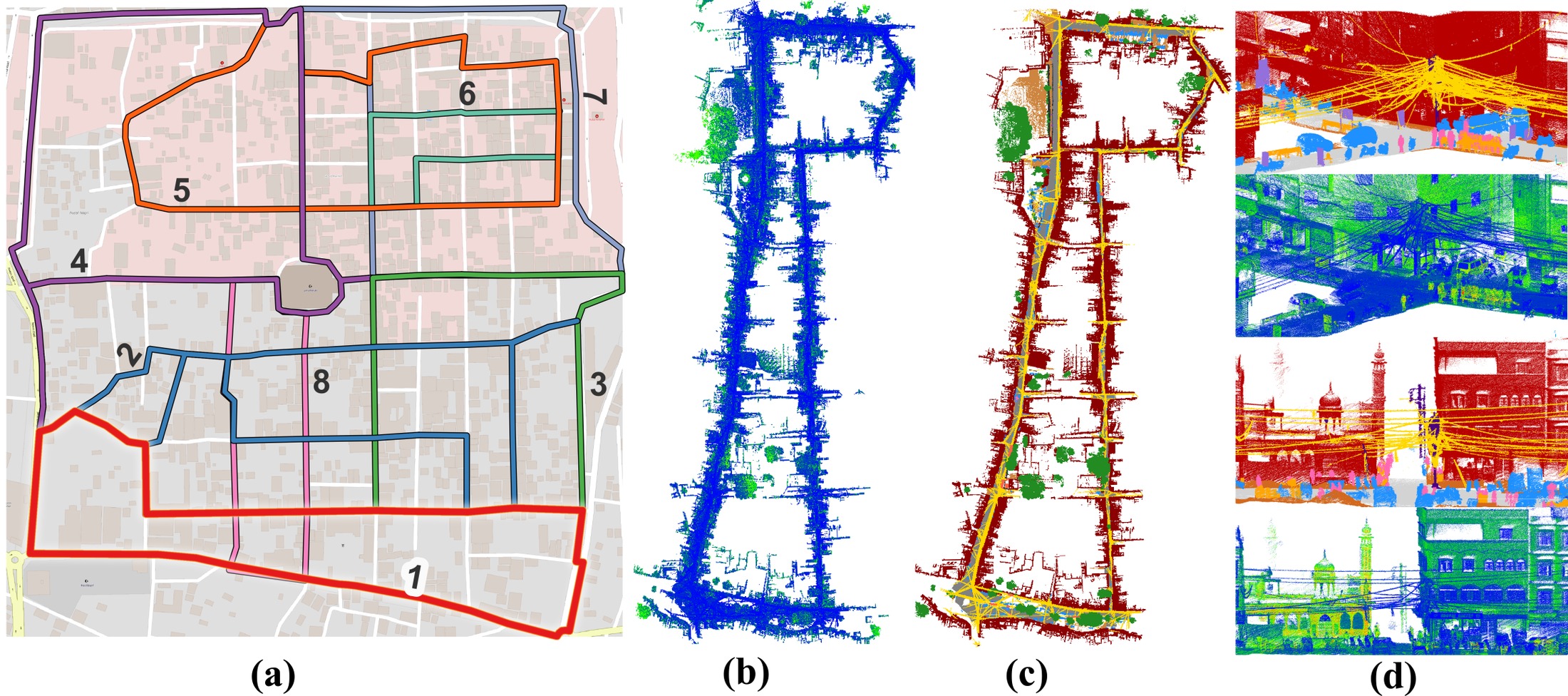}%
  \llap{\makebox[\textwidth][l]{\hspace{2.7pt}\raisebox{19.8pt}{{\setlength{\fboxsep}{1pt}\colorbox{white}{\fontsize{5}{5}\selectfont\textcopyright\ OpenStreetMap contributors}}}}}%
  \caption{\textbf{BhopalMLS LiDAR dataset study area and Path 1 visualization.}
  \textbf{(a)} Road network segmentation showing 8 distinct path segments with Path 1 highlighted in red.
  \textbf{(b)} Top view of Path 1 raw point cloud with intensity.
  \textbf{(c)} Top view of Path 1 annotated point cloud with semantic classification.
  \textbf{(d)} Multi-scale zoom instances with raw point cloud intensity and corresponding annotations (12 major classes with color code), demonstrating annotation quality and point density across diverse urban features.}
  \label{fig:study_area}
\end{figure*}

\section{Related Work}
\label{sec:related}

\subsection{3D Outdoor Scene Datasets}
MLS provides dense ground-level coverage of urban scenes, capturing detailed 3D geometry of roads, buildings, and infrastructure. Early MLS datasets such as Oakland \cite{munoz2009oakland} and Paris-rue-Madame \cite{serna2014paris} offered limited scale and class diversity, while larger efforts including Paris-Lille-3D \cite{roynard2018paris}, Toronto-3D \cite{tan2020toronto}, and TUM-MLS \cite{zhu2020tum} expanded spatial coverage and annotation granularity. SemanticKITTI \cite{behley2019semantickitti} and KITTI-360 \cite{liao2022kitti}, collected with vehicle-mounted spinning LiDAR, became widely adopted benchmarks for autonomous driving but capture a narrower vertical field of view due to the sensor's fixed mounting height. Complementary modalities include Terrestrial Laser Scanning (TLS), as in Semantic3D \cite{hackel2017semantic3d}, which provides ultra-dense static scans, and Airborne Laser Scanning (ALS), as in DALES \cite{varney2020dales}, which covers large areas at lower point density. More recently, WHU-Urban3D \cite{han2024whu} combined MLS and ALS, with 35 classes in its MLS part, and Campus3D \cite{li2020campus3d} provided dense UAV-photogrammetry coverage of a university campus. Closest to ours in geography, the recent Ke-MLS~\cite{kumar2026ke} is also an Indian MLS dataset, but it captures a \emph{semi-urban}, vegetation-dominated environment, in contrast to the dense, building-dominated commercial fabric of BhopalMLS. A summary is provided in Table~\ref{tab:dataset_comparison}. Despite this progress, all existing large-scale benchmarks are collected in well-planned urban environments with orderly layouts and standardized infrastructure. To our knowledge, no comparable MLS dataset captures the dense, unstructured, and heterogeneous urban fabric that BhopalMLS provides.

\subsection{3D Scene Perception}

3D semantic segmentation assigns semantic labels to each point in a point cloud, enabling detailed structural understanding of complex environments. Early architectures such as PointNet \cite{qi2017pointnet} and PointNet++ \cite{qi2017pointnet++} pioneered direct point-based learning through shared MLPs and hierarchical feature extraction. Subsequent methods including DGCNN \cite{wang2019dgcnn} and PointCNN \cite{li2018pointcnn} incorporated dynamic graph learning and $\chi$-convolutions, while kernel-based approaches such as KPConv \cite{thomas2019kpconv} improved geometric fidelity using deformable kernel operations. Transformer-based models, including Stratified Transformer \cite{lai2022stratified}, PTv2 \cite{wu2022point}, and PTv3 \cite{wu2024point}, extended self-attention to 3D domains for efficient long-range context modeling. Hybrid architectures such as KPConvX \cite{thomas2024kpconvx} and LitePT \cite{yuelitept2026} integrate convolutional priors with attention-driven reasoning to balance local precision and global awareness.

More recently, self-supervised pretraining has emerged as a powerful paradigm for 3D perception. Sonata \cite{wu2025sonata} learns reliable point representations through self-supervision, while Concerto \cite{zhang2025concerto} leverages joint 2D-3D self-supervised learning for transferable representations. Utonia \cite{zhang2026utonia} extends this direction toward a unified encoder for diverse point cloud domains. These models can be adapted via linear probing, decoder probing, LoRA, or fine-tuning, often matching or surpassing fully supervised baselines.

\section{Dataset Overview}

\subsection{Data Acquisition Setup}
Data were collected in the Azad Market region, located within the Peer Gate area of Bhopal, Madhya Pradesh, India. Azad Market is a dense commercial hub with residential and recreational spaces, characterized by narrow streets, mixed building types, heavy pedestrian and vehicular movement, and a complex arrangement of roadside establishments (see Fig.~\ref{fig:study_area} and \supfig{dataset-example}). The compact urban fabric, irregular geometry, and high levels of activity make it a challenging site for benchmarking 3D scene understanding in dense, unstructured urban environments.
A Velodyne VLP-16 LiDAR sensor integrated with a GeoSLAM ZEB Horizon handheld mapping system was employed for data acquisition. Although the sensor has only 16 channels, the GeoSLAM unit mounts it on a continuously rotating head that scans along all three axes, effectively multiplying the angular coverage. Combined with a low acquisition speed (on foot at approximately 3\,km/h, and on a slow-moving two-wheeler at comparable speed on longer stretches), this configuration produces dense urban point clouds with a mean point spacing of 4.9\,cm. Returns stream continuously and are SLAM-registered into one accumulated cloud, so there are no discrete frames: scenes are 5\,m \emph{spatial} segments and the point count measures surface sampling, not frame overlap. Our scale thus comes from sampling density and class diversity, not spatial extent or instance count. The handheld form factor also enabled flexible data collection through narrow and congested streets inaccessible to vehicle-mounted scanners. A 360$^\circ$ panoramic camera was co-mounted for colorization. GPS multipath in narrow lanes degrades the projection pose, so rather than assign uncertain color we apply a reprojection-consistency check and write per-point RGB only where it agrees with geometry (null elsewhere), giving validated registration for ${\sim}$40\% of points and trading coverage for correctness.

\subsection{Data Collection and Processing}
The study area was systematically divided into eight traversable paths (Path 01 to Path 08) covering both major and minor roads to achieve complete spatial coverage (see Fig.~\ref{fig:study_area}). Data were collected over three days between 12:00 p.m. and 3:00 p.m.; the survey spanned a total length of approximately 10 kilometers. Individual paths ranged from 0.66\,km to 1.85\,km in length (\suptab{path_diversity}).

A consistent low speed was maintained to ensure uniform point density and minimize motion artifacts. Overlapping coverage between adjacent paths was planned to reduce occlusions caused by buildings, vehicles, and other obstacles common in dense urban settings.

A DGPS-based survey was conducted to minimize positional errors and obtain accurately georeferenced data. Following acquisition, the raw point clouds were aligned using DGPS timestamps, and noise and artifacts were removed to refine spatial accuracy.

\begin{figure*}[!htbp]
  \centering
  \includegraphics[width=\textwidth]{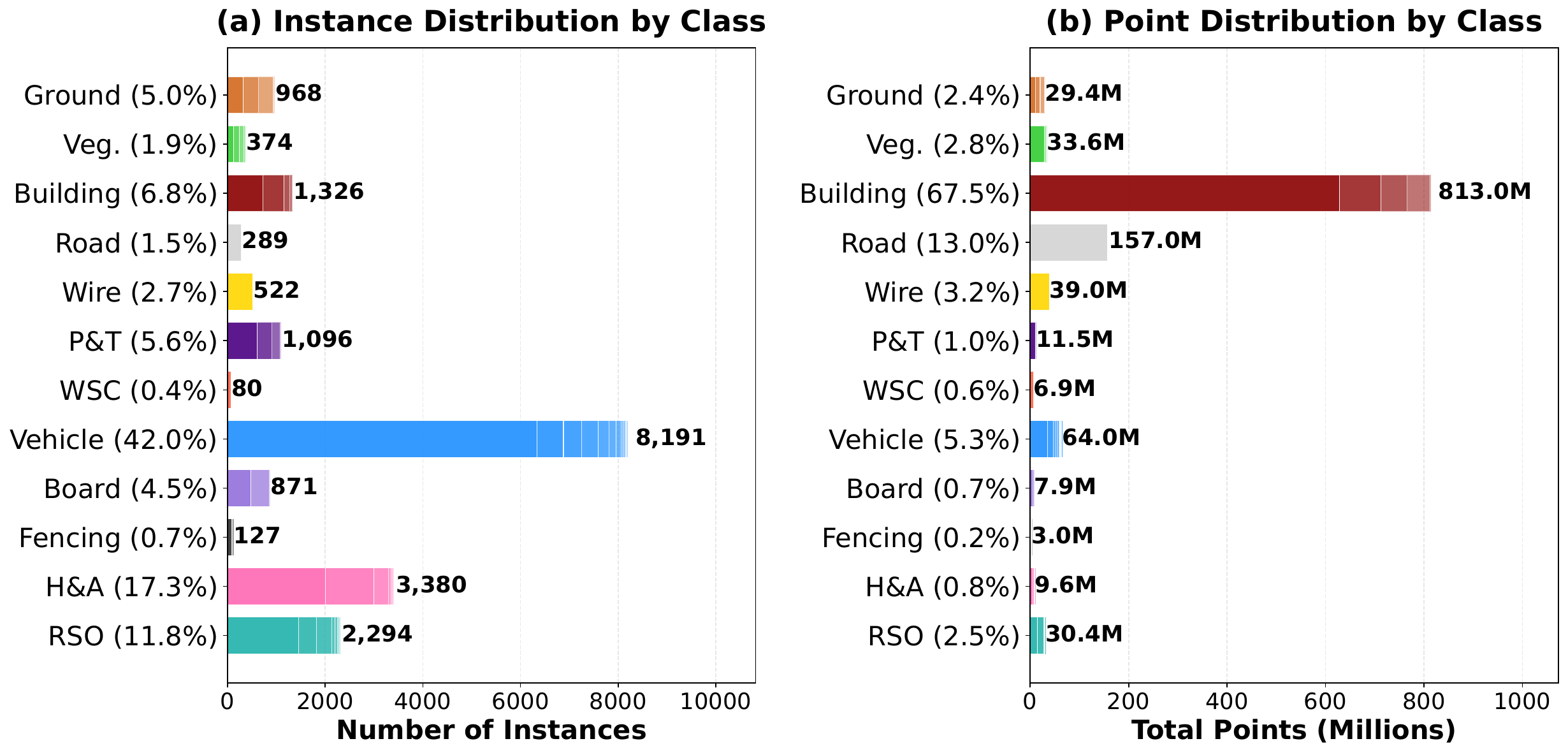}
  \caption{\textbf{BhopalMLS class distribution.}
\textbf{(a)} Instance count: Vehicle (42.0\%), Human and Animal (17.3\%), and Road Side Object (11.8\%) are the most frequent categories.
\textbf{(b)} Point count: Building (67.5\%) and Road (13.0\%) dominate spatial coverage (both panels over the 62 static classes; 66.0\% / 12.8\% over all 68 classes, Table~\ref{tab:class_distribution}).
Stacked segments represent subclass diversity within each major class (major-class totals in Table~\ref{tab:class_distribution}; full 62-subclass breakdown in \suptab{class_distribution_full}).
Class abbreviations (P\&T, WSC, H\&A, RSO, Veg.) are defined in Table~\ref{tab:class_distribution}.}
  \label{fig:class_dist_full}
\end{figure*}

\subsection{Annotation Process}

\begin{figure}[t]
\centering
\includegraphics[width=\linewidth]{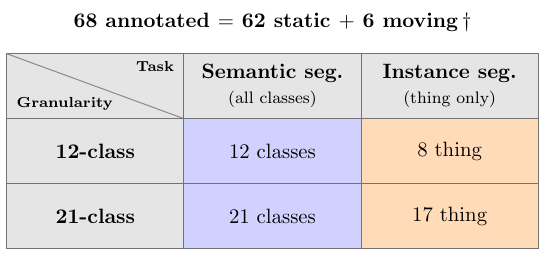}
\caption{\textbf{Taxonomy and tasks.} The benchmark comprises \textbf{68} annotated classes ($=$ \textbf{62 static} $+$ \textbf{6 moving}\,$\dagger$). The dataset supports \textbf{semantic} segmentation (all classes) and \textbf{instance} segmentation (thing classes; the 4 stuff classes are excluded from instance grouping), each at \textbf{12-class} (main) and \textbf{21-class} (supplementary) granularity; instance results are in supplementary Tables~\ref*{S-tab:instance_seg_12c}--\ref*{S-tab:instance_seg_21c}. The $\dagger$ variants train jointly on the 6 moving subclasses, folded into their static parents (\textit{Vehicle}, \textit{Human \& Animal}) at evaluation. The 21-class taxonomy refines Building, Pole \& Tower, Vehicle, Road Side Object, and Ground, and merges the rare Wire-Structure Connector into Pole (other).}
\label{fig:taxonomy}
\end{figure}

The dataset was annotated with 62 fine-grained static semantic categories, selected to balance three criteria. \emph{Comparability}: the 12 major classes match those common to existing MLS benchmarks (Building, Road, Ground, Vegetation, Pole, Vehicle, Fencing), so standard mIoU and cross-dataset transfer apply directly. \emph{Region-specific coverage}: dedicated classes for objects absent from those benchmarks but pervasive in dense Indian corridors (auto-/cycle-rickshaws, hand carts, shop shades, mannequins, wire-structure connectors). \emph{Instance support}: the distribution is extremely long-tailed (\suptab{class_distribution_full}), with \textit{Building} and \textit{Road} alone holding 80.5\% of static-class points, while twelve of the 62 classes carry $\leq$5 instances (six just one) and are merged into their nearest parent rather than kept as standalone categories. A further 6 moving-class subclasses extend the taxonomy to 68 classes (detailed below). The 62 static classes form two taxonomies (Fig.~\ref{fig:taxonomy}): a coarse \textbf{12-class} setting (main; Fig.~\ref{fig:class_dist_full}, Table~\ref{tab:class_distribution}) and a finer \textbf{21-class} setting (supplementary) that refines only the well-populated categories (Building, Pole \& Tower, Vehicle, Road Side Object, Ground; e.g., 2-/3-/4-Wheeler and Hand Cart, each with hundreds of instances) while the sparse tail stays grouped.

The primary benchmark uses the 62 static classes (19{,}518 instances, 1{,}205.2M points). An additional 3 categories (unlabeled, High Noise, Low Noise; 5{,}124 instances) are excluded entirely from benchmarking. The 6 moving-class subclasses (dynamic humans, vehicles, riders, and animals; 7{,}181 instances, 26.2M points) capture dynamic objects whose LiDAR points form motion-induced tails during scanning (examples in \supfig{class-samples-grid}). Because this tail geometry is inconsistent with that of their static counterparts, these subclasses are folded into their static parents (\textit{Vehicle}, \textit{Human \& Animal}) at evaluation; they are retained as separate fine-grained labels (68-class taxonomy) for use as additional supervision during training and for future work on dynamic-object detection.

\begin{table}[!htbp]
\centering
\small
\setlength{\tabcolsep}{8pt}
\renewcommand{\arraystretch}{1.05}
\caption{%
\textbf{BhopalMLS class distribution (12 major classes).}
The primary benchmark uses \textbf{62 static} semantic classes (19{,}518 instances, 1{,}205.2M points) grouped into the 12 major categories below; \textbf{6 moving-class subclasses} (7{,}181 instances, 26.2M points) extend the taxonomy to \textbf{68 classes} and are folded into their static parents (\textit{Vehicle}, \textit{Human \& Animal}) at evaluation. Percentages are over all 68 classes (1{,}231.4M points). \colorbox{hltint}{Shaded} \textit{Building}/\textit{Road} hold 78.8\% of points. The full 62-subclass breakdown is given in \suptab{class_distribution_full}.
}
\label{tab:class_distribution}
\resizebox{\columnwidth}{!}{%
\begin{tabular}{l rrr}
\toprule
\textbf{Major Class} & \textbf{Inst.} & \textbf{Pts (M)} & \textbf{Pts (\%)} \\
\midrule
\rowcolor{hltint} Building & 1{,}326 & 813.0 & 66.0 \\
\rowcolor{hltint} Road & 289 & 157.0 & 12.8 \\
Vehicle & 9{,}888 & 77.2 & 6.3 \\
Wire & 522 & 39.0 & 3.2 \\
Vegetation (Veg.) & 374 & 33.6 & 2.7 \\
Road Side Object (RSO) & 2{,}294 & 30.4 & 2.5 \\
Ground & 968 & 29.4 & 2.4 \\
Human \& Animal (H\&A) & 8{,}864 & 22.5 & 1.8 \\
Pole \& Tower (P\&T) & 1{,}096 & 11.5 & 0.9 \\
Board & 871 & 7.9 & 0.6 \\
Wire-Structure Connector (WSC) & 80 & 6.9 & 0.6 \\
Fencing & 127 & 3.0 & 0.2 \\
\midrule
\textbf{Combined (68 classes)} & \textbf{26{,}699} & \textbf{1{,}231.4} & \textbf{100.0} \\
\bottomrule
\end{tabular}%
}
\end{table}

\subsubsection{Annotation Methodology}

To manage complexity, each path was divided into smaller segments, with two independent annotators labeling each segment. As a measure of annotation reliability, we computed the point-level inter-annotator disagreement rate between the two annotators for each path; mismatched points were then jointly reviewed and corrected until this rate fell below 0.5\% per path. Paths exceeding this threshold were relabeled by a third annotator, and a reviewer manually verified all labeled paths, resolving any remaining discrepancies through discussion with the annotators.

A Standard Operating Procedure (SOP) was developed to ensure consistency and quality, as no established guidelines existed for LiDAR annotation of this complexity.
Annotators also used 360$^\circ$ RGB images to assist with difficult cases, particularly for distinguishing between similar object types or handling occlusions.

\subsubsection{Point Cloud Attributes and Data Formats}

The dataset is released in LAS, PLY, HDF5, CSV, NPZ and BIN formats. Each point carries georeferenced XYZ (stored as float64), per-point RGB from co-registered 360$^\circ$ panoramas (valid for ${\sim}$40\% of points, zero elsewhere), return intensity, GPS time, a semantic label (62 static classes; 68 including the moving subclasses), and a 9-digit instance identifier encoding path, class hierarchy and instance number. Field-level specifications are given in the supplementary.

\begin{table}[!htbp]
\centering
\small
\setlength{\tabcolsep}{2pt}
\renewcommand{\arraystretch}{0.9}
\caption{Quantitative comparison of \textbf{BhopalMLS}~(BM) against four MLS/AD-LiDAR datasets: Ke-MLS~(KM)~\cite{kumar2026ke}, Paris-Lille-3D~(PL)~\cite{roynard2018paris}, Toronto-3D~(TR)~\cite{tan2020toronto}, and SemanticKITTI~(SK)~\cite{behley2019semantickitti}, under a fair voxel-normalized protocol (5\,m$\times$5\,m XY tiles, 25\,cm voxel). Mann--Whitney $p<10^{-170}$ for every Indian-vs-Western comparison; Cohen's $d \in [0.46, 2.59]$. \colorbox{hltint}{Tinted} columns: Indian (BM, KM); \textbf{bold}: Indian $>$ all Western.}
\label{tab:dataset_complexity}
\resizebox{\columnwidth}{!}{%
\begin{tabular}{l|>{\columncolor{hltint}}c>{\columncolor{hltint}}c|ccc}
\toprule
& \multicolumn{2}{c|}{\cellcolor{hltint}\textbf{\textit{Indian MLS}}} & \multicolumn{3}{c}{\textbf{\textit{Western}}} \\
\cmidrule(lr){2-3} \cmidrule(lr){4-6}
\textbf{Metric} & \textbf{BM} & \textbf{KM} & \textbf{PL} & \textbf{TR} & \textbf{SK} \\
& (Ours) & & & & \\
\textit{Sensor} & {\scriptsize handheld} & {\scriptsize vehicle} & {\scriptsize vehicle} & {\scriptsize vehicle} & {\scriptsize spinning} \\
& {\scriptsize ZEB} & {\scriptsize Riegl VQ-450} & {\scriptsize Velo-32} & {\scriptsize MLS} & {\scriptsize 64-ch} \\
\textit{Region} & {\scriptsize India} & {\scriptsize India} & {\scriptsize France} & {\scriptsize Canada} & {\scriptsize Germany} \\
\midrule
\multicolumn{6}{l}{\textit{Density}} \\
Point spacing & & & & & \\
(mean, cm) & \multirow{-2}{*}{4.9} & \multirow{-2}{*}{4.9} & \multirow{-2}{*}{3.8} & \multirow{-2}{*}{9.1} & \multirow{-2}{*}{10.6} \\
Point density (pts/m$^2$) & 2{,}829 & 1{,}787 & 1{,}519 & 466 & 63.6 \\
Voxel density & & & & & \\
(pts/voxel) & \multirow{-2}{*}{28.9} & \multirow{-2}{*}{33.3} & \multirow{-2}{*}{39.9} & \multirow{-2}{*}{14.4} & \multirow{-2}{*}{4.2} \\
\midrule
\multicolumn{6}{l}{\textit{Heterogeneity} ($\uparrow$)} \\
Density CoV & \textbf{1.15} & \textbf{1.33} & 0.97 & 0.83 & 0.58 \\
Density Gini & \textbf{0.53} & \textbf{0.55} & 0.47 & 0.37 & 0.29 \\
Density variance & \textbf{3{,}927} & \textbf{6{,}589} & 1{,}988 & 1{,}804 & 36.6 \\
\midrule
\multicolumn{6}{l}{\textit{Complexity} ($\uparrow$)} \\
Z-height extent (m) & \textbf{10.1} & \textbf{11.6} & 8.4 & 7.7 & 2.2 \\
\midrule
\textit{Tiles sampled ($n$)} & 29.1k & 55.7k & 3.2k & 4.4k & 74.4k \\
\bottomrule
\end{tabular}
}%
\end{table}

\section{Benchmarks}
\subsection{Data Partitioning}

Data from paths 1--4 are used for training, 5--6 for validation, and 7--8 for testing, following an approximate 60/21/19 point ratio (\supfig{split_distribution}). This path-based partitioning strategy, similar to S3DIS \cite{armeni20163d} and Waymo \cite{sun2020waymo}, ensures that each subset represents different yet comparable urban areas while preserving spatial consistency.

The dataset follows a long-tailed class distribution: \textit{Building} (66.0\% of points) and \textit{Road} (12.8\%) dominate, while rare classes such as \textit{Fencing} (0.2\%) and \textit{Wire-Structure Connector} (0.6\%) are underrepresented. This imbalance reflects real urban conditions and provides a realistic benchmark. To mitigate class imbalance during training, we apply two strategies: (1) at the data level, the dominant \textit{Building} class is downsampled by 10\% during scene construction, and (2) at the loss level, we use square-root scaled inverse-frequency class weights in the cross-entropy loss combined with a Lov\'asz loss, ensuring that rare classes receive higher gradient contributions without destabilizing training.

\subsection{Evaluation Metrics}

To ensure comparability with established 3D benchmarks, we adopt the mean Intersection-over-Union (mIoU) as our primary evaluation metric, consistent with protocols used in SemanticKITTI \cite{behley2019semantickitti}, S3DIS \cite{armeni20163d}, and SensatUrban \cite{hu2021semantic3d}.
For each class $i$, $\text{IoU}_i = c_{ii}/(c_{ii} + \sum_{j \neq i} c_{ij} + \sum_{k \neq i} c_{ki})$ (true positives over the union of predicted and ground-truth points), and the mean IoU is $\overline{\text{IoU}} = \frac{1}{N} \sum_{i=1}^{N} \text{IoU}_i$ over the $N$ classes.

We also report Overall Accuracy (OA) and mean class Accuracy (mAcc) for completeness, though OA can overestimate performance on imbalanced datasets \cite{hackel2017semantic3d, tan2020toronto}. mIoU serves as the primary metric for all comparisons.

\subsection{Semantic Segmentation Benchmark}

Table~\ref{tab:bhopal_iou_benchmark_style_rotated_vline} reports the 12-class semantic segmentation benchmark, from which two findings emerge: per-class accuracy is governed more by geometric distinctiveness than by class frequency, and pretrained foundation models with fine-tuning surpass supervised baselines trained from scratch.

Per-class, the highest IoU scores are achieved either by abundant classes (\textit{Vegetation}, \textit{Building}, \textit{Vehicle}, \textit{Road}) or by classes with strong, spatially continuous geometric signatures despite lower frequency (\textit{Wire}). Notably, \textit{Wire-Structure Connector} achieves 85.26\% IoU despite its rarity (0.6\% of points), indicating that distinctive geometric cues can compensate for low frequency. In contrast, \textit{Fencing} and \textit{Ground} remain the most challenging classes, as discussed in Section~\ref{sec:challenges}.

Among the benchmarked approaches, pretrained foundation models outperform fully supervised baselines: Utonia(L) with fine-tuning (trained jointly with the 6 moving-class subclasses, marked $^{\dagger}$) achieves the best mIoU of 72.12\% under our standard training recipe, with Concerto(L) close behind at 71.06\%, both surpassing PTv3 (68.04\%) trained from scratch. This indicates that large-scale pretraining transfers effectively to these complex urban scenes. However, this recipe is not optimal for every architecture: an ablation (\suptab{class_imbalance_ablation}) shows that removing the Lov\'asz term raises Utonia(L) by 1.92 mIoU while slightly lowering Concerto(L), an architecture-dependent interaction with the loss.

Despite only approximately 40\% of points having valid RGB values, switching from geometry-only inputs (3ch/4ch) to 9ch (XYZ + partial RGB + normals) improves performance in 5 of 7 model comparisons (supervised 4ch$\rightarrow$9ch and linear-probe 3ch$\rightarrow$9ch), with Sonata(L) LP gaining +11.14 mIoU. Even incomplete color therefore provides useful discriminative cues, especially for foundation models pretrained with color.

Comparing pretraining strategies, Concerto(L) (general-purpose) consistently outperforms Concerto(O) (outdoor) across all adaptation methods (e.g., 71.06\% vs 68.10\% FT-9ch), indicating that general-purpose pretraining transfers better to MLS data than outdoor-specific pretraining. Utonia(L) also achieves the best mAcc (83.89\%), while Concerto(L) FT attains the best per-class IoU on several structural categories (Building 93.71\%, Wire 87.91\%, and Wire-Structure Connector 85.26\%; Table~\ref{tab:bhopal_iou_benchmark_style_rotated_vline}). Fine-grained 21-class results, covering India-specific categories such as 3-Wheeler, Hand Cart, Shop Shade, and Mannequin, are reported in \suptab{semseg_21c_full}, along with qualitative predictions and instance segmentation results.

\begin{table*}[t]
\centering
\small
\setlength{\tabcolsep}{4pt}
\renewcommand{\arraystretch}{0.92}
\caption{Semantic segmentation benchmark on the \textbf{BhopalMLS} dataset (12-class). LP = Linear Probe, DP = Decoder Probe, LoRA = Low-Rank Adaptation, FT = Fine-Tuning. 3ch = XYZ; 4ch = XYZ + intensity; 9ch = XYZ + Partial RGB + normals. Sonata(L), Concerto(L) and Utonia(L) = Large (general-purpose pretrained), Concerto(O) = outdoor-pretrained. $^{\dagger}$=trained jointly with the 6 moving-class subclasses (Moving Vehicle, Moving Humans, Moving Female, Moving Male, Moving Animal, Mobike Rider Moving) as additional fine-grained supervision; folded into their static parents at evaluation. \colorbox{bestgreen}{\textbf{Bold green}} = overall best per column; \underline{underline} = best within category; \textcolor{red}{red} = overall lowest.}
\label{tab:bhopal_iou_benchmark_style_rotated_vline}
\resizebox{\textwidth}{!}{%
\begin{tabular}{l|ccc|cccccccccccc}
\hline
\textbf{Model} &
\rotatebox{90}{\textbf{OA (\%)}} &
\rotatebox{90}{\textbf{mAcc (\%)}} &
\rotatebox{90}{\textbf{mIoU (\%)}} &
\rotatebox{90}{\textbf{Board}} &
\rotatebox{90}{\textbf{Building}} &
\rotatebox{90}{\textbf{Fencing}} &
\rotatebox{90}{\textbf{Ground}} &
\rotatebox{90}{\textbf{Human\&Animal}} &
\rotatebox{90}{\textbf{Pole\&Tower}} &
\rotatebox{90}{\textbf{Road}} &
\rotatebox{90}{\textbf{Road Side Obj.}} &
\rotatebox{90}{\textbf{Vegetation}} &
\rotatebox{90}{\textbf{Vehicle}} &
\rotatebox{90}{\textbf{Wire}} &
\rotatebox{90}{\textbf{Wire-Structure Conn.}} \\
\hline
\multicolumn{16}{l}{\textbf{Supervised Learning -- 4ch (XYZ + intensity)}} \\
\hline
PTv3 \cite{wu2024point} & \underline{92.60} & 78.75 & \underline{67.88} & 57.05 & \underline{92.75} & 13.87 & 22.24 & \underline{66.09} & \underline{69.41} & \underline{87.39} & \underline{51.03} & 97.11 & \underline{91.19} & \underline{87.74} & 78.75 \\
LitePT-L \cite{yuelitept2026} & 91.13 & \underline{80.45} & 67.53 & \underline{57.19} & 91.47 & 20.24 & 22.10 & 63.88 & 66.69 & 85.07 & 48.76 & 97.18 & 88.23 & 86.06 & \underline{83.47} \\
KpConvX-L \cite{thomas2024kpconvx} & 92.00 & 80.10 & 66.78 & 55.82 & 92.31 & \underline{20.74} & \underline{24.28} & 60.30 & 62.87 & 86.20 & 49.12 & \underline{97.24} & 89.40 & 86.25 & 76.80 \\
\hline
\multicolumn{16}{l}{\textbf{Supervised Learning -- 9ch (XYZ + Partial RGB + normals)}} \\
\hline
PTv3 & \underline{92.17} & \underline{79.57} & \underline{68.04} & \underline{59.37} & 92.42 & 18.18 & 22.69 & 62.22 & \underline{68.23} & 86.65 & \underline{49.37} & \underline{97.25} & \underline{90.78} & \underline{86.61} & \underline{82.71} \\
LitePT-L & 91.16 & 77.54 & 65.18 & 48.18 & 91.95 & \underline{21.39} & 24.47 & 61.46 & 62.40 & 85.13 & 43.10 & 96.25 & 85.83 & 82.22 & 79.78 \\
KpConvX-L & 92.02 & 79.14 & 66.97 & 56.82 & \underline{92.43} & 20.90 & \underline{24.70} & \underline{62.90} & 63.23 & \underline{86.69} & 47.99 & 96.76 & 89.30 & 86.07 & 75.79 \\
\hline
\multicolumn{16}{l}{\textbf{Linear Probe (3ch: XYZ)}} \\
\hline
Sonata(L) \cite{wu2025sonata} & \textcolor{red}{84.38} & \textcolor{red}{62.25} & \textcolor{red}{44.60} & \textcolor{red}{27.93} & \textcolor{red}{84.90} & 6.58 & \textcolor{red}{19.28} & \textcolor{red}{28.34} & \textcolor{red}{36.00} & \textcolor{red}{82.94} & \textcolor{red}{29.38} & \textcolor{red}{78.43} & \textcolor{red}{63.95} & \textcolor{red}{57.40} & \textcolor{red}{20.11} \\
Concerto(O) & 87.32 & 69.90 & 52.18 & 33.74 & 88.21 & 16.78 & 21.92 & 35.24 & 41.76 & 83.92 & 34.43 & 91.80 & 72.67 & 65.87 & 39.85 \\
Concerto(L) \cite{zhang2025concerto} & 90.64 & 77.06 & 61.29 & 53.06 & 91.18 & 21.57 & 23.04 & \underline{48.59} & 54.92 & \underline{86.49} & 42.87 & 95.45 & 81.49 & 80.70 & 56.07 \\
Utonia(L) \cite{zhang2026utonia} & \underline{90.90} & \underline{77.10} & \underline{63.16} & \underline{55.51} & \underline{91.50} & \underline{21.95} & \underline{24.70} & 46.35 & \underline{58.96} & 85.86 & \underline{45.32} & \underline{96.55} & \underline{86.45} & \underline{81.46} & \underline{63.28} \\
\hline
\multicolumn{16}{l}{\textbf{Linear Probe (9ch: XYZ + Partial RGB + normals)}} \\
\hline
Sonata(L) & 88.68 & 74.59 & 55.74 & 41.87 & 89.51 & 22.88 & 23.13 & 37.76 & 48.10 & 85.54 & 38.68 & 92.75 & 74.23 & 75.36 & 39.04 \\
Concerto(O) & 88.70 & 68.01 & 52.41 & 35.69 & 89.54 & 15.17 & 22.54 & 36.72 & 40.83 & 85.58 & 34.44 & 92.80 & 75.11 & 70.78 & 29.74 \\
Concerto(O)$^{\dagger}$ & 88.51 & 69.39 & 54.19 & 36.03 & 89.40 & 15.96 & 22.25 & 53.69 & 41.55 & 85.79 & 35.22 & 93.45 & 72.43 & 71.69 & 32.86 \\
Concerto(L) & 89.81 & 78.11 & 60.34 & 52.32 & 90.08 & 24.61 & 22.56 & 45.30 & 54.24 & 86.33 & 41.91 & 95.39 & 81.00 & 78.66 & 51.67 \\
Utonia(L) & \underline{91.32} & 79.55 & 64.18 & \underline{56.74} & \underline{91.72} & 34.49 & \underline{25.02} & 47.40 & 58.64 & \underline{87.43} & \underline{44.77} & \underline{96.76} & \underline{86.25} & 81.47 & 59.44 \\
Utonia(L)$^{\dagger}$ & 91.14 & \underline{80.04} & \underline{65.74} & 56.04 & 91.51 & \underline{37.63} & 25.00 & \underline{62.95} & \underline{58.95} & 87.42 & 43.57 & 96.61 & 84.05 & \underline{81.73} & \underline{63.40} \\
\hline
\multicolumn{16}{l}{\textbf{Decoder Probe (9ch: XYZ + Partial RGB + normals)}} \\
\hline
Sonata(L) & 92.31 & 79.05 & 67.52 & \underline{61.80} & 92.79 & 19.81 & 25.04 & 60.16 & 64.19 & 86.80 & \underline{49.82} & 96.91 & 89.32 & 85.31 & 78.27 \\
Concerto(O) & 91.47 & 76.31 & 63.37 & 54.85 & 92.10 & 9.32 & 23.87 & 55.94 & 58.94 & 86.15 & 47.24 & 96.28 & 87.85 & 82.44 & 65.53 \\
Concerto(O)$^{\dagger}$ & 91.83 & 74.46 & 64.19 & 51.18 & 92.31 & 13.32 & 23.03 & 73.24 & 58.69 & 86.50 & 45.56 & 95.84 & 87.83 & 82.89 & 59.86 \\
Concerto(L) & 92.23 & 78.74 & 67.05 & 61.75 & 92.64 & 11.33 & 24.25 & 62.20 & \underline{66.03} & 86.81 & 48.32 & 97.21 & 89.78 & 85.02 & \underline{79.25} \\
Utonia(L) & 91.89 & \underline{79.53} & 67.03 & 58.76 & 92.48 & \underline{21.41} & 25.74 & 65.76 & 60.52 & 85.84 & 47.72 & 97.17 & \underline{91.35} & 84.50 & 73.10 \\
Utonia(L)$^{\dagger}$ & \underline{92.84} & 77.11 & \underline{68.22} & 56.46 & \underline{93.18} & 10.56 & \underline{26.25} & \underline{79.63} & 65.16 & \underline{87.21} & 47.41 & \underline{97.48} & 90.91 & \underline{87.12} & 77.29 \\
\hline
\multicolumn{16}{l}{\textbf{LoRA (9ch: XYZ + Partial RGB + normals)}} \\
\hline
Sonata(L) & 90.76 & 79.48 & 64.07 & 56.20 & 91.16 & 24.44 & 24.65 & 47.86 & 58.65 & 86.37 & 44.18 & 95.68 & 84.65 & 81.48 & 73.51 \\
Concerto(O) & 91.27 & 80.07 & 65.29 & 57.71 & 91.97 & 26.00 & 25.08 & 49.50 & 59.79 & 85.24 & \underline{48.78} & 96.87 & 87.54 & 81.42 & 73.57 \\
Concerto(L) & 91.69 & 80.64 & 66.57 & 59.99 & 92.01 & 26.42 & 24.85 & 54.16 & 61.98 & 87.09 & 47.63 & 96.78 & 87.64 & 83.36 & 76.91 \\
Utonia(L) & \underline{92.60} & \underline{81.76} & \underline{69.77} & \underline{63.84} & \underline{92.76} & \underline{38.12} & \cellcolor{bestgreen}\underline{\textbf{27.02}} & \underline{59.85} & \underline{66.77} & \cellcolor{bestgreen}\underline{\textbf{87.94}} & 48.33 & \cellcolor{bestgreen}\underline{\textbf{97.82}} & \underline{90.86} & \underline{85.29} & \underline{78.64} \\
\hline
\multicolumn{16}{l}{\textbf{Fine-Tuning (9ch: XYZ + Partial RGB + normals)}} \\
\hline
Sonata(L) & 92.12 & 83.30 & 68.72 & 60.93 & 92.25 & 28.45 & 24.21 & 62.17 & 63.55 & 87.23 & 52.05 & 97.27 & 92.56 & 84.02 & 79.96 \\
Concerto(O) & 92.58 & 78.78 & 68.10 & 66.56 & 92.81 & \textcolor{red}{6.57} & 25.42 & 64.30 & 68.20 & \underline{87.40} & 51.71 & 97.03 & 92.04 & 84.63 & 80.54 \\
Concerto(O)$^{\dagger}$ & 92.70 & 79.89 & 70.41 & 59.53 & 92.97 & 25.31 & 24.80 & 80.61 & 68.68 & 86.41 & 47.96 & 97.37 & 91.44 & 87.29 & 82.57 \\
Concerto(L) & \cellcolor{bestgreen}\underline{\textbf{93.42}} & 80.44 & 71.06 & \cellcolor{bestgreen}\underline{\textbf{67.72}} & \cellcolor{bestgreen}\underline{\textbf{93.71}} & 20.14 & 25.41 & 69.88 & \cellcolor{bestgreen}\underline{\textbf{72.48}} & 87.11 & \cellcolor{bestgreen}\underline{\textbf{53.07}} & 97.67 & 92.34 & \cellcolor{bestgreen}\underline{\textbf{87.91}} & \cellcolor{bestgreen}\underline{\textbf{85.26}} \\
Utonia(L) & 92.69 & \cellcolor{bestgreen}\underline{\textbf{83.89}} & 71.28 & 67.36 & 93.38 & \cellcolor{bestgreen}\underline{\textbf{38.79}} & 26.27 & 64.24 & 68.61 & 85.66 & 51.51 & 97.41 & 91.54 & 87.53 & 83.03 \\
Utonia(L)$^{\dagger}$ & 93.32 & 81.90 & \cellcolor{bestgreen}\underline{\textbf{72.12}} & 64.92 & 93.58 & 28.27 & \underline{26.40} & \cellcolor{bestgreen}\underline{\textbf{82.44}} & 69.25 & 87.19 & 52.31 & \underline{97.80} & \cellcolor{bestgreen}\underline{\textbf{92.87}} & 87.78 & 82.60 \\
\hline
\end{tabular}%
}%
\end{table*}

\subsection{Instance Segmentation Benchmark}
BhopalMLS provides unique instance IDs for all 19{,}518 annotated objects, enabling instance segmentation without additional labeling. Following the standard \emph{thing}/\emph{stuff} convention, only thing classes (countable objects, \eg vehicles, poles, humans) are clustered and scored, while the four stuff classes (Ground, Vegetation, Road, Wire) are predicted semantically but excluded from instance grouping; this leaves 8 thing classes in the 12-class setting and 17 in the 21-class setting. Using a PointGroup~\cite{jiang2020pointgroup} head with 9ch input on the same test split, fine-tuning again dominates: Utonia(L)~FT attains the best 44.07\% mAP50 (12-class, $2.9\times$ the best linear probe) and 39.83\% mAP50 (21-class), ahead of Sonata(L) and Concerto(L), while supervised PTv3 stays competitive. Thin \textit{Fencing} and small India-specific objects (Shop Shade, Mannequin) remain the hardest, mirroring the semantic-segmentation trends. Full per-class results and qualitative predictions are reported in supplementary Tables~\ref*{S-tab:instance_seg_12c}--\ref*{S-tab:instance_seg_21c} and Figs.~\ref*{S-fig:instance-predictions-12c}--\ref*{S-fig:instance-predictions-21c}.

\section{Analysis and Open Challenges}
\label{sec:challenges}

\definecolor{bmsrcgreen}{rgb}{0.85,0.95,0.85}
\definecolor{tgtsrcred}{rgb}{1.00,0.88,0.88}

\setlength{\textfloatsep}{4pt plus 2pt minus 1pt}
\begin{table}[t]
\centering
\footnotesize
\captionsetup{skip=2pt}
\setlength{\tabcolsep}{3pt}
\renewcommand{\arraystretch}{1.0}
\caption{\textbf{Cross-region generalization} of Utonia(L) and Concerto(L) (FT-9ch, native resolution). $n_c$ = number of classes shared by the BM-12 taxonomy and the target dataset. Ke-MLS (KM) provides no RGB, so KM rows are geometry-only (9ch with zero RGB). Per-class IoU, and Concerto(O) on the SK pairs and BM$\rightarrow$PL, are in \suptab{cross_city_full}. \colorbox{bmsrcgreen}{Green rows}: BM-source (BM$\rightarrow$target). \colorbox{tgtsrcred}{Red rows}: target-source (target$\rightarrow$BM). \textbf{Bold}: the higher mIoU of each BM$\leftrightarrow$target pair.}
\label{tab:cross_region_1cm}
\resizebox{0.92\columnwidth}{!}{%
\begin{tabular}{l|l|c|cc|cc}
\toprule
& \textbf{Source $\rightarrow$} & & \multicolumn{2}{c|}{\textbf{Utonia(L)}} & \multicolumn{2}{c}{\textbf{Concerto(L)}} \\
\cmidrule(lr){4-5} \cmidrule(lr){6-7}
\textbf{Category} & \textbf{Target} & \textbf{$n_c$} & \textbf{mAcc} & \textbf{mIoU} & \textbf{mAcc} & \textbf{mIoU} \\
\midrule
\multirow{5}{*}{\begin{tabular}[c]{@{}l@{}}\textbf{Within-}\\\textbf{domain}\end{tabular}}
 & BM $\rightarrow$ BM & 12 & 83.89 & 71.28 & 80.44 & 71.06 \\
 & KM $\rightarrow$ KM & 12 & 77.05 & 66.66 & 74.79 & 67.58 \\
 & TR $\rightarrow$ TR & 7 & 93.25 & 88.38 & 93.25 & 86.96 \\
 & PL $\rightarrow$ PL & 11 & 92.11 & 84.77 & 92.03 & 83.43 \\
 & SK $\rightarrow$ SK & 9 & 84.86 & 77.14 & 83.67 & 75.09 \\
\midrule
\multirow{2}{*}{\begin{tabular}[c]{@{}l@{}}\textbf{Intra-}\\\textbf{Indian}\end{tabular}}
 & \cellcolor{bmsrcgreen}BM $\rightarrow$ KM & \cellcolor{bmsrcgreen}12 & \cellcolor{bmsrcgreen}74.94 & \cellcolor{bmsrcgreen}\textbf{56.78} & \cellcolor{bmsrcgreen}66.43 & \cellcolor{bmsrcgreen}\textbf{49.23} \\
 & \cellcolor{tgtsrcred}KM $\rightarrow$ BM & \cellcolor{tgtsrcred}12 & \cellcolor{tgtsrcred}53.90 & \cellcolor{tgtsrcred}33.96 & \cellcolor{tgtsrcred}51.29 & \cellcolor{tgtsrcred}32.70 \\
\midrule
\multirow{6}{*}{\begin{tabular}[c]{@{}l@{}}\textbf{Inter-}\\\textbf{region}\end{tabular}}
 & \cellcolor{bmsrcgreen}BM $\rightarrow$ TR & \cellcolor{bmsrcgreen}7 & \cellcolor{bmsrcgreen}90.36 & \cellcolor{bmsrcgreen}\textbf{77.27} & \cellcolor{bmsrcgreen}84.67 & \cellcolor{bmsrcgreen}\textbf{75.20} \\
 & \cellcolor{tgtsrcred}TR $\rightarrow$ BM & \cellcolor{tgtsrcred}7 & \cellcolor{tgtsrcred}60.66 & \cellcolor{tgtsrcred}51.65 & \cellcolor{tgtsrcred}49.52 & \cellcolor{tgtsrcred}39.11 \\
 & \cellcolor{bmsrcgreen}BM $\rightarrow$ PL & \cellcolor{bmsrcgreen}11 & \cellcolor{bmsrcgreen}76.95 & \cellcolor{bmsrcgreen}\textbf{55.68} & \cellcolor{bmsrcgreen}67.55 & \cellcolor{bmsrcgreen}\textbf{48.84} \\
 & \cellcolor{tgtsrcred}PL $\rightarrow$ BM & \cellcolor{tgtsrcred}11 & \cellcolor{tgtsrcred}40.42 & \cellcolor{tgtsrcred}27.06 & \cellcolor{tgtsrcred}45.85 & \cellcolor{tgtsrcred}31.07 \\
 & \cellcolor{bmsrcgreen}BM $\rightarrow$ SK & \cellcolor{bmsrcgreen}9 & \cellcolor{bmsrcgreen}57.88 & \cellcolor{bmsrcgreen}\textbf{40.05} & \cellcolor{bmsrcgreen}47.23 & \cellcolor{bmsrcgreen}\textbf{32.07} \\
 & \cellcolor{tgtsrcred}SK $\rightarrow$ BM & \cellcolor{tgtsrcred}9 & \cellcolor{tgtsrcred}55.28 & \cellcolor{tgtsrcred}35.54 & \cellcolor{tgtsrcred}27.89 & \cellcolor{tgtsrcred}12.29 \\
\bottomrule
\end{tabular}%
}%
\end{table}

\subsection{Boundary-Sensitive Semantic Segmentation}
Semantic segmentation of LiDAR point clouds is particularly sensitive to object boundaries where geometric and radiometric cues become indistinct \cite{hu2021semantic3d}. In dense urban scenes, overhead \textit{Wires} often overlap with nearby \textit{Wire-Structure Connectors}, and adjacent \textit{Buildings} frequently share walls or narrow gaps (see \supfig{dataset-example}).

While well-represented classes like \textit{Vehicle} (92.87\%) and \textit{Road} (87.94\%) achieve strong IoU even near cluttered boundaries, thin and geometrically ambiguous structures remain challenging. \textit{Fencing} achieves only 38.79\% IoU at best, as its narrow planar geometry is easily confused with nearby walls, \textit{Boards}, and \textit{Buildings} (47--65\% of \textit{Fencing} points are predicted as \textit{Building}; \supfig{confusion-matrices}). Similarly, \textit{Ground} peaks at 27.02\% IoU due to gradual elevation transitions that blur boundaries with \textit{Road} and \textit{Vegetation}. Fine-scale structural boundaries thus remain open, and this dataset's dense overlap patterns offer a testbed for boundary-aware architectures.

\subsection{Asymmetric Cross-Region Transferability}
Models trained on BhopalMLS transfer to structured benchmarks substantially better than the reverse. To quantify it, we fine-tune Utonia(L)~\cite{zhang2026utonia} and Concerto(L)~\cite{zhang2025concerto} under FT-9ch and evaluate transfer between BhopalMLS and the four datasets of Table~\ref{tab:cross_region_1cm} at native resolution.

The most severe drop is on SK$\leftrightarrow$BM, which differs most in acquisition (SK: vehicle spinning 64-beam, 10.6\,cm spacing vs.\ 4.9\,cm): both Concerto variants fall from above 74\% mIoU on their own domain to 12.3\% and 3.9\% on BhopalMLS, with safety-critical classes near zero (\suptab{cross_city_full}). Because this single pair conflates sensor, field-of-view, and density, we do not rest the asymmetry claim on it.

Conversely, BhopalMLS-trained models transfer out well: the BM-source direction wins on \emph{all four} BM$\leftrightarrow$target pairs for both architectures ($+$4.5 to $+$36.1 mIoU). This resists both confounds: it holds intra-Indian against a \emph{different} sensor (KM uses a vehicle Riegl VQ-450), ruling out the ZEB pattern; and for BM$\leftrightarrow$PL, where Paris-Lille-3D is \emph{denser} than BhopalMLS (3.8 vs.\ 4.9\,cm, Table~\ref{tab:dataset_complexity}), BM-source still wins by $+$28.6/$+$17.8 mIoU, the opposite of what density-driven degradation predicts. A uniform asymmetry across targets differing in sensor, density and region therefore indicates complementary scene geometry, not an acquisition artifact. A post-hoc CORAL~\cite{sun2016coral} domain-adaptation baseline (\suptab{coral_da}) reinforces this: closed-form second-order feature alignment fails to close the BM$\leftrightarrow$PL gap and even degrades transfer, indicating the gap is structural rather than a global covariance shift.

\makeatletter
\renewcommand\paragraph{\@startsection{paragraph}{4}{\z@}{1ex \@plus .5ex \@minus .2ex}{-1em}{\normalfont\normalsize\bfseries}}
\makeatother
\paragraph{What drives the asymmetry.} We retrain Utonia(L) on four subsets of the training split that share an identical labelled-point budget and each destroy one candidate factor: a random subsample (\emph{size}), every class collapsed to its dominant fine subclass (\emph{variety}), only regular corridors retained (\emph{geometry}), and that same collapse with \textit{Vehicle} keeping \textit{Car} instead of the dominant two-wheeler (\emph{subtype}). Against the size control, mean BM$\rightarrow$X falls 13.0 mIoU points under \emph{variety} and is unchanged under \emph{geometry}; changing only \emph{which} subtype survives then recovers $+$7.5 of that deficit and restores \textit{Vehicle} IoU from 4--14\% to 70--95\%. The advantage is therefore one of \emph{coverage}: BhopalMLS's 15 vehicle subtypes contain the car these benchmarks are built from, alongside the auto-rickshaws and hand-carts they lack. Repeating all four arms for Concerto(L) reproduces both the ranking and this coverage signature, and both contrasts hold on SemanticKITTI too. Protocol, per-arm results and controls are in \supsec{asym_attribution}.

\section{Conclusion}
\textbf{BhopalMLS} is a 1.2-billion-point MLS dataset of a dense, unstructured Indian urban environment. Pretrained foundation models outperform fully supervised baselines, and a cross-region study exposes a consistent \emph{asymmetric transferability}: transfer from BhopalMLS beats the reverse on every pair, holding even where the target is denser. Controlled retraining attributes this to the \emph{coverage} of BhopalMLS's fine-grained subtype inventory rather than to scale, so models trained solely on conventional benchmarks miss safety-critical features in dense, unstructured scenes. BhopalMLS provides a testbed for more geographically diverse 3D perception.

\paragraph{Data, Code, and Access.}
Data, protocols, splits, code and configs are on the project page given in the abstract. The cloud is open in India under the National Geospatial Policy; researchers abroad obtain a derivative withholding only the absolute geographic offset; metrics are unaffected (supp.).

\paragraph{Acknowledgements.}
We thank our annotators, Arjun, Meena and Kavita Meena. We also thank Rownak Tiwari for contributions to an early draft of this work. This work is funded by iHub-Data, IIIT Hyderabad (P. No. IIIT-H/IHub/Project/Data Foundation \textbar2024-25\textbar DF-001) and IISER Bhopal. The work is also supported by Madhya Pradesh State Electronics Development Corporation Ltd. (MPSeDC), Department of Science and Technology (DST), Government of Madhya Pradesh (P. No. \mbox{F 6-1/2022/41-2}).

{
    \small
    \bibliographystyle{ieeenat_fullname}
    \bibliography{main}
}

\clearpage
\appendix
\setlength{\textfloatsep}{20pt plus 2pt minus 4pt}
\makeatletter
\renewcommand\paragraph{\@startsection{paragraph}{4}{\z@}{3.25ex \@plus1ex \@minus.2ex}{-1em}{\normalfont\normalsize\bfseries}}
\makeatother
\renewcommand{\thesection}{S\arabic{section}}
\renewcommand{\theHsection}{S\arabic{section}}
\setcounter{table}{0}
\renewcommand{\thetable}{S\arabic{table}}
\renewcommand{\theHtable}{S\arabic{table}}
\setcounter{figure}{0}
\renewcommand{\thefigure}{S\arabic{figure}}
\renewcommand{\theHfigure}{S\arabic{figure}}
\setcounter{equation}{0}
\renewcommand{\theequation}{S\arabic{equation}}
\renewcommand{\theHequation}{S\arabic{equation}}
\renewcommand{\dbltopfraction}{0.9}
\renewcommand{\dblfloatpagefraction}{0.75}
\twocolumn[{\centering\Large\bfseries Supplementary Material\par\vspace{1.5em}}]

This supplementary material is organised in four parts. \textbf{The dataset}
(\S S1--\S S4): visual characterization of the 62 semantic classes, class and split
statistics, survey-path diversity, the per-point data format, annotation
consistency measured across repeated survey passes, and a quantitative analysis
of spatial overlap between the benchmark splits. \textbf{The benchmark}
(\S S5--\S S8): extended fine-grained (21-class) semantic segmentation results, a
class-imbalance ablation, instance segmentation baselines with qualitative
visualizations, and foundation-model feature visualizations, prediction error
maps and confusion-matrix analysis. \textbf{Cross-region transfer}
(\S S9--\S S11): per-class breakdowns behind Table~\ref{tab:cross_region_1cm}, a factor-attribution
study that identifies what drives the asymmetric transferability, and a post-hoc
CORAL domain-adaptation analysis. \textbf{Ethics} (\S S12): the collection
context, consent and institutional review, and the privacy and release
protocol.

\section{Dataset Visualizations and Statistics}

This section provides a visual characterization of the BhopalMLS dataset: the acquisition setup, class-wise point cloud and RGB samples, annotation examples, and the full class and train/validation/test split distribution statistics.

\subsection{Acquisition Setup}

\begin{figure*}[!tp]
    \centering
    \includegraphics[width=\textwidth]{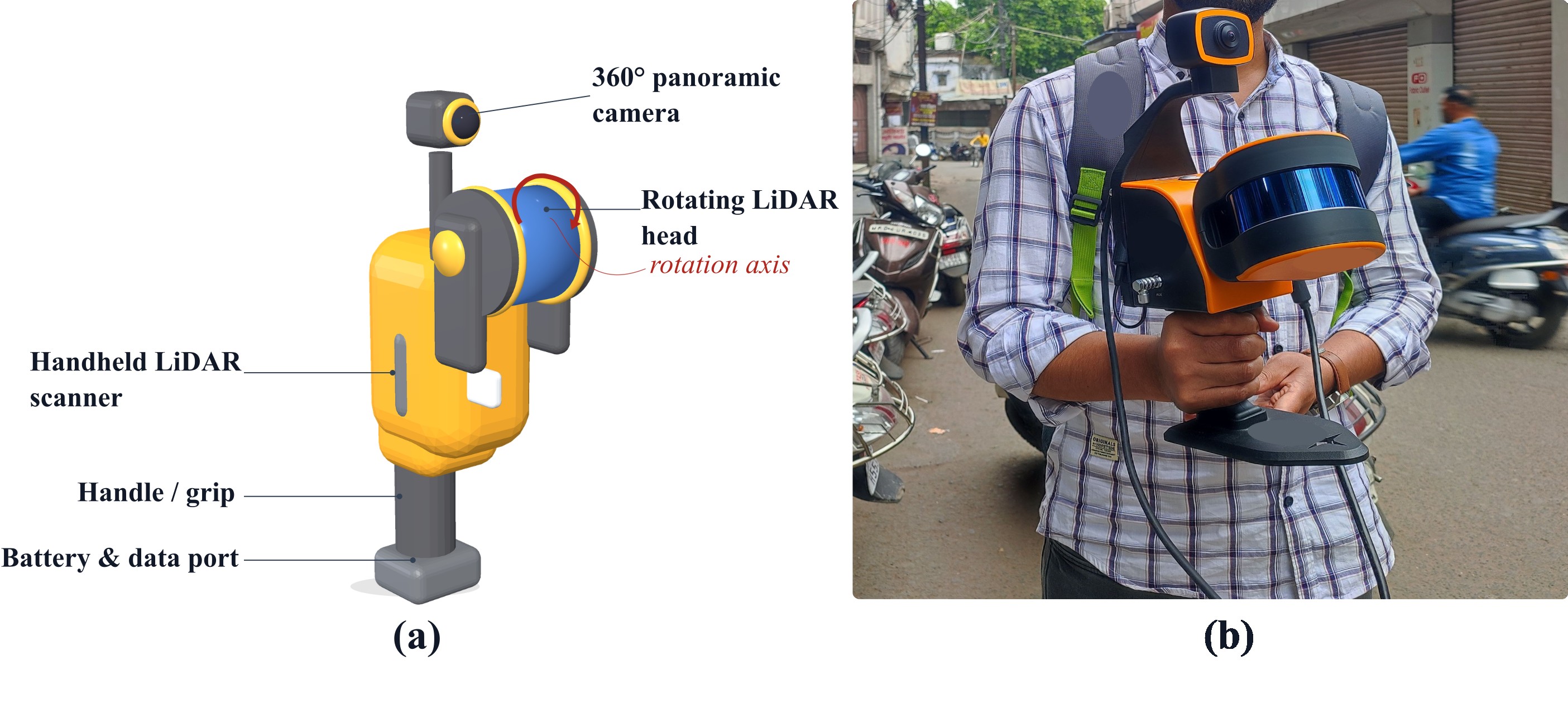}
    \caption{\textbf{Handheld scanning acquisition.} \textbf{(a)}~Schematic of the handheld LiDAR scanner, showing the rotating LiDAR head (with its rotation axis), the co-mounted 360$^\circ$ panoramic camera used for colorization, the handle/grip, and the battery and data port. \textbf{(b)}~The system in use along a dense Bhopal street corridor; the handheld form factor enables data collection in narrow, congested lanes that are inaccessible to vehicle-mounted scanners.}
    \label{S-fig:acquisition}
\end{figure*}

BhopalMLS was captured with a handheld GeoSLAM ZEB Horizon LiDAR scanner (a Velodyne VLP-16 mounted on a continuously rotating head), with a 360$^\circ$ panoramic camera co-mounted for colorization. Figure~\ref{S-fig:acquisition} shows \textbf{(a)} the scanner schematic and \textbf{(b)} the system in use along a dense Bhopal corridor. The handheld form factor enables coverage of narrow, congested streets that are inaccessible to vehicle-mounted platforms.

\subsection{Class-wise Point Cloud Samples}

\begin{figure*}[!tp]
    \centering
    \includegraphics[width=\textwidth, height=0.90\textheight, keepaspectratio]{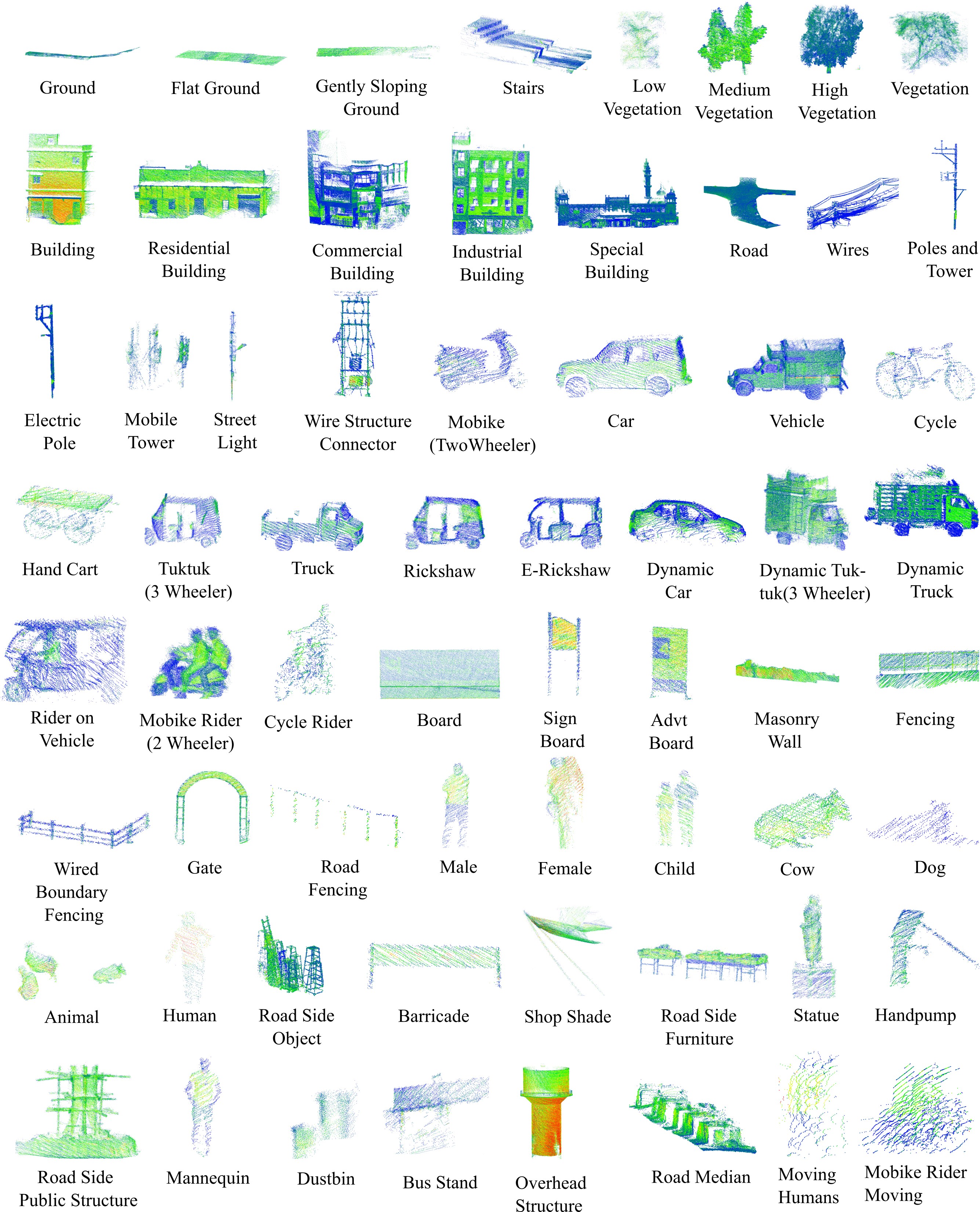}
    \caption{Intensity-encoded point cloud samples (blue = low intensity, red = high intensity) for each of the 62 static classes, plus two moving-class examples (\textit{Moving Humans} and \textit{Mobike Rider Moving}) showing the motion-induced point geometry of dynamic objects. Visually ambiguous classes are grouped under broader general categories for clarity. Corresponding RGB imagery is shown in Fig.~\ref{S-fig:class-samples-grid-rgb}.}
    \label{S-fig:class-samples-grid}
\end{figure*}

Figure~\ref{S-fig:class-samples-grid} presents intensity-encoded point cloud samples for all 62 semantic categories, together with two moving-class examples (\textit{Moving Humans} and \textit{Mobike Rider Moving}) that show the motion-induced point tails of dynamic objects. These samples illustrate the geometric diversity, scale variation, and structural complexity present across the dataset. Classes that are visually ambiguous or difficult to distinguish in isolation are grouped under broader general categories (\eg, general Building, general Board, general Vehicle) for clarity.

\subsection{Class-wise RGB Samples}

\begin{figure*}[!tp]
    \centering
    \includegraphics[width=\textwidth, height=0.90\textheight, keepaspectratio, trim=36bp 0 0 0, clip]{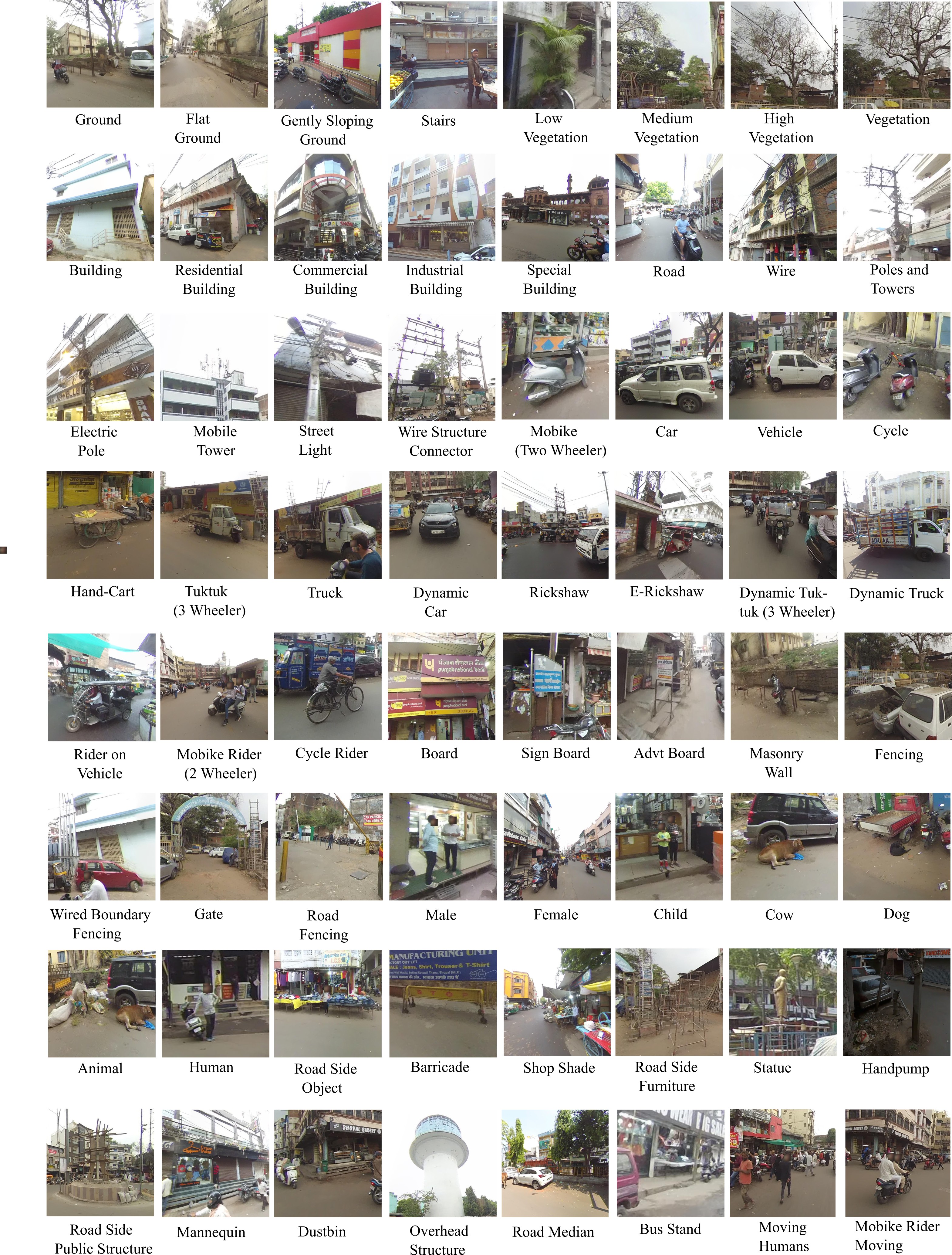}%
    \caption{Class-wise RGB images corresponding to Fig.~\ref{S-fig:class-samples-grid}, illustrating the 62 static semantic classes plus two moving-class examples (\textit{Moving Humans} and \textit{Mobike Rider Moving}) in the dataset.}
    \label{S-fig:class-samples-grid-rgb}
\end{figure*}

Figure~\ref{S-fig:class-samples-grid-rgb} provides the corresponding RGB images for all 62 semantic classes and the two moving-class examples (\textit{Moving Humans} and \textit{Mobike Rider Moving}), complementing the LiDAR point cloud views in Fig.~\ref{S-fig:class-samples-grid}. These samples demonstrate appearance-level variations across urban, roadside, vegetation, and object-level categories, offering visual context to the LiDAR-only views.

\subsection{LiDAR Annotation Examples}

\begin{figure*}[!tp]
    \centering
    \includegraphics[width=\textwidth, height=0.90\textheight, keepaspectratio]{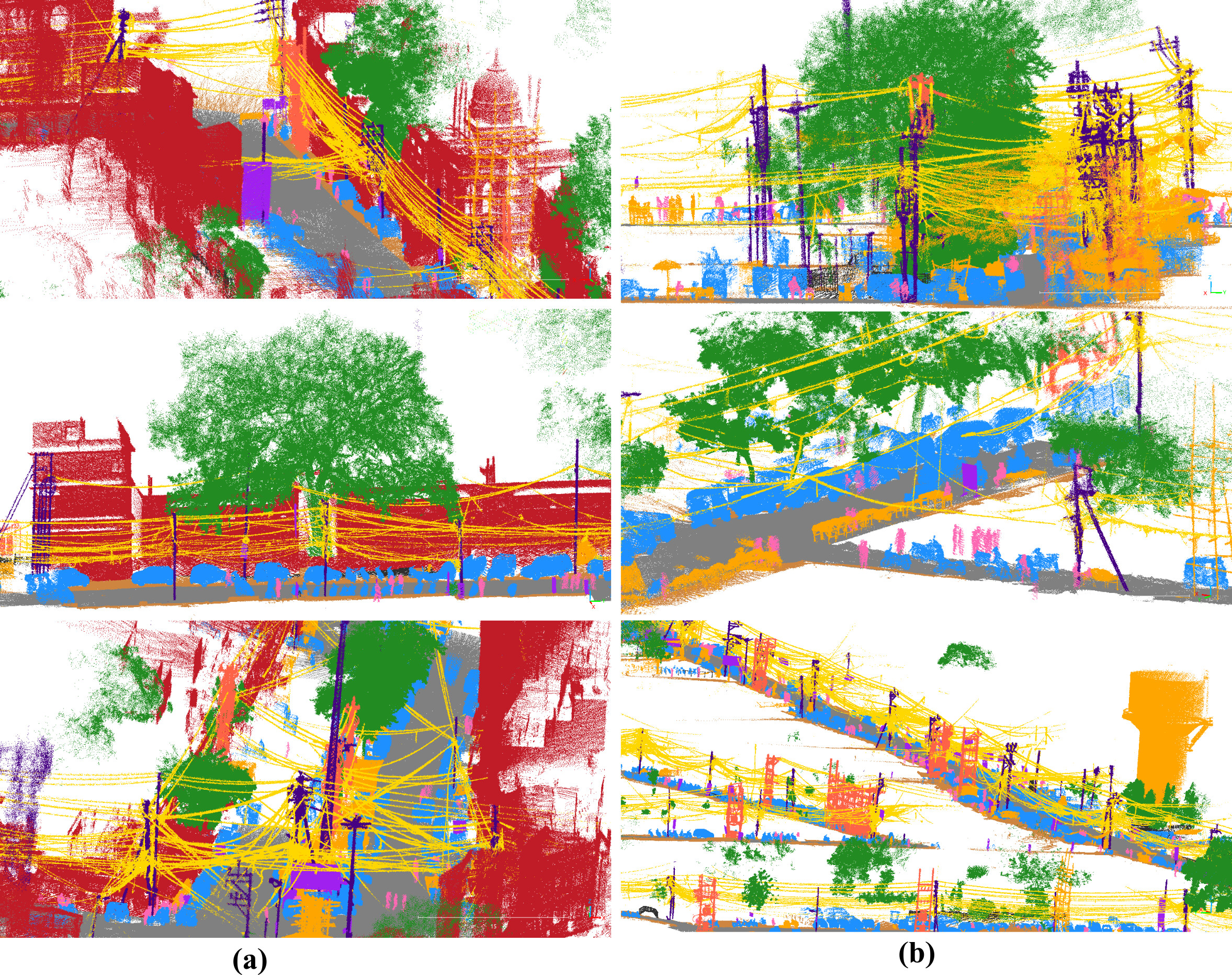}
    \caption{\textbf{BhopalMLS LiDAR annotation examples.} \textbf{(a)} Different views of annotations showing the complexity of the dense urban environment. \textbf{(b)} Overview of the larger study area with annotations \textbf{excluding buildings} for better visualization of multiple features present.}
    \label{S-fig:lidar-scenes-overview}
\end{figure*}

Figure~\ref{S-fig:lidar-scenes-overview} presents BhopalMLS LiDAR annotation examples that illustrate the spatial richness and annotation density of the survey area. Panel \textbf{(a)} shows multiple views of annotated scenes to highlight the complexity of dense urban environments, occlusions, and fine-grained object boundaries. Panel \textbf{(b)} presents an overview of a larger portion of the study region with annotations rendered excluding buildings, enabling clearer visualization of smaller structures, linear objects, road furniture, and other subtle features.

\begin{figure*}[!tp]
  \centering
  \includegraphics[width=0.9\textwidth]{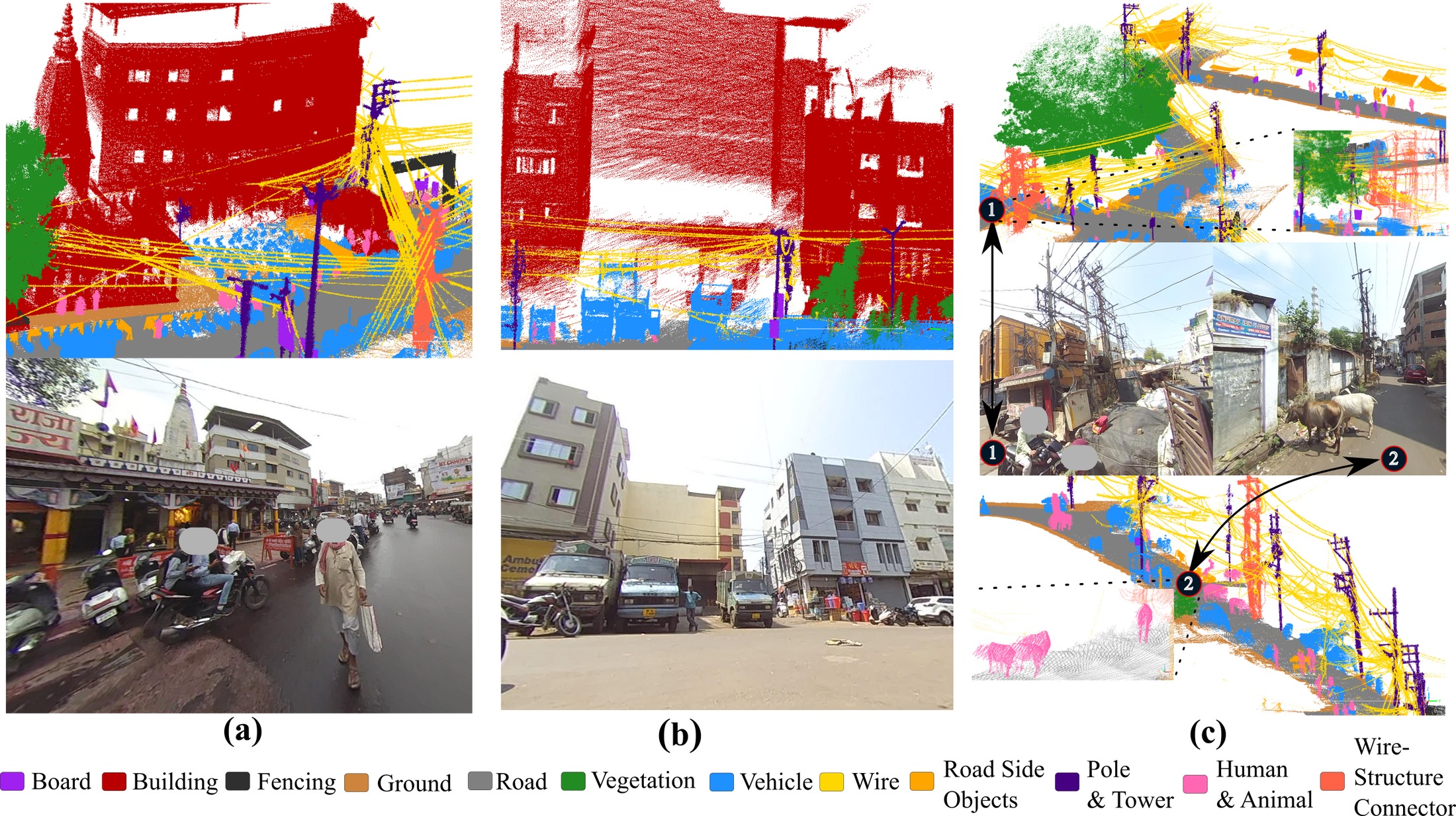}
\caption{\textbf{Annotation examples with corresponding RGB images.} \textbf{(a)} and \textbf{(b)} show annotated point clouds (top row) and corresponding RGB images (bottom row), illustrating varied architectural styles and dense urban features. RGB images show approximate viewpoints, not exact point cloud perspectives. \textbf{(c)} Top view of annotations with buildings removed for clarity; viewpoints (1) and (2) show zoomed RGB images and corresponding point clouds.}
  \label{S-fig:dataset-example}
\end{figure*}

Figure~\ref{S-fig:dataset-example} further pairs annotated point clouds with their corresponding 360$^\circ$ RGB imagery, illustrating varied architectural styles and dense urban features (RGB views are approximate, not exact point-cloud perspectives).

\subsection{Dataset Statistics}
\begin{figure*}[!tp]
\centering
\includegraphics[width=\textwidth]{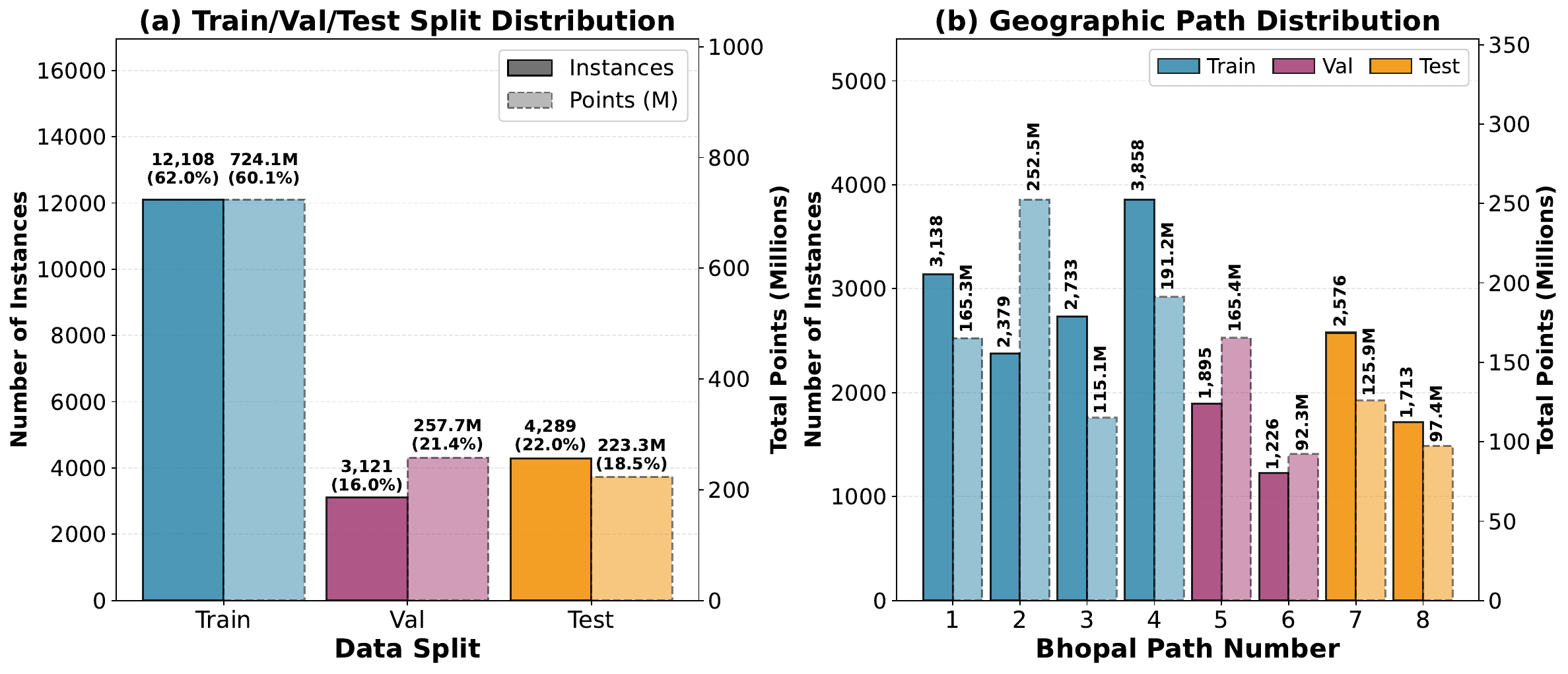}
\caption{%
\textbf{Dataset split and geographic path distribution.}
\textbf{(a)} Train/Val/Test split by instance count and point count (Millions), with percentages; the split follows an approximate 60/21/19 point ratio (Sec.~4.1).
\textbf{(b)} Per-path distribution (Paths 1--8) of instances and points, colored by split (Train: Paths 1--4; Val: 5--6; Test: 7--8).
In both panels, solid bars are instance counts and dashed bars are point counts (Millions); colors denote the split.
This path-based partitioning assigns entire geographic paths to splits, preserving spatial coherence while keeping the split ratio balanced.
}
\label{S-fig:split_distribution}
\end{figure*}

Table~\ref{S-tab:class_distribution_full} gives the complete 62-subclass breakdown summarized by the 12 major classes in Table~\ref{tab:class_distribution}, including the 6 moving-class subclasses (marked $^{\dagger}$, shaded \colorbox{red!12}{light red}). Figure~\ref{S-fig:split_distribution} shows the class-wise train/validation/test split distribution (Sec.~4.1).

\begin{table*}[!tp]
\centering
\small
\caption{%
\textbf{BhopalMLS full semantic class distribution.}
The primary benchmark uses \textbf{62 static} semantic classes (19{,}518 instances, 1{,}205.2M points) in 12 major categories. \textbf{6 moving-class subclasses} (marked $^{\dagger}$, shaded \colorbox{red!12}{light red}; 7{,}181 instances, 26.2M points) extend the taxonomy to \textbf{68 classes}; at evaluation they fold into their static parents (\textit{Vehicle}, \textit{Human \& Animal}). Subclasses are sorted by point contribution within each major class.
}
\label{S-tab:class_distribution_full}
\setlength{\tabcolsep}{2pt}
\resizebox{\textwidth}{!}{%
\begin{tabular}{@{}llrrr@{\hspace{0.5em}}llrrr@{}}
\toprule
\multicolumn{5}{c@{\hspace{0.5em}}}{\textbf{Major Classes (Part 1)}} & \multicolumn{5}{c}{\textbf{Major Classes (Part 2)}} \\
\cmidrule(r){1-5}\cmidrule(l){6-10}
\textbf{Major Class} & \textbf{Subclass} & \textbf{Inst.} & \textbf{Pts(M)} & \textbf{Pts(\%)} & \textbf{Major Class} & \textbf{Subclass} & \textbf{Inst.} & \textbf{Pts(M)} & \textbf{Pts(\%)} \\
\midrule
\textbf{Building} & \textit{Total} & \textbf{1,326} & \textbf{813.0} & \textbf{66.0} & \textbf{Human \& Animal} & \textit{Total} & \textbf{8,864} & \textbf{22.5} & \textbf{1.8} \\
 & {\scriptsize |-- Commercial} & 724 & 628.8 & 51.1 &  & \cellcolor{red!12}{\scriptsize |-- Moving Humans$^{\dagger}$} & \cellcolor{red!12}5,386 & \cellcolor{red!12}12.6 & \cellcolor{red!12}1.0 \\
 & {\scriptsize |-- Residential} & 106 & 84.0 & 6.8 &  & {\scriptsize |-- Human \& Animal} & 2,005 & 4.9 & 0.4 \\
 & {\scriptsize |-- Building (gen.)} & 439 & 52.6 & 4.3 &  & {\scriptsize |-- Male} & 996 & 3.2 & 0.3 \\
 & {\scriptsize |-- Special} & 56 & 47.2 & 3.8 &  & {\scriptsize |-- Female} & 294 & 1.1 & 0.1 \\
 & {\scriptsize |-- Industrial} & 1 & 0.5 & 0.0 &  & {\scriptsize |-- Cow} & 32 & 0.2 & 0.0 \\
& & & & &  & \cellcolor{red!12}{\scriptsize |-- Moving Female$^{\dagger}$} & \cellcolor{red!12}40 & \cellcolor{red!12}0.2 & \cellcolor{red!12}0.0 \\
& & & & &  & \cellcolor{red!12}{\scriptsize |-- Moving Male$^{\dagger}$} & \cellcolor{red!12}57 & \cellcolor{red!12}0.2 & \cellcolor{red!12}0.0 \\
& & & & &  & {\scriptsize |-- Child} & 49 & 0.1 & 0.0 \\
& & & & &  & {\scriptsize |-- Animal} & 1 & 0.0 & 0.0 \\
& & & & &  & \cellcolor{red!12}{\scriptsize |-- Moving Animal$^{\dagger}$} & \cellcolor{red!12}1 & \cellcolor{red!12}0.0 & \cellcolor{red!12}0.0 \\
& & & & &  & {\scriptsize |-- Dog} & 3 & 0.0 & 0.0 \\
\addlinespace[0.5em]
\textbf{Road} & \textit{Single} & \textbf{289} & \textbf{157.0} & \textbf{12.8} & \textbf{Pole \& Tower} & \textit{Total} & \textbf{1,096} & \textbf{11.5} & \textbf{0.9} \\
& & & & &  & {\scriptsize |-- Electric Pole} & 615 & 10.3 & 0.8 \\
& & & & &  & {\scriptsize |-- Pole \& Tower} & 178 & 0.6 & 0.0 \\
& & & & &  & {\scriptsize |-- Street Light} & 289 & 0.4 & 0.0 \\
& & & & &  & {\scriptsize |-- Mobile Tower} & 14 & 0.2 & 0.0 \\
\addlinespace[0.5em]
\textbf{Vehicle} & \textit{Total} & \textbf{9,888} & \textbf{77.2} & \textbf{6.3} & \textbf{Board} & \textit{Total} & \textbf{871} & \textbf{7.9} & \textbf{0.6} \\
 & {\scriptsize |-- Mobike (Two-Wheeler)} & 6,335 & 35.0 & 2.8 &  & {\scriptsize |-- Advt Board} & 484 & 5.4 & 0.4 \\
 & {\scriptsize |-- Car} & 544 & 11.5 & 0.9 &  & {\scriptsize |-- Board (gen.)} & 370 & 2.4 & 0.2 \\
 & \cellcolor{red!12}{\scriptsize |-- Mobike Rider Moving$^{\dagger}$} & \cellcolor{red!12}1,544 & \cellcolor{red!12}10.7 & \cellcolor{red!12}0.9 &  & {\scriptsize |-- Sign Board} & 17 & 0.1 & 0.0 \\
 & {\scriptsize |-- Hand Cart} & 347 & 4.2 & 0.3 & \textbf{Vehicle} {\scriptsize(cont.)} & {\scriptsize |-- Dynamic Car} & 14 & 0.3 & 0.0 \\
 & {\scriptsize |-- Tuktuk (Three-Wheeler)} & 214 & 4.0 & 0.3 &  & {\scriptsize |-- Rickshaw} & 15 & 0.3 & 0.0 \\
 & {\scriptsize |-- Mobike Rider} & 366 & 3.6 & 0.3 &  & {\scriptsize |-- Rider on Vehicle} & 16 & 0.3 & 0.0 \\
 & \cellcolor{red!12}{\scriptsize |-- Moving Vehicle$^{\dagger}$} & \cellcolor{red!12}153 & \cellcolor{red!12}2.6 & \cellcolor{red!12}0.2 &  & {\scriptsize |-- Cycle} & 102 & 0.2 & 0.0 \\
 & {\scriptsize |-- Vehicle (gen.)} & 151 & 1.7 & 0.1 &  & {\scriptsize |-- Dynamic Truck} & 1 & 0.1 & 0.0 \\
 & {\scriptsize |-- Truck} & 31 & 1.4 & 0.1 &  & {\scriptsize |-- Cycle Rider} & 9 & 0.1 & 0.0 \\
 & {\scriptsize |-- Dynamic Tuktuk} & 44 & 1.2 & 0.1 &  & {\scriptsize |-- E-Rickshaw} & 2 & 0.0 & 0.0 \\
\addlinespace[0.5em]
\textbf{Wire} & \textit{Single} & \textbf{522} & \textbf{39.0} & \textbf{3.2} & \textbf{Wire-Structure Conn.} & \textit{Single} & \textbf{80} & \textbf{6.9} & \textbf{0.6} \\
\addlinespace[0.5em]
\textbf{Vegetation} & \textit{Total} & \textbf{374} & \textbf{33.6} & \textbf{2.7} & \textbf{Fencing} & \textit{Total} & \textbf{127} & \textbf{3.0} & \textbf{0.2} \\
 & {\scriptsize |-- High Vegetation} & 120 & 29.1 & 2.4 &  & {\scriptsize |-- Masonry Wall} & 18 & 1.1 & 0.1 \\
 & {\scriptsize |-- Vegetation (gen.)} & 91 & 2.1 & 0.2 &  & {\scriptsize |-- Fencing (gen.)} & 84 & 1.1 & 0.1 \\
 & {\scriptsize |-- Medium Vegetation} & 36 & 1.6 & 0.1 &  & {\scriptsize |-- Gate of Boundary} & 21 & 0.7 & 0.1 \\
 & {\scriptsize |-- Low Vegetation} & 127 & 0.8 & 0.1 &  & {\scriptsize |-- Wired Boundary Fencing} & 3 & 0.0 & 0.0 \\
 & & & & &  & {\scriptsize |-- Road Fencing} & 1 & 0.0 & 0.0 \\
\addlinespace[0.5em]
\textbf{Road Side Object} & \textit{Total} & \textbf{2,294} & \textbf{30.4} & \textbf{2.5} & \textbf{Ground} & \textit{Total} & \textbf{968} & \textbf{29.4} & \textbf{2.4} \\
 & {\scriptsize |-- Shop Shade} & 315 & 15.1 & 1.2 &  & {\scriptsize |-- Stairs} & 310 & 10.5 & 0.8 \\
 & {\scriptsize |-- Road Side Furniture} & 1,455 & 11.8 & 1.0 &  & {\scriptsize |-- Flat Ground} & 322 & 9.9 & 0.8 \\
 & {\scriptsize |-- Mannequin} & 365 & 1.6 & 0.1 &  & {\scriptsize |-- Ground (gen.)} & 310 & 8.1 & 0.7 \\
 & {\scriptsize |-- Overhead Structure} & 4 & 0.8 & 0.1 &  & {\scriptsize |-- Gently Sloping} & 26 & 0.9 & 0.1 \\
 & {\scriptsize |-- Road Side Object (gen.)} & 54 & 0.4 & 0.0 & \textbf{R.S. Object} {\scriptsize(cont.)} & {\scriptsize |-- Handpump} & 17 & 0.0 & 0.0 \\
 & {\scriptsize |-- Barricade} & 68 & 0.4 & 0.0 &  & {\scriptsize |-- Dustbin} & 5 & 0.0 & 0.0 \\
 & {\scriptsize |-- Road Median} & 6 & 0.2 & 0.0 &  & {\scriptsize |-- Statue} & 1 & 0.0 & 0.0 \\
 & {\scriptsize |-- Public Structure} & 3 & 0.1 & 0.0 &  & {\scriptsize |-- Bus Stand} & 1 & 0.0 & 0.0 \\
\midrule
\multicolumn{10}{l}{\textbf{Primary benchmark (62 static classes)}: \textbf{19,518 instances}, \textbf{1,205.2M points}} \\
\multicolumn{10}{l}{\colorbox{red!12}{\textbf{Moving-class subclasses (6 classes, $^{\dagger}$)}: \textbf{7,181 instances}, \textbf{26.2M points}}} \\
\multicolumn{10}{l}{\textbf{Combined (68 subclasses)}: \textbf{26,699 instances}, \textbf{1,231.4M points} (100.0\%)} \\
\bottomrule
\end{tabular}
}%
\end{table*}

\subsection{Survey Path Characteristics and Diversity}
\label{S-sec:path_diversity}

\begin{figure*}[!tp]
  \centering
  \includegraphics[width=0.82\textwidth]{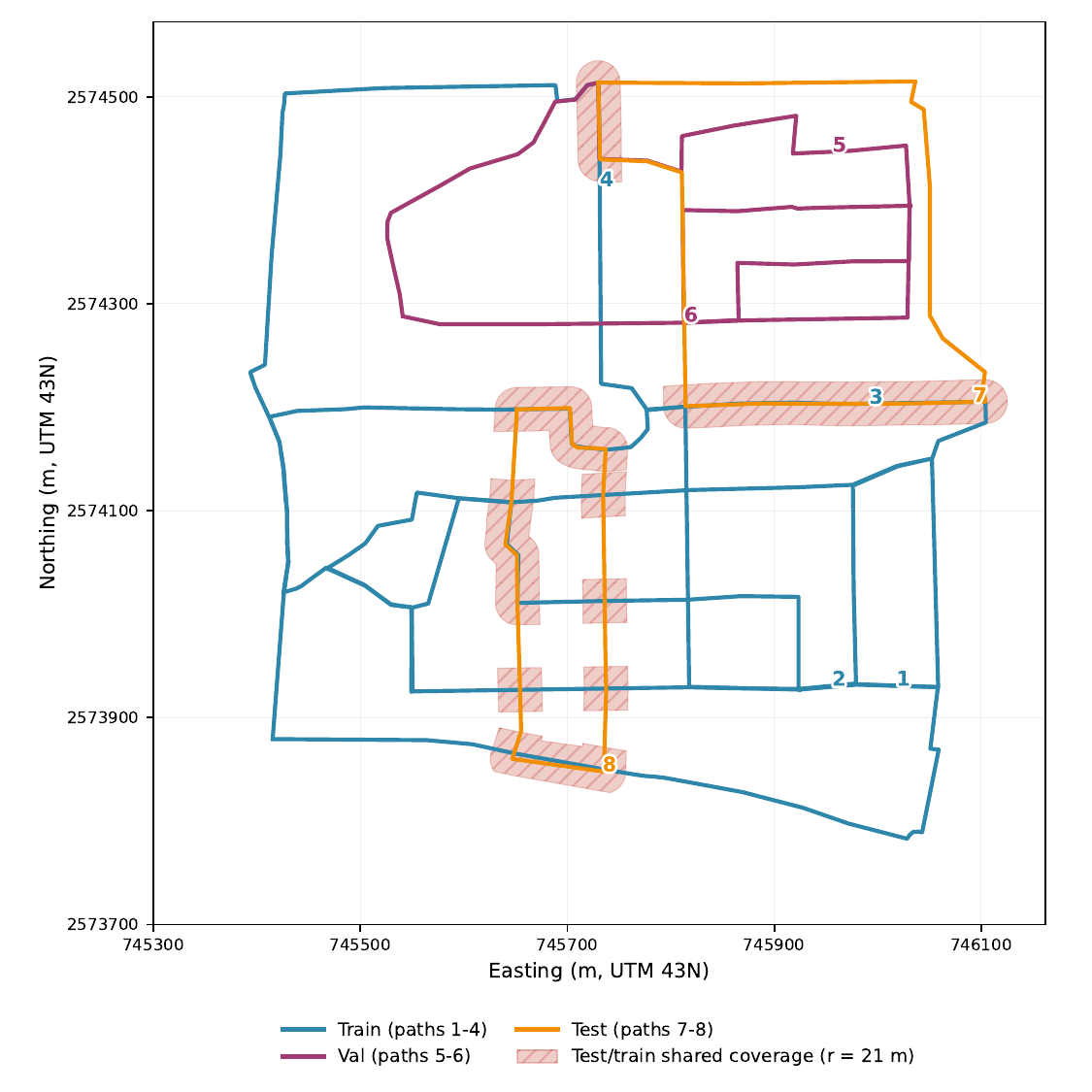}
  \caption{\textbf{Survey path layout and split assignment.} The eight
  handheld survey paths in UTM 43N, colored by benchmark split, with shaded
  zones marking where test coverage overlaps training coverage at the
  sensor's effective reach (21\,m).}
  \label{S-fig:path_split_map}
\end{figure*}

The eight survey paths are separate passes over the district's road
network, not repeated traversals of one route. Table~\ref{S-tab:path_diversity}
summarizes their geometry and composition: lengths range from 657 to
1{,}846\,m, mean carriageway widths from 2.6 to 3.5\,m, cumulative turn rates
from 46 to 83$^\circ$ per 100\,m, and building point shares from 57.5\% to
74.3\%. The validation paths (5 and 6) traverse the narrowest, most
building-enclosed lanes of the old market core (widths 2.6--2.7\,m, building
share above 72\%), whereas the training and test paths cover wider mixed
corridors with up to 6.5\% vegetation. Figure~\ref{S-fig:path_split_map} shows
the spatial layout of all paths with their split assignment, including where
test paths pass near training coverage. That overlap is quantified, and its
effect on the benchmark measured, in \S\ref{S-sec:leakage}.

\begin{table}[!tb]
\centering
\footnotesize
\setlength{\tabcolsep}{3pt}
\caption{\textbf{Survey path characteristics.} Length, benchmark scenes and
points, cumulative turn rate, mean carriageway width (5th--95th percentile
extent of road points per scene), and dominant class shares per path.}
\label{S-tab:path_diversity}
\resizebox{\columnwidth}{!}{%
\begin{tabular}{lcrrrrrrrrr}
\toprule
\textbf{Path} & \textbf{Split} & \textbf{Len.\,(m)} & \textbf{Scenes} & \textbf{Pts\,(M)} & \textbf{Turn\,($^\circ$/100m)} & \textbf{Width\,(m)} & \textbf{Bld.\,\%} & \textbf{Road\,\%} & \textbf{Veg.\,\%} & \textbf{Veh.\,\%} \\
\midrule
1 & train & 1{,}663 & 418 & 137.2 & 56.1 & 3.39 & 57.5 & 16.6 & 5.1 & 5.9 \\
2 & train & 1{,}846 & 270 & 147.4 & 82.5 & 3.47 & 68.8 & 13.9 & 1.0 & 6.9 \\
3 & train & 1{,}110 & 331 &  98.9 & 46.2 & 3.20 & 61.8 & 14.9 & 5.2 & 7.4 \\
4 & train & 1{,}845 & 334 & 156.8 & 82.3 & 3.32 & 60.6 & 15.2 & 3.9 & 6.2 \\
5 & val   & 1{,}452 & 380 & 143.2 & 74.4 & 2.69 & 72.9 & 11.2 & 1.0 & 3.2 \\
6 & val   &   657 & 188 &  83.4 & 77.5 & 2.59 & 74.3 & 10.6 & 0.1 & 4.2 \\
7 & test  & 1{,}318 & 285 & 111.4 & 55.1 & 3.53 & 58.5 & 14.7 & 6.5 & 7.0 \\
8 & test  &   865 & 204 &  90.4 & 68.9 & 3.30 & 69.6 & 12.5 & 0.8 & 4.0 \\
\bottomrule
\end{tabular}%
}
\end{table}

\section{Per-Point Data Format}
\label{S-sec:format}
The dataset is provided in multiple formats: LAS for GIS compatibility, PLY for 3D visualization, HDF5 for deep learning with optimized batched loading, and CSV, NPZ, BIN for general-purpose access. Each point includes the following attributes:
\begin{enumerate}
    \item \textbf{X, Y, Z:} 3D coordinates in the georeferenced coordinate system (float64 storage precision; positional accuracy is set by the SLAM survey and its DGPS alignment, Sec.~3.2, not by the storage format).
    \item \textbf{RGB:} Per-point color from co-registered 360$^\circ$ panoramic images (uint8 per channel); valid for approximately 40\% of points (limited by image registration in poor-GPS lanes), zero elsewhere.
    \item \textbf{Intensity:} LiDAR return intensity (uint16, range 0--65535).
    \item \textbf{GPS Time:} Timestamp for temporal referencing (float64, seconds).
    \item \textbf{Semantic Labels:} 62 static semantic classes (uint16); 6 additional moving-class subclasses provide an optional fine-grained dynamic-object labeling (68 classes total).
    \item \textbf{Instance or Global ID:} Unique 9-digit identifier encoding path, class hierarchy, and instance number (NXXZZYYYY format: N=path 1--8, XX=major class 01--12, ZZ=subclass, YYYY=instance number).
\end{enumerate}

\section{Annotation Consistency Across Survey Passes}
\label{S-sec:agreement}

The main paper reports that annotators iterated until point-level disagreement
fell below 0.5\% per path. That figure is an aggregate, and aggregates over a
long-tailed dataset are dominated by the largest classes. This section reports
consistency per class.

\paragraph{Method.} Where two survey paths cover the same street, the same
physical surfaces were annotated twice, in independent work units; comparing
the labels assigned at the same location measures annotation consistency. The
two passes are separate SLAM trajectories and register to within 8--25\,cm, so
a naive nearest-neighbour comparison would report misalignment as
disagreement. We therefore (i) estimate and remove a per-pair rigid
translation, (ii) compare only \emph{core} points, whose $k{=}8$ nearest
neighbours in their own pass share their label, and require the matched
neighbourhood in the other pass to be homogeneous as well, and (iii) report
movable classes separately, since vehicles and people genuinely move between
passes. Agreement is accumulated over the 62 fine classes and folded into the
coarser taxonomies, so within-parent disagreements are counted correctly. Over
200 co-located scene pairs this compares 2.1M points.

\paragraph{Results.} Overall agreement is 93.4\% in the 12-class taxonomy
(94.3\% excluding movable classes), 82.2\% at 21 classes and 76.8\% over the
62 fine classes. It is far from uniform (Table~\ref{S-tab:agreement}):
\textit{Building} and \textit{Road} carry 85\% of the compared points and both
exceed 94\%, so an aggregate threshold is effectively a statement about those
two classes and can be met while rarer classes are labelled far less
consistently. Two classes are genuinely ambiguous rather than mislabelled:
\textit{Ground} (38.0\%, confused with \textit{Road} where unpaved surfaces
grade into the carriageway) and \textit{Fencing} (64.3\%, confused with
\textit{Building} on boundary walls that are functionally both). These are the
same confusions our models make (Fig.~\ref{S-fig:confusion-matrices}).

\paragraph{Implication for the benchmark.} Per-class agreement correlates
strongly with per-class model IoU (Spearman $\rho{=}0.77$, $p{=}0.009$, over
the 10 static classes): the classes on which humans agree least are those on
which models score lowest. A substantial part of the residual error on
\textit{Ground} and \textit{Fencing} is therefore label ambiguity intrinsic to
unstructured streets rather than model failure, and these classes should be
read as having a lower attainable ceiling. The same effect explains why
agreement drops at finer granularity: the 21-class taxonomy splits
\textit{Building} into \textit{Commercial} and \textit{Building (other)}, a
distinction on which the two passes frequently disagree.

\paragraph{Limitations.} This is a proxy for inter-annotator agreement, not a
direct measurement: it cannot detect an error both annotators made identically,
it measures object interiors rather than boundary delineation (boundaries are
excluded to survive registration drift), and movable classes are not
interpretable as annotation error.

\begin{table}[t]
\centering
\footnotesize
\setlength{\tabcolsep}{4pt}
\caption{\textbf{Cross-pass label agreement per class} (12-class taxonomy,
200 co-located scene pairs, 2.1M points compared). \emph{Share} is the
fraction of compared points. Movable classes ($^{\dagger}$) differ between
passes for physical reasons and are excluded from the aggregate. Model IoU is
Utonia(L)~FT from Table~\ref{tab:bhopal_iou_benchmark_style_rotated_vline}.}
\label{S-tab:agreement}
\begin{tabular}{lrrr}
\toprule
\textbf{Class} & \textbf{Share (\%)} & \textbf{Agreement (\%)} & \textbf{Model IoU} \\
\midrule
Wire-Structure Conn. & 2.8 & 99.8 & 83.0 \\
Building & 64.0 & 96.4 & 93.4 \\
Vegetation & 1.3 & 94.6 & 97.4 \\
Road & 21.3 & 94.1 & 85.7 \\
Wire & 2.1 & 92.2 & 87.5 \\
Road Side Object & 2.1 & 88.1 & 51.5 \\
Board & 0.5 & 77.6 & 67.4 \\
Fencing & 0.2 & 64.3 & 38.8 \\
Pole and Tower & 0.5 & 54.5 & 68.6 \\
Ground & 2.2 & 38.0 & 26.3 \\
\midrule
Vehicle$^{\dagger}$ & 2.6 & 83.6 & 91.5 \\
Human \& Animal$^{\dagger}$ & 0.3 & 4.7 & 64.2 \\
\bottomrule
\end{tabular}
\end{table}

\section{Spatial Overlap Between Splits}
\label{S-sec:leakage}

BhopalMLS is split by survey path (paths 1--4 train, 5--6 val, 7--8 test), but
paths in a compact 10\,km district inevitably run near one another. Lilja
\etal~\cite{lilja2024localization} show that when splits of online-mapping
datasets overlap geographically, evaluation of static classes such as buildings
and road topology degenerates towards localisation. We therefore quantify the
overlap in our splits and, more importantly, measure whether it inflates the
benchmark.

\subsection{How much overlap there is}

Table~\ref{S-tab:leakage_overlap} reports proximity between split trajectories and
the share of each split's observed area that is also observed in training. The
overlap is real and concentrated in the test split: \textbf{35.0\% of the test
trajectory lies within 5\,m of a training trajectory} against 6.7\% for val, and
median test-to-train distance is 22.7\,m against 89.7\,m for val. Measured by
area rather than trajectory, \textbf{49.8\% of the region the test set observes
is also observed during training} (val 15.7\%), where ``observes'' uses the
sensor's median effective half-extent of 21.2\,m.
Figure~\ref{S-fig:path_split_map} shows where.

\begin{table}[!tb]
\centering
\footnotesize
\setlength{\tabcolsep}{4pt}
\caption{\textbf{Spatial overlap of the benchmark splits.} Trajectory proximity
is the share of a split's trajectory samples within the stated distance of any
training trajectory; area overlap is the share of the split's observed footprint
(sensor half-extent 21.2\,m) intersecting the training footprint. The test split
overlaps training substantially more than val does.}
\label{S-tab:leakage_overlap}
\begin{tabular}{lcc}
\toprule
& \textbf{val} & \textbf{test} \\
\midrule
within 5\,m of train      & \phantom{0}6.7\% & \textbf{35.0\%} \\
within 10\,m of train     & \phantom{0}7.6\% & 39.7\% \\
within 25\,m of train     & 10.4\% & 51.8\% \\
median distance to train  & 89.7\,m & 22.7\,m \\
\midrule
observed-area overlap     & 15.7\% & \textbf{49.8\%} \\
\bottomrule
\end{tabular}
\end{table}

\subsection{Whether it inflates the benchmark}

Overlap matters only if it changes what the benchmark measures. We partition the
test split by whether a scene's axis-aligned bounding box intersects the union
of training-scene boxes, giving a \textbf{leakage-free subset of 180 scenes and
75.7M points that retains all 12 classes}, and an overlapping remainder of 306.
Three path-7 scenes are empty (zero points, hence no bounding box) and fall in
neither subset, which is why the two sum to 486 rather than the split's 489. Every
cached evaluation in our model zoo is then re-scored on both subsets:
\textbf{36 evaluations} in total, 10 at 12 classes and 26 at 21 classes,
spanning supervised baselines and all four adaptation regimes
(Table~\ref{S-tab:leakage_rescore}). As a correctness check, every recomputed
full-test mIoU reproduces its logged value exactly.

The benchmark is not inflated. The mean leakage-free\,$-$\,overlapping
difference is $+0.33$\,pp at 12 classes and $-0.08$\,pp at 21 classes, i.e.\
models score marginally \emph{higher} on the leakage-free subset at 12 classes,
the opposite of what inflation predicts. Individual differences scatter in both
directions ($-3.16$ to $+4.13$) without a systematic sign. Most importantly the
\emph{ranking} is preserved: Spearman correlation between full-test and
leakage-free mIoU is \textbf{0.891} (12-class) and \textbf{0.997} (21-class),
and mean $|$full\,$-$\,leakage-free$|$ is 0.62 and 0.65\,pp. Utonia(L)\,FT, the
top model, scores 71.29 leakage-free against 71.28 on the full test split.

One caveat qualifies this. The leakage-free subset draws 162 scenes from path 7
and 18 from path 8, while the overlapping remainder draws 120 and 186
respectively, so the difference between subsets partly measures path-7-versus-8
character rather than overlap alone. The
near-zero result argues against inflation regardless, since a genuine
localisation effect would have to be cancelled by an equal and opposite path
effect to produce it.

\textbf{The leakage-free subset can be rebuilt from the scene bounds stored in
the released benchmark file}, so future work can report both numbers.

\begin{table*}[!tp]
\centering
\footnotesize
\setlength{\tabcolsep}{4pt}
\renewcommand{\arraystretch}{1.05}
\caption{\textbf{Every cached evaluation re-scored on the leakage-free and
overlapping halves of the test split}: 36 evaluations, 10 at 12 classes and 26
at 21 classes, spanning supervised baselines and all four adaptation regimes.
\textbf{full} is the published full-test mIoU, which every rerun reproduces
exactly; \textbf{free} is the 180-scene leakage-free subset; \textbf{ovl} is the
overlapping remainder. If geographic overlap inflated the benchmark, \textbf{free}
would sit systematically below \textbf{ovl}. It does not: mean $\Delta$ is
$+0.33$\,pp at 12 classes and $-0.08$\,pp at 21, individual values scatter in
both directions, and the full-\vs-leakage-free rank correlation is 0.891 and
0.997. Coverage is every model whose cached precise-test predictions reproduce
its published number; cells without such a cache (e.g.\ Concerto DP and
Concerto(L) LP at 12 classes) are absent. Concerto(O) denotes the outdoor-pretrained variant.}
\label{S-tab:leakage_rescore}
\begin{minipage}[t]{0.45\textwidth}
\centering
\textbf{12-class (10 evaluations)}\\[2pt]
\begin{tabular}{lcccc}
\toprule
\textbf{model} & \textbf{full} & \textbf{free} & \textbf{ovl} & $\Delta$ \\
\midrule
Utonia(L) FT & 71.28 & 71.29 & 69.64 & $+1.65$ \\
Utonia(L) LoRA & 69.77 & 69.85 & 68.21 & $+1.64$ \\
Concerto(O) FT & 68.10 & 66.59 & 69.21 & $-2.62$ \\
PTv3 sup. & 67.88 & 66.81 & 68.52 & $-1.71$ \\
LitePT-L sup. & 67.53 & 66.96 & 67.51 & $-0.55$ \\
Utonia(L) DP & 67.03 & 66.55 & 67.13 & $-0.58$ \\
KPConvX sup. & 66.78 & 66.67 & 65.62 & $+1.05$ \\
Utonia(L) LP & 64.18 & 64.84 & 62.18 & $+2.66$ \\
Utonia(L) LP 3ch & 63.16 & 62.12 & 62.76 & $-0.64$ \\
Concerto(O) LP & 52.41 & 53.07 & 50.63 & $+2.44$ \\
\midrule
\multicolumn{4}{l}{\textbf{mean} $\Delta$} & $\mathbf{+0.33}$ \\
\multicolumn{4}{l}{Spearman full \vs free} & 0.891 \\
\bottomrule
\end{tabular}
\end{minipage}\hfill
\begin{minipage}[t]{0.53\textwidth}
\centering
\textbf{21-class (26 evaluations)}\\[2pt]
\begin{tabular}{lcccc}
\toprule
\textbf{model} & \textbf{full} & \textbf{free} & \textbf{ovl} & $\Delta$ \\
\midrule
Utonia(L) FT & 61.73 & 60.96 & 62.76 & $-1.80$ \\
Concerto(L) FT & 60.83 & 60.37 & 61.48 & $-1.11$ \\
Sonata(L) FT & 59.68 & 58.63 & 61.07 & $-2.44$ \\
Utonia(L) LoRA & 58.97 & 58.16 & 59.30 & $-1.14$ \\
Concerto(O) FT & 58.97 & 58.10 & 60.19 & $-2.09$ \\
PTv3 9ch & 58.51 & 57.28 & 60.27 & $-2.99$ \\
PTv3 4ch & 58.35 & 56.87 & 60.03 & $-3.16$ \\
Utonia(L) DP & 57.77 & 56.84 & 58.99 & $-2.15$ \\
Concerto(L) DP & 56.40 & 56.29 & 55.09 & $+1.20$ \\
Concerto(L) LoRA & 56.01 & 55.87 & 54.82 & $+1.05$ \\
Sonata(L) DP & 55.90 & 56.23 & 54.75 & $+1.48$ \\
LitePT-L 9ch & 55.87 & 55.09 & 56.95 & $-1.86$ \\
Concerto(O) LoRA & 55.16 & 54.82 & 55.54 & $-0.72$ \\
LitePT-L 4ch & 55.09 & 54.08 & 55.99 & $-1.91$ \\
KPConvX 4ch & 54.47 & 54.14 & 53.55 & $+0.59$ \\
KPConvX 9ch & 53.96 & 53.63 & 53.94 & $-0.31$ \\
Utonia(L) LP & 53.72 & 53.95 & 53.56 & $+0.39$ \\
Utonia(L) LP 3ch & 53.09 & 52.57 & 52.69 & $-0.12$ \\
Concerto(O) DP & 52.30 & 52.05 & 52.56 & $-0.51$ \\
Sonata(L) LoRA & 52.13 & 52.55 & 50.47 & $+2.08$ \\
Concerto(L) LP 3ch & 50.23 & 49.81 & 49.75 & $+0.06$ \\
Concerto(L) LP & 50.23 & 50.44 & 49.41 & $+1.03$ \\
Sonata(L) LP & 46.39 & 47.64 & 43.51 & $+4.13$ \\
Concerto(O) LP & 44.61 & 45.32 & 42.84 & $+2.48$ \\
Concerto(O) LP 3ch & 43.20 & 44.02 & 41.22 & $+2.80$ \\
Sonata(L) LP 3ch & 34.30 & 35.26 & 32.23 & $+3.03$ \\
\midrule
\multicolumn{4}{l}{\textbf{mean} $\Delta$} & $\mathbf{-0.08}$ \\
\multicolumn{4}{l}{Spearman full \vs free} & 0.997 \\
\bottomrule
\end{tabular}
\end{minipage}
\end{table*}

\section{Extended Semantic Segmentation Results (21-class)}
\label{S-sec:semseg_21c}

Table~\ref{S-tab:semseg_21c_full} reports fine-grained 21-class semantic segmentation results on BhopalMLS, complementing the 12-class benchmark in the main paper (Table~4). The 21-class taxonomy is a finer grouping of the same 62 static classes. It refines five of the 12 major categories: Building (into Commercial and Building-other), Pole \& Tower (into Electric Pole, Street Light, and Pole-other, the last absorbing the rare Wire-Structure Connector), Vehicle (into 2-/3-/4-Wheeler and Hand Cart), Road Side Object (into Road Side Furniture, Shop Shade, Mannequin, and other), and Ground (into Ground and Stairs); Vegetation, Road, Wire, Board, and Fencing are unchanged, and Human \& Animal is reduced to Human (the rare Animal/Dog/Cow subclasses are excluded). We report all metrics and per-class IoU under the same supervised and transfer-learning protocols as the 12-class benchmark. These results should be read with one
caveat. The cross-pass agreement analysis (Sec.~\ref{S-sec:agreement}) shows that
label consistency is markedly lower at this granularity than at 12 classes
(82.2\% \vs 93.4\% overall), because the 21-class taxonomy splits precisely the
categories on which independent annotation passes agree least, most notably
\textit{Commercial} \vs \textit{Building (other)}. The 21-class benchmark
therefore carries more label noise than the 12-class one, and small differences
between methods on it should not be over-interpreted; we regard the 12-class
benchmark as the primary comparison and the 21-class one as a harder,
finer-grained secondary task.

\begin{table*}[!tp]
\centering
\small
\setlength{\tabcolsep}{4pt}
\renewcommand{\arraystretch}{1.15}
\caption{\textbf{Semantic segmentation (21-class)} on BhopalMLS. Per-class IoU~(\%). LP = Linear Probe, DP = Decoder Probe, LoRA = Low-Rank Adaptation, FT = Fine-Tuning. 3ch = XYZ; 4ch = XYZ + intensity; 9ch = XYZ + Partial RGB + normals. Sonata(L), Concerto(L) and Utonia(L) = Large (general-purpose pretrained), Concerto(O) = outdoor-pretrained. $^{\dagger}$=trained jointly with the 6 moving-class subclasses (Moving Vehicle, Moving Humans, Moving Female, Moving Male, Moving Animal, Mobike Rider Moving) as additional fine-grained supervision; folded into their static parents at evaluation. \colorbox{bestgreen}{\textbf{Bold green}} = overall best per column; \underline{underline} = best within category; \textcolor{red}{red} = overall lowest.}
\label{S-tab:semseg_21c_full}
\resizebox{\textwidth}{!}{%
\begin{tabular}{l|ccc|ccccccccccccccccccccc}
\hline
\textbf{Model} &
\rotatebox{90}{\textbf{OA (\%)}} &
\rotatebox{90}{\textbf{mAcc (\%)}} &
\rotatebox{90}{\textbf{mIoU (\%)}} &
\rotatebox{90}{\textbf{Ground}} &
\rotatebox{90}{\textbf{Stairs}} &
\rotatebox{90}{\textbf{Vegetation}} &
\rotatebox{90}{\textbf{Building (other)}} &
\rotatebox{90}{\textbf{Commercial}} &
\rotatebox{90}{\textbf{Road}} &
\rotatebox{90}{\textbf{Wire}} &
\rotatebox{90}{\textbf{Pole (other)}} &
\rotatebox{90}{\textbf{Electric Pole}} &
\rotatebox{90}{\textbf{Street Light}} &
\rotatebox{90}{\textbf{2-Wheeler}} &
\rotatebox{90}{\textbf{3-Wheeler}} &
\rotatebox{90}{\textbf{4-Wheeler}} &
\rotatebox{90}{\textbf{Hand Cart}} &
\rotatebox{90}{\textbf{Board}} &
\rotatebox{90}{\textbf{Fencing}} &
\rotatebox{90}{\textbf{Human}} &
\rotatebox{90}{\textbf{Road Side Furniture}} &
\rotatebox{90}{\textbf{Shop Shade}} &
\rotatebox{90}{\textbf{Mannequin}} &
\rotatebox{90}{\textbf{Road Side Obj (other)}} \\
\hline
\multicolumn{25}{l}{\textbf{Supervised Learning -- 4ch (XYZ + intensity)}} \\
\hline
PTv3 \cite{wu2024point} & \underline{86.47} & 71.13 & \underline{58.35} & \underline{16.80} & \underline{14.25} & \underline{97.04} & 30.44 & \underline{84.02} & \underline{87.08} & \underline{86.61} & \underline{70.47} & \underline{66.03} & \underline{58.03} & \underline{95.09} & \underline{86.28} & \underline{83.00} & \underline{62.66} & \underline{60.82} & 21.06 & \underline{65.08} & \underline{38.43} & 58.90 & \underline{28.56} & \underline{14.78} \\
LitePT-L \cite{yuelitept2026} & 84.73 & \underline{71.70} & 55.09 & 16.67 & 13.51 & 96.52 & 29.57 & 82.25 & 86.53 & 85.08 & 65.95 & 64.61 & 52.38 & 93.57 & 85.79 & 73.33 & 47.63 & 51.39 & \underline{22.48} & 63.32 & 34.26 & 57.10 & 25.18 & 9.83 \\
KpConvX-L \cite{thomas2024kpconvx} & 84.09 & 71.23 & 54.47 & 15.96 & 13.59 & 97.03 & \underline{32.13} & 81.21 & 85.34 & 85.66 & 68.74 & 62.49 & 47.15 & 92.24 & 74.84 & 77.13 & 51.69 & 55.42 & 20.36 & 60.75 & 36.74 & \underline{61.08} & 19.82 & 4.47 \\
\hline
\multicolumn{25}{l}{\textbf{Supervised Learning -- 9ch (XYZ + Partial RGB + normals)}} \\
\hline
PTv3             & \underline{86.61} & 70.70 & \underline{58.51} & \underline{18.40} & 13.51 & \underline{96.79} & 28.15 & \underline{84.19} & \underline{87.06} & \underline{84.76} & \underline{71.14} & \underline{66.89} & \underline{56.15} & \underline{94.16} & \underline{88.92} & \underline{84.17} & \underline{64.74} & 57.27 & \underline{22.24} & \underline{65.89} & \underline{38.83} & \underline{59.68} & \underline{31.55} & \underline{14.20} \\
LitePT-L         & 84.48 & \underline{71.99} & 55.87 & 16.82 & 13.57 & 96.39 & 26.49 & 81.96 & 85.99 & 83.21 & 69.80 & 64.95 & 53.16 & 93.68 & 82.73 & 74.80 & 56.29 & \underline{58.03} & 22.18 & 64.17 & 34.78 & 59.00 & 25.48 & 9.72 \\
KpConvX-L        & 84.30 & 71.24 & 53.96 & 17.29 & \underline{14.52} & 96.46 & \underline{29.41} & 82.01 & 85.23 & 82.61 & 65.80 & 60.43 & 49.59 & 91.94 & 77.20 & 78.35 & 49.32 & 54.45 & 17.20 & 58.62 & 38.25 & 58.99 & 21.68 & 3.82 \\
\hline
\multicolumn{25}{l}{\textbf{Linear Probe (3ch: XYZ)}} \\
\hline
Sonata(L) \cite{wu2025sonata} & \textcolor{red}{77.40} & \textcolor{red}{52.31} & \textcolor{red}{34.30} & \textcolor{red}{15.31} & \textcolor{red}{10.24} & \textcolor{red}{75.52} & \textcolor{red}{26.18} & \textcolor{red}{75.84} & \textcolor{red}{82.42} & \textcolor{red}{56.81} & \textcolor{red}{18.87} & \textcolor{red}{34.55} & \textcolor{red}{24.65} & \textcolor{red}{70.91} & \textcolor{red}{39.12} & \textcolor{red}{34.51} & \textcolor{red}{14.30} & \textcolor{red}{29.78} & 10.13 & \textcolor{red}{32.16} & \textcolor{red}{14.97} & \textcolor{red}{43.00} & \textcolor{red}{8.61} & 2.44 \\
Concerto(O)          & 80.38 & 62.01 & 43.20 & 17.01 & 12.39 & 91.82 & 30.33 & 78.49 & 83.31 & 65.92 & 38.06 & 42.61 & 31.78 & 75.55 & 63.92 & 62.71 & 37.05 & 34.43 & 15.24 & 38.53 & 21.13 & 51.39 & 12.46 & 3.03 \\
Concerto(L) \cite{zhang2025concerto} & 84.62 & 69.35 & 50.23 & 18.06 & \underline{14.20} & \underline{95.14} & 34.65 & \underline{82.72} & \underline{86.29} & 80.73 & 51.15 & 54.30 & 46.40 & 87.87 & 60.65 & 61.29 & 49.20 & 51.47 & 20.46 & 50.40 & 32.71 & \underline{56.52} & 17.05 & 3.58 \\
Utonia(L) \cite{zhang2026utonia} & \underline{84.72} & \underline{71.53} & \underline{53.09} & \underline{18.21} & 13.98 & \underline{96.46} & \underline{35.33} & 82.56 & 85.86 & \underline{82.35} & \underline{54.09} & \underline{58.96} & \underline{47.63} & \underline{89.96} & \underline{73.45} & \underline{81.66} & \underline{50.08} & \underline{52.03} & \underline{24.10} & \underline{50.67} & \underline{33.66} & 56.14 & \underline{19.95} & \underline{7.68} \\
\hline
\multicolumn{25}{l}{\textbf{Linear Probe (9ch: XYZ + Partial RGB + normals)}} \\
\hline
Sonata(L)            & 82.99 & 64.52 & 46.39 & 17.17 & 13.16 & 92.58 & 28.24 & 81.03 & 86.01 & 75.40 & 35.24 & 49.20 & 45.76 & 81.99 & 61.61 & 56.09 & 41.06 & 41.13 & 27.31 & 44.40 & 24.00 & 52.46 & 17.41 & 2.88 \\
Concerto(O)          & 82.78 & 61.71 & 44.61 & 15.84 & 14.35 & 92.81 & 31.82 & 81.10 & 85.71 & 72.19 & 29.91 & 42.73 & 33.68 & 77.72 & 67.78 & 72.32 & 35.15 & 36.11 & 15.60 & 41.83 & 20.62 & 53.40 & 13.99 & 2.04 \\
Concerto(O)$^{\dagger}$ & 82.31 & 61.21 & 43.98 & 15.51 & 13.82 & 92.75 & 32.19 & 80.65 & 85.49 & 72.22 & 29.41 & 41.90 & 34.97 & 73.87 & 57.69 & 65.24 & 32.98 & 35.16 & 16.49 & 56.44 & 19.42 & 52.86 & 12.65 & \textcolor{red}{1.85} \\
Concerto(L)          & 84.60 & 69.39 & 50.23 & 17.98 & 15.62 & 95.20 & 34.32 & \underline{82.51} & 86.33 & 80.20 & 46.52 & 55.40 & 44.77 & 88.51 & 63.78 & 65.40 & 46.67 & 50.46 & 22.55 & 52.71 & 30.69 & \underline{55.20} & 17.90 & 2.06 \\
Utonia(L)            & \underline{84.77} & \underline{72.92} & \underline{53.72} & 18.26 & \underline{15.88} & 96.68 & 36.05 & 82.09 & \underline{87.23} & \underline{82.04} & 54.44 & \underline{59.01} & \underline{50.38} & \underline{89.90} & \underline{73.63} & \underline{79.96} & \underline{47.17} & 54.15 & \underline{34.42} & 52.58 & \underline{30.92} & 54.19 & \underline{19.33} & \underline{9.90} \\
Utonia(L)$^{\dagger}$ & 84.76 & 71.36 & 53.44 & \underline{18.50} & 15.38 & \underline{96.69} & \underline{36.50} & 82.29 & 86.94 & 81.98 & \underline{55.10} & 58.56 & 49.78 & 86.12 & 71.54 & 79.59 & 47.16 & \underline{54.62} & 28.19 & \underline{64.04} & 29.42 & 54.12 & 16.48 & 9.19 \\
\hline
\multicolumn{25}{l}{\textbf{Decoder Probe (9ch: XYZ + Partial RGB + normals)}} \\
\hline
Sonata(L)            & 86.42 & 69.67 & 55.90 & 18.08 & 15.40 & 96.15 & 33.00 & 84.58 & 86.86 & 84.45 & 66.13 & 62.31 & 59.88 & 92.97 & 78.55 & 79.86 & 50.89 & 56.98 & \underline{20.42} & 61.60 & 35.54 & 55.29 & 29.34 & 5.65 \\
Concerto(O)          & 84.62 & 67.16 & 52.30 & 16.02 & 15.08 & 96.00 & 29.72 & 82.41 & 86.38 & 81.49 & 52.07 & 56.51 & 50.60 & 91.35 & 70.32 & 73.34 & 49.89 & 46.62 & 18.51 & 57.96 & 33.12 & 55.81 & 26.86 & 8.30 \\
Concerto(O)$^{\dagger}$ & 82.52 & 66.43 & 51.20 & 15.57 & 14.05 & 94.37 & 27.37 & 79.18 & 85.86 & 79.39 & 50.92 & 53.91 & 44.04 & 88.27 & 70.85 & 74.22 & 44.75 & 51.38 & 18.95 & 70.16 & 31.25 & 55.36 & 21.51 & 3.89 \\
Concerto(L)          & 85.45 & 70.98 & 56.40 & 18.18 & \underline{16.16} & 96.90 & 30.14 & 82.72 & 86.82 & 85.07 & 67.27 & \underline{64.66} & \underline{60.87} & 94.08 & 81.24 & 80.03 & 57.49 & \underline{59.13} & 16.98 & 63.28 & 37.13 & 55.30 & 26.38 & 4.63 \\
Utonia(L)            & \underline{87.44} & 71.97 & 57.77 & \underline{18.37} & 16.05 & 97.19 & \underline{36.01} & \underline{85.65} & \underline{87.46} & 86.20 & 68.16 & 64.04 & 60.24 & \underline{94.61} & \underline{81.41} & 85.98 & 58.40 & 57.99 & 13.20 & 64.79 & 37.01 & \underline{57.88} & 28.72 & 13.84 \\
Utonia(L)$^{\dagger}$ & 87.23 & \underline{72.67} & \underline{58.94} & 18.14 & 15.78 & \underline{97.55} & 30.99 & 85.13 & 87.34 & \underline{86.38} & \underline{69.04} & 64.41 & 57.80 & 93.45 & 81.04 & \underline{87.70} & \underline{60.20} & \underline{59.13} & 18.74 & \underline{79.74} & \underline{38.22} & 57.83 & \underline{31.54} & \underline{17.61} \\
\hline
\multicolumn{25}{l}{\textbf{LoRA (9ch: XYZ + Partial RGB + normals)}} \\
\hline
Sonata(L)            & 84.36 & 71.21 & 52.13 & 18.30 & 15.04 & 95.63 & 29.75 & 82.01 & 86.55 & 82.38 & 57.14 & 57.76 & 46.71 & 88.95 & 71.90 & 73.43 & 44.43 & 52.47 & 25.88 & 52.39 & 31.40 & 55.83 & 18.67 & 8.15 \\
Concerto(O)          & 85.49 & 71.86 & 55.16 & 17.63 & \underline{16.61} & 96.86 & 36.23 & \underline{83.24} & 86.57 & 80.80 & 62.11 & 59.63 & 54.16 & 89.95 & 79.32 & 83.99 & 52.76 & 53.50 & 22.03 & 54.20 & 34.30 & 55.58 & \underline{24.82} & 14.10 \\
Concerto(L)          & 85.46 & 72.45 & 56.01 & \cellcolor{bestgreen}\underline{\textbf{20.05}} & 16.30 & 96.88 & 35.48 & 82.63 & \underline{87.25} & 82.21 & 62.38 & 58.69 & 55.75 & 92.57 & 79.95 & 80.42 & \underline{58.95} & 57.86 & 28.04 & 57.97 & 36.59 & \underline{58.77} & 21.39 & 5.98 \\
Utonia(L)            & \underline{85.88} & \cellcolor{bestgreen}\underline{\textbf{75.22}} & \underline{58.97} & 19.16 & 15.60 & \underline{97.61} & \cellcolor{bestgreen}\underline{\textbf{36.75}} & 82.99 & 87.08 & \underline{84.25} & \underline{68.10} & \underline{64.65} & \underline{58.98} & \underline{93.34} & \underline{88.02} & \cellcolor{bestgreen}\underline{\textbf{91.51}} & 54.03 & \underline{61.07} & \cellcolor{bestgreen}\underline{\textbf{34.88}} & \underline{60.68} & \underline{37.01} & 55.90 & 24.56 & \cellcolor{bestgreen}\underline{\textbf{22.24}} \\
\hline
\multicolumn{25}{l}{\textbf{Fine-Tuning (9ch: XYZ + Partial RGB + normals)}} \\
\hline
Sonata(L)            & 86.54 & 72.93 & 59.68 & 18.40 & 17.66 & 97.43 & 31.07 & 83.96 & 87.21 & 86.69 & 70.76 & 67.87 & 57.81 & 95.07 & 92.69 & 89.08 & 67.75 & 62.84 & 10.85 & 65.70 & 41.48 & 61.81 & 32.24 & 14.85 \\
Concerto(O)          & \cellcolor{bestgreen}\underline{\textbf{88.07}} & 71.62 & 58.97 & 17.06 & 14.50 & 97.36 & 34.99 & \cellcolor{bestgreen}\underline{\textbf{86.17}} & 87.59 & 86.49 & 63.63 & 65.59 & 53.82 & 94.67 & \cellcolor{bestgreen}\underline{\textbf{92.81}} & 90.60 & 62.55 & 62.87 & 15.86 & 67.84 & 39.99 & 56.07 & 33.55 & 14.34 \\
Concerto(O)$^{\dagger}$ & 87.74 & 72.65 & 59.24 & 18.17 & 14.18 & 97.65 & \underline{36.33} & 85.81 & 86.70 & 85.36 & 71.07 & 66.43 & 59.13 & 94.25 & 76.37 & 82.77 & 68.76 & 62.00 & \textcolor{red}{9.93} & 81.20 & 40.30 & 60.49 & 30.08 & 17.08 \\
Concerto(L)          & 86.63 & \underline{74.60} & 60.83 & 17.29 & 16.34 & 97.53 & 31.53 & 83.80 & 87.60 & 87.26 & \cellcolor{bestgreen}\underline{\textbf{72.53}} & \cellcolor{bestgreen}\underline{\textbf{70.22}} & 62.40 & 95.23 & 91.15 & 89.24 & \cellcolor{bestgreen}\underline{\textbf{69.12}} & 63.38 & 21.90 & 67.38 & 44.50 & 60.11 & \cellcolor{bestgreen}\underline{\textbf{33.79}} & 15.17 \\
Utonia(L)            & 87.43 & 74.17 & 61.73 & 18.90 & \cellcolor{bestgreen}\underline{\textbf{18.50}} & \cellcolor{bestgreen}\underline{\textbf{97.94}} & 30.75 & 84.83 & \cellcolor{bestgreen}\underline{\textbf{88.09}} & 87.69 & 70.76 & 68.44 & \cellcolor{bestgreen}\underline{\textbf{63.07}} & \cellcolor{bestgreen}\underline{\textbf{95.84}} & 92.12 & 90.43 & 66.85 & \cellcolor{bestgreen}\underline{\textbf{70.03}} & 23.94 & 72.60 & \cellcolor{bestgreen}\underline{\textbf{44.54}} & 62.23 & 32.89 & 15.91 \\
Utonia(L)$^{\dagger}$ & 87.92 & 74.41 & \cellcolor{bestgreen}\underline{\textbf{61.95}} & \underline{19.27} & 13.27 & 97.73 & 35.40 & 85.50 & 87.30 & \cellcolor{bestgreen}\underline{\textbf{88.08}} & 70.35 & 67.89 & 59.52 & 95.47 & 89.11 & \underline{90.99} & 67.03 & 67.10 & \underline{26.98} & \cellcolor{bestgreen}\underline{\textbf{83.41}} & 42.94 & \cellcolor{bestgreen}\underline{\textbf{63.24}} & 30.43 & \underline{19.88} \\
\hline
\end{tabular}%
}%
\end{table*}

Consistent with the 12-class benchmark, Utonia(L) with fine-tuning (trained jointly with the 6 moving-class subclasses, $^{\dagger}$) achieves the best overall mIoU (61.95\%), followed by Utonia(L) FT (61.73\%) and Concerto(L) FT (60.83\%). Fine-grained Indian-urban classes such as 2-Wheeler (95.84\% IoU at best), 3-Wheeler (92.81\%), and Commercial Building (86.17\%) reach high performance under fine-tuning, while structurally ambiguous categories (Fencing, Stairs, Road Side Object) remain the hardest classes in the dataset.

\begin{figure*}[!tp]
    \centering
    \includegraphics[width=\textwidth]{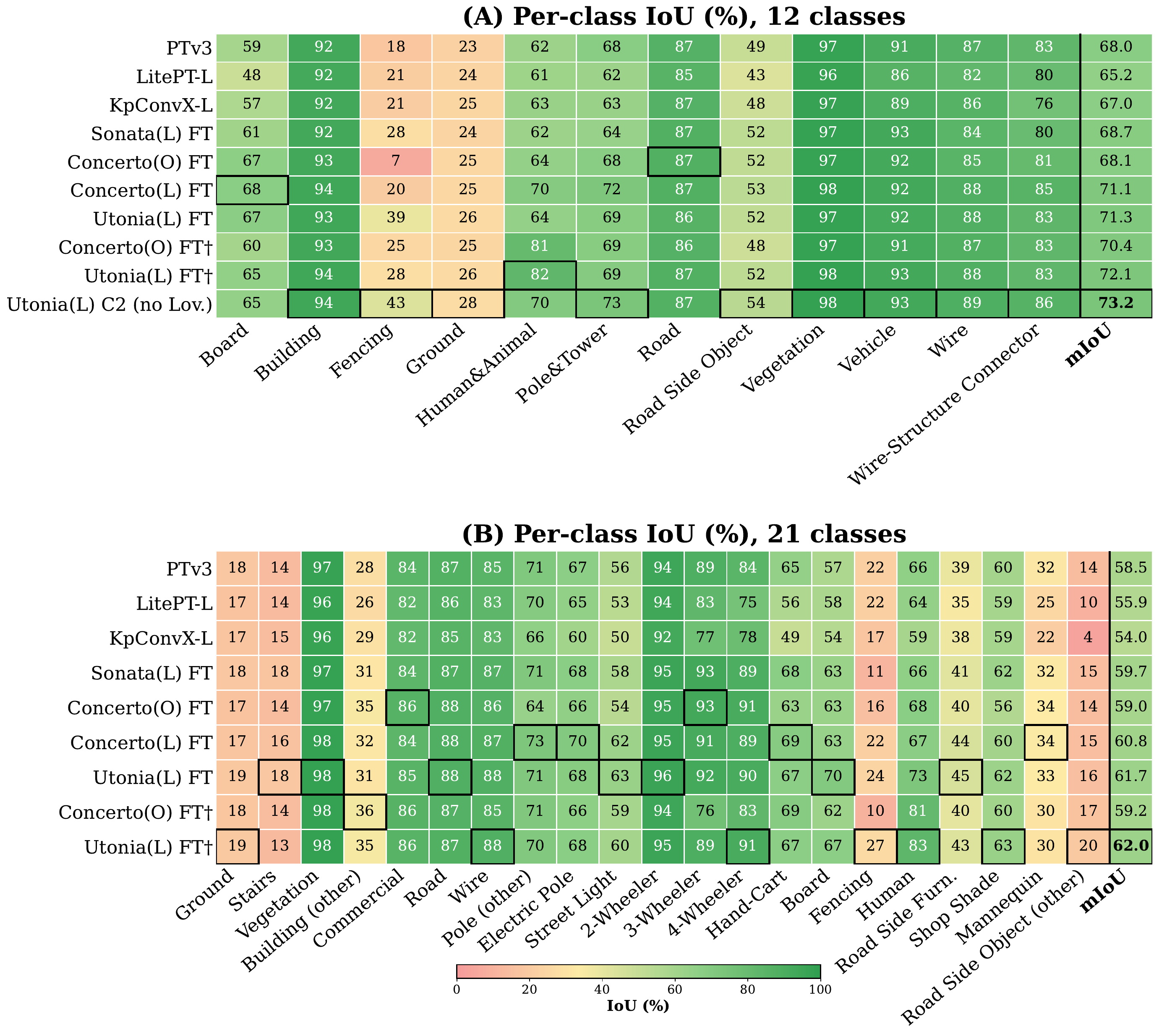}
    \caption{\textbf{Per-class IoU heatmaps} for \textbf{(A)} the 12-class benchmark (Table~\ref{tab:bhopal_iou_benchmark_style_rotated_vline}) and \textbf{(B)} the 21-class benchmark (Table~\ref{S-tab:semseg_21c_full}). Each cell is the per-class IoU (\%) of a method (row) on a class (column), shaded light-to-dark with IoU; the best method in each column is outlined in black. Rows cover the supervised 9ch baselines (PTv3, LitePT-L, KpConvX-L), the fine-tuned foundation models (Sonata(L), Concerto(O), Concerto(L), Utonia(L)), their joint moving-class $^{\dagger}$ variants (Concerto(O), Utonia(L)), and the best class-imbalance recipe Utonia(L)~C2 (no Lov\'asz; 12-class only). The separated rightmost column reports each method's overall mIoU (\%), again with the best boxed.}
    \label{S-fig:classwise-radar}
\end{figure*}

Figure~\ref{S-fig:classwise-radar} presents the per-class IoU of the strongest configurations as a heatmap: the three supervised 9ch baselines (PTv3, LitePT-L, KpConvX-L) and the fine-tuned foundation models (Sonata(L), Concerto(O), Concerto(L), Utonia(L), and the joint moving-class $^{\dagger}$ variants of Concerto(O) and Utonia(L)), plus the best class-imbalance recipe Utonia(L)~C2 (Sec.~\ref{S-sec:class_imbalance_ablation}; 12-class only). The color encodes IoU and each column's best method is boxed, making the per-class spread explicit: vegetation, building, road, and vehicle are handled well across all models, whereas Fencing, Ground, Stairs, and Road Side Object remain low; the rightmost column reports each method's overall mIoU.

\section{Class-Imbalance Ablation}
\label{S-sec:class_imbalance_ablation}

To understand the contribution of each component of our class-imbalance handling recipe (Section~4.1 of the main paper), Table~\ref{S-tab:class_imbalance_ablation} reports a controlled ablation of the three ingredients (square-root inverse-frequency class weights, Lov\'asz loss, and 10\% Building downsampling) across two architectures (Concerto(L)~\cite{zhang2025concerto} and Utonia(L)~\cite{zhang2026utonia}). The five settings are:
\begin{itemize}
    \item \textbf{(C0)} full recipe: class-weights + Lov\'asz + Building DS (baseline; matches Table~\ref{tab:bhopal_iou_benchmark_style_rotated_vline} FT-9ch).
    \item \textbf{(C1)} no class-weights: Lov\'asz + Building DS only.
    \item \textbf{(C2)} no Lov\'asz: class-weights + Building DS only.
    \item \textbf{(C3)} CE only ($\equiv$ C1$+$C2): plain cross-entropy + Building DS, no class-weights, no Lov\'asz.
    \item \textbf{(C4)} no Building DS: class-weights + Lov\'asz, no Building downsampling.
\end{itemize}

\begin{table*}[!tp]
\centering
\small
\setlength{\tabcolsep}{4pt}
\renewcommand{\arraystretch}{1.15}
\caption{\textbf{Class-imbalance ablation} on BhopalMLS (12-class, FT-9ch) across two architectures. (C0) is the full recipe (matches Table~\ref{tab:bhopal_iou_benchmark_style_rotated_vline}); (C1--C4) remove or change one ingredient at a time. All per-class values are IoU (\%); mIoU column shows $\Delta$ vs (C0) in parentheses. \colorbox{bestgreen}{\textbf{Bold green}} = overall best across all 10 rows for that column.}
\label{S-tab:class_imbalance_ablation}
\resizebox{\textwidth}{!}{%
\begin{tabular}{l|l|ccc|cccccccccccc}
\toprule
\textbf{Model} & \textbf{Setting} &
\rotatebox{90}{\textbf{OA (\%)}} &
\rotatebox{90}{\textbf{mAcc (\%)}} &
\rotatebox{90}{\textbf{mIoU (\%)}} &
\rotatebox{90}{\textbf{Board}} &
\rotatebox{90}{\textbf{Building}} &
\rotatebox{90}{\textbf{Fencing}} &
\rotatebox{90}{\textbf{Ground}} &
\rotatebox{90}{\textbf{Human\&Animal}} &
\rotatebox{90}{\textbf{Pole\&Tower}} &
\rotatebox{90}{\textbf{Road}} &
\rotatebox{90}{\textbf{Road Side Obj.}} &
\rotatebox{90}{\textbf{Vegetation}} &
\rotatebox{90}{\textbf{Vehicle}} &
\rotatebox{90}{\textbf{Wire}} &
\rotatebox{90}{\textbf{Wire-Structure Conn.}} \\
\midrule
\multirow{5}{*}{\textbf{Concerto(L)}}
& (C0) full recipe                       & 93.42 & 80.44 & 71.06            & 67.72 & 93.71 & 20.14 & 25.41 & 69.88 & 72.48 & 87.11 & 53.07 & 97.67 & 92.34 & 87.91 & 85.26 \\
& (C1) no class-weights                  & 93.46 & 77.84 & 69.44\,($-$1.62) & 62.52 & 93.63 &  8.20 & 23.52 & 71.07 & 71.98 & 87.29 & 50.74 & 97.64 & 91.80 & 88.62 & \cellcolor{bestgreen}\textbf{86.29} \\
& (C2) no Lov\'asz                       & 93.03 & 81.22 & 70.67\,($-$0.39) & 66.45 & 93.26 & 22.50 & 25.04 & 69.08 & 71.93 & 86.62 & 50.74 & 97.69 & 92.69 & 87.39 & 84.64 \\
& (C3) CE only (C1$+$C2)                 & 93.52 & 78.16 & 70.21\,($-$0.85) & 68.09 & 93.79 & 14.63 & 24.52 & 68.66 & 71.76 & 87.42 & 50.17 & 97.62 & 92.20 & 88.81 & 84.80 \\
& (C4) no Building downsampling          & 93.48 & 79.64 & 70.44\,($-$0.62) & 64.74 & 93.81 & 21.27 & 23.51 & 69.63 & 72.85 & 86.48 & 48.83 & 97.76 & 92.84 & 88.11 & 85.44 \\
\midrule
\multirow{5}{*}{\textbf{Utonia(L)}}
& (C0) full recipe                       & 92.69 & 83.89 & 71.28            & 67.36 & 93.38 & 38.79 & 26.27 & 64.24 & 68.61 & 85.66 & 51.51 & 97.41 & 91.54 & 87.53 & 83.03 \\
& (C1) no class-weights                  & \cellcolor{bestgreen}\textbf{93.83} & 79.60 & 71.65\,($+$0.37) & \cellcolor{bestgreen}\textbf{69.93} & \cellcolor{bestgreen}\textbf{94.08} & 16.30 & 26.30 & \cellcolor{bestgreen}\textbf{71.78} & \cellcolor{bestgreen}\textbf{73.45} & 87.54 & \cellcolor{bestgreen}\textbf{55.20} & 97.83 & 93.13 & 88.86 & 85.36 \\
& (C2) no Lov\'asz                       & 93.43 & \cellcolor{bestgreen}\textbf{83.90} & \cellcolor{bestgreen}\textbf{73.20\,($+$1.92)} & 64.55 & 93.74 & \cellcolor{bestgreen}\textbf{43.08} & \cellcolor{bestgreen}\textbf{27.70} & 70.16 & 73.12 & 86.88 & 53.62 & 97.82 & 92.90 & 88.77 & 86.02 \\
& (C3) CE only (C1$+$C2)                 & 93.78 & 78.90 & 70.97\,($-$0.31) & 66.65 & \cellcolor{bestgreen}\textbf{94.08} & 13.54 & 26.03 & 71.54 & 72.75 & \cellcolor{bestgreen}\textbf{87.65} & 53.34 & 97.73 & \cellcolor{bestgreen}\textbf{93.19} & \cellcolor{bestgreen}\textbf{89.06} & 86.05 \\
& (C4) no Building downsampling          & 93.71 & 80.49 & 70.52\,($-$0.76) & 64.25 & 94.05 & 18.42 & 25.74 & 69.43 & 71.90 & 86.99 & 53.96 & \cellcolor{bestgreen}\textbf{97.88} & 92.44 & 88.82 & 82.43 \\
\bottomrule
\end{tabular}%
}%
\end{table*}

The full per-class results reveal a sharp architecture-dependent picture:

\noindent\textbf{Concerto(L)} loses mIoU under every ablation (max drop: C1 $-$1.62), confirming that class-weights, Lov\'asz loss, and Building DS all contribute. The cost concentrates on the rarest class: Fencing drops by $-$11.94 (C1), $-$5.51 (C3) when class-weights are removed; class-weights are decisive for Fencing.

\noindent\textbf{Utonia(L)} responds very differently. (C2) ``no Lov\'asz'' is the best setting at \textbf{73.20\%} mIoU ($+$1.92 vs C0), and 11 of 12 classes \emph{improve} simultaneously, including a notable Fencing gain from 38.79 to \textbf{43.08} ($+$4.29). Lov\'asz appears to over-regularize Utonia's representations on minority-class boundaries. However, removing class-weights (C1) still collapses Fencing for Utonia ($-$22.49, the largest single per-class loss in the entire table), and the combined CE-only setting (C3) is even worse on Fencing ($-$25.25), confirming super-additive interaction: dropping both ingredients is more harmful than the sum of dropping each.

\noindent\textbf{Building downsampling (C4)} has the mildest effect on both architectures ($-$0.62 / $-$0.76 mIoU). It slightly \emph{helps} Building IoU as expected ($+$0.10 Co / $+$0.67 Ut), but the cost falls on small, cluttered classes that share spatial context with buildings: Road Side Object loses $-$4.24 (Co) and Board loses $-$2.98 (Co), $-$3.11 (Ut). Across all 10 rows, Utonia variants hold the column-best on \textbf{14 of 15} columns; only Wire-Structure Connector is won by Concerto(L) C1 (86.29).

\section{Instance Segmentation Baselines}

To demonstrate the dataset's utility beyond semantic segmentation, we evaluate instance segmentation with two supervised baselines trained from scratch (PTv3~\cite{wu2024point} and LitePT-L~\cite{yuelitept2026}) and the pretrained foundation models Sonata(L)~\cite{wu2025sonata}, Concerto~\cite{zhang2025concerto} (general-purpose Concerto(L) and outdoor Concerto(O) variants), and Utonia(L)~\cite{zhang2026utonia} under three adaptation strategies: Linear Probe (LP), Decoder Probe (DP), and Fine-Tuning (FT). The supervised baselines let us compare foundation-model fine-tuning against training from scratch. All architectures natively support the PointGroup instance head within the Pointcept framework. The instance-segmentation task setup (thing/stuff split, instance IDs, and the 12- and 21-class granularities) is introduced in the main paper; here we provide the full per-class results.

\subsection{Thing vs Stuff Classes}

The four stuff classes are predicted semantically but excluded from instance clustering and mAP; the remaining thing classes are grouped into instances. The exact membership in each setting is:

\textbf{12-class setting} (Table~\ref{S-tab:instance_seg_12c}): 4 stuff classes (Ground, Vegetation, Road, Wire) and 8 thing classes (Building, Pole\&Tower, Wire-Structure Conn., Vehicle, Board, Fencing, Human\&Animal, Road Side~Object).

\textbf{21-class setting} (Table~\ref{S-tab:instance_seg_21c}): The same 4 stuff classes (Ground, Vegetation, Road, Wire), with 17 fine-grained thing classes obtained by splitting Building, Pole \& Tower, Vehicle, and Road Side Object into finer categories, carving Stairs out of Ground, and merging the rare Wire-Structure Connector into Pole (other). Specifically: Stairs, Building (other), Commercial Building, Pole (other), Electric Pole, Street Light, 2-Wheeler, 3-Wheeler, 4-Wheeler, Hand Cart, Board, Fencing, Human, Road Side Furniture, Shop Shade, Mannequin, and Road Side Object (other).

\subsection{Experimental Setup}

All models use the PointGroup~\cite{jiang2020pointgroup} (PG-v1m2) instance segmentation head built on Pointcept, which jointly predicts semantic labels and instance offsets, with 9-channel input (XYZ + partial RGB + normals). Models are evaluated on the same test split (Path~07--08) as the semantic segmentation benchmarks.

\subsection{Results}

Beyond the headline mAP50 reported in the main paper, the per-class breakdown shows that fine-tuning yields the largest gains on object classes that linear probing handles poorly: for Utonia(L), Board AP50 rises from 8.22\% (LP) to 55.67\% (FT), Wire-Structure Connector from 7.21\% to 59.47\%, and Building from 17.06\% to 53.70\%, while structurally ambiguous Fencing remains the hardest class (0.31\%$\rightarrow$5.43\%). The 21-class evaluation further tests whether models can delineate fine-grained categories such as 2-wheelers (6{,}335 instances), 3-wheelers, and hand carts that are unique to dense, unstructured urban scenes and absent from existing benchmarks.

In the 12-class setting (Table~\ref{S-tab:instance_seg_12c}), Utonia(L) FT is the overall best on mAP25/mAP50/mAP and on per-class \textit{Building} (53.70), \textit{Wire-Structure Conn.} (59.47), and \textit{Board} (55.67), while Concerto(L)~FT remains overall best on \textit{Vehicle} (73.68), \textit{Fencing} (6.32), and \textit{Road Side Object} (19.00); supervised LitePT-L wins \textit{Pole \& Tower} (39.15) and \textit{Human \& Animal} (54.84). In the 21-class setting (Table~\ref{S-tab:instance_seg_21c}), Utonia(L) FT again leads (39.83\% mAP50, 30.63\% mAP), outperforming Concerto(L)~FT (36.71\%) and Concerto(O)~FT (35.47\%), with the largest margins on India-specific fine-grained classes: \textit{Commercial Building} (62.44 vs.\ 40.73), \textit{Building (other)} (31.11 vs.\ 11.12), \textit{Road Side Object (other)} (6.12 vs.\ 3.94), and \textit{Board} (55.76 vs.\ 52.83). Supervised baselines remain competitive: PTv3 9ch (33.07\% mAP50) outperforms all LP/DP foundation-model variants, and LitePT-L still holds the best \textit{Human} AP50 (55.02) and \textit{Street Light} AP50 (39.06) overall, demonstrating that training from scratch can still excel on specific categories. Comparing pretraining strategies, Concerto(L) consistently outperforms Concerto(O) across all adaptation methods (36.71 vs.\ 35.47 mAP50 for FT in 21-class), mirroring the semantic segmentation trends. Concerto(O) LP is particularly weak (\textcolor{red}{4.07\%} mAP50 in 12-class and \textcolor{red}{4.44\%} in 21-class, both the overall lowest), indicating that outdoor-specific pretraining alone provides insufficient inductive bias for instance-level grouping in complex urban scenes.

Across both granularities, Fencing remains the hardest class in the 12-class setting (best AP50: 6.32\%), reflecting its thin planar geometry and frequent confusion with adjacent structures. In the 21-class setting, Stairs (best AP50: 2.60\%) and Fencing (7.80\%) are the most challenging classes, while fine-grained building distinction also proves difficult: Building (other) peaks at 31.11\% AP50 despite Commercial Building reaching 62.44\%, suggesting that generic building instances lack discriminative geometric cues. India-specific classes such as Shop Shade (best 31.41\%) and Mannequin (best 11.93\%) highlight the unique challenges of dense, unstructured urban scenes that are absent from existing instance segmentation benchmarks.

\begin{table*}[!tp]
\centering
\footnotesize
\setlength{\tabcolsep}{4pt}
\renewcommand{\arraystretch}{1.1}
\caption{\textbf{Instance segmentation results (12-class)} on BhopalMLS using PTv3 and LitePT-L (trained from scratch) and the pretrained foundation models Sonata, Concerto, and Utonia with LP, DP, and FT. All models use 9ch input. mAP, mAP50, and mAP25 are computed over 8 thing classes at IoU thresholds averaged over 50:5:95, 50\%, and 25\%, respectively; 4 stuff classes (Ground, Vegetation, Road, Wire) are excluded from instance evaluation. Per-class values are AP50 (\%). Sonata(L), Concerto(L) and Utonia(L) = Large (general-purpose pretrained), Concerto(O) = outdoor-pretrained. \colorbox{bestgreen}{\textbf{Bold green}} = overall best per column; \underline{underline} = best within category; \textcolor{red}{red} = overall lowest.}
\label{S-tab:instance_seg_12c}
\resizebox{\textwidth}{!}{%
\begin{tabular}{l|ccc|cccccccc}
\hline
\textbf{Model} &
\rotatebox{90}{\textbf{mAP25 (\%)}} &
\rotatebox{90}{\textbf{mAP50 (\%)}} &
\rotatebox{90}{\textbf{mAP (\%)}} &
\rotatebox{90}{\textbf{Building}} &
\rotatebox{90}{\textbf{Pole \& Tower}} &
\rotatebox{90}{\textbf{Wire-Structure Conn.}} &
\rotatebox{90}{\textbf{Vehicle}} &
\rotatebox{90}{\textbf{Board}} &
\rotatebox{90}{\textbf{Fencing}} &
\rotatebox{90}{\textbf{Human \& Animal}} &
\rotatebox{90}{\textbf{Road Side Object}} \\
\hline
\multicolumn{12}{l}{\textbf{Supervised Learning -- 9ch (XYZ + Partial RGB + normals)}} \\
\hline
PTv3 \cite{wu2024point} & 42.07 & 35.13 & 25.78 & 23.71 & 28.57 & \underline{46.14} & \underline{70.62} & 41.98 & 0.13 & 54.04 & 15.87 \\
LitePT-L \cite{yuelitept2026} & \underline{43.73} & \underline{36.76} & \underline{26.90} & \underline{34.03} & \cellcolor{bestgreen}\underline{\textbf{39.15}} & 30.54 & 69.43 & \underline{45.87} & \underline{3.86} & \cellcolor{bestgreen}\underline{\textbf{54.84}} & \underline{16.33} \\
\hline
\multicolumn{12}{l}{\textbf{Linear Probe (9ch: XYZ + Partial RGB + normals)}} \\
\hline
Sonata(L) \cite{wu2025sonata} & 23.28 & 14.66 & 7.72 & \underline{20.41} & 2.33 & \underline{13.04} & 49.87 & 6.72 & \underline{2.96} & 19.27 & 2.63 \\
Concerto(O)          & \textcolor{red}{11.86} & \textcolor{red}{4.07} & \textcolor{red}{1.66} & \textcolor{red}{4.41} & \textcolor{red}{1.83} & \textcolor{red}{0.00} & \textcolor{red}{2.35} & \underline{8.39} & \textcolor{red}{0.01} & \textcolor{red}{14.86} & \textcolor{red}{0.75} \\
Concerto(L) \cite{zhang2025concerto} & 23.91 & 14.22 & 7.80 & 16.25 & 8.12 & 9.34 & 48.29 & \textcolor{red}{3.42} & 0.11 & \underline{26.68} & 1.52 \\
Utonia(L) \cite{zhang2026utonia} & \underline{26.25} & \underline{15.27} & \underline{8.15} & 17.06 & \underline{9.80} & 7.21 & \underline{49.98} & 8.22 & 0.31 & 25.75 & \underline{3.86} \\
\hline
\multicolumn{12}{l}{\textbf{Decoder Probe (9ch: XYZ + Partial RGB + normals)}} \\
\hline
Sonata(L)            & 37.90 & 29.59 & 20.62 & 18.25 & 19.17 & 26.08 & 66.22 & 45.08 & 1.40 & 46.88 & 13.66 \\
Concerto(O)          & 36.79 & 27.46 & 19.02 & 22.42 & 20.92 & 15.36 & 64.51 & 40.35 & 2.39 & 42.41 & 11.33 \\
Concerto(L)          & 39.85 & 31.88 & 22.75 & 20.06 & 22.99 & \underline{32.77} & 67.24 & 46.09 & 2.71 & 49.12 & 14.07 \\
Utonia(L)            & \underline{43.87} & \underline{35.75} & \underline{25.48} & \underline{41.72} & \underline{26.39} & 28.86 & \underline{69.55} & \underline{47.87} & \underline{5.16} & \underline{51.55} & \underline{14.89} \\
\hline
\multicolumn{12}{l}{\textbf{Fine-Tuning (9ch: XYZ + Partial RGB + normals)}} \\
\hline
Sonata(L)            & 44.73 & 38.22 & 28.47 & 28.14 & \underline{35.22} & 43.97 & 71.92 & 50.58 & 3.68 & \underline{53.48} & 18.74 \\
Concerto(O)          & 42.30 & 36.21 & 26.39 & 34.82 & 31.90 & 38.96 & 72.32 & 46.66 & 0.09 & 47.89 & 17.08 \\
Concerto(L)          & 44.37 & 37.06 & 26.68 & 24.73 & 30.96 & 37.30 & \cellcolor{bestgreen}\underline{\textbf{73.68}} & 51.35 & \cellcolor{bestgreen}\underline{\textbf{6.32}} & 53.12 & \cellcolor{bestgreen}\underline{\textbf{19.00}} \\
Utonia(L)            & \cellcolor{bestgreen}\underline{\textbf{50.56}} & \cellcolor{bestgreen}\underline{\textbf{44.07}} & \cellcolor{bestgreen}\underline{\textbf{33.09}} & \cellcolor{bestgreen}\underline{\textbf{53.70}} & 33.46 & \cellcolor{bestgreen}\underline{\textbf{59.47}} & 72.77 & \cellcolor{bestgreen}\underline{\textbf{55.67}} & 5.43 & 53.27 & 18.75 \\
\hline
\end{tabular}%
}%
\end{table*}

\begin{table*}[!tp]
\centering
\scriptsize
\setlength{\tabcolsep}{4pt}
\renewcommand{\arraystretch}{1.1}
\caption{\textbf{Instance segmentation results (21-class)} on BhopalMLS using PTv3 and LitePT-L (trained from scratch) and the pretrained foundation models Sonata, Concerto, and Utonia with LP, DP, and FT. All models use 9ch input. mAP, mAP50, and mAP25 are computed over 17 thing classes at IoU thresholds averaged over 50:5:95, 50\%, and 25\%, respectively; 4 stuff classes (Ground, Vegetation, Road, Wire) are excluded from instance evaluation. Per-class values are AP50 (\%). Sonata(L), Concerto(L) and Utonia(L) = Large (general-purpose pretrained), Concerto(O) = outdoor-pretrained. \colorbox{bestgreen}{\textbf{Bold green}} = overall best per column; \underline{underline} = best within category; \textcolor{red}{red} = overall lowest.}
\label{S-tab:instance_seg_21c}
\resizebox{\textwidth}{!}{%
\begin{tabular}{l|ccc|ccccccccccccccccc}
\hline
\textbf{Model} &
\rotatebox{90}{\textbf{mAP25 (\%)}} &
\rotatebox{90}{\textbf{mAP50 (\%)}} &
\rotatebox{90}{\textbf{mAP (\%)}} &
\rotatebox{90}{\textbf{2-Wheeler}} &
\rotatebox{90}{\textbf{3-Wheeler}} &
\rotatebox{90}{\textbf{4-Wheeler}} &
\rotatebox{90}{\textbf{Hand Cart}} &
\rotatebox{90}{\textbf{Human}} &
\rotatebox{90}{\textbf{Commercial Building}} &
\rotatebox{90}{\textbf{Building (other)}} &
\rotatebox{90}{\textbf{Electric Pole}} &
\rotatebox{90}{\textbf{Street Light}} &
\rotatebox{90}{\textbf{Pole (other)}} &
\rotatebox{90}{\textbf{Board}} &
\rotatebox{90}{\textbf{Fencing}} &
\rotatebox{90}{\textbf{Road Side Furniture}} &
\rotatebox{90}{\textbf{Shop Shade}} &
\rotatebox{90}{\textbf{Mannequin}} &
\rotatebox{90}{\textbf{Stairs}} &
\rotatebox{90}{\textbf{Road Side Object (other)}} \\
\hline
\multicolumn{21}{l}{\textbf{Supervised Learning -- 9ch (XYZ + Partial RGB + normals)}} \\
\hline
PTv3 \cite{wu2024point} & \underline{39.06} & \underline{33.07} & \underline{24.31} & 68.43 & \underline{85.20} & \underline{59.22} & \underline{55.28} & 52.55 & \underline{34.00} & \underline{6.69} & \underline{50.79} & 25.78 & \underline{18.92} & \underline{43.58} & \underline{1.57} & \underline{19.22} & 24.43 & \cellcolor{bestgreen}\underline{\textbf{11.93}} & \underline{2.08} & 2.43 \\
LitePT-L \cite{yuelitept2026} & 33.26 & 25.81 & 18.58 & \underline{70.65} & 73.15 & 42.54 & 50.67 & \cellcolor{bestgreen}\underline{\textbf{55.02}} & \textcolor{red}{0.00} & \textcolor{red}{0.15} & 20.94 & \cellcolor{bestgreen}\underline{\textbf{39.06}} & 2.32 & 28.94 & 1.33 & 15.00 & \underline{24.92} & 7.25 & 1.72 & \underline{5.17} \\
\hline
\multicolumn{21}{l}{\textbf{Linear Probe (9ch: XYZ + Partial RGB + normals)}} \\
\hline
Sonata(L) \cite{wu2025sonata} & 24.37 & 16.07 & 9.46 & 52.38 & 52.75 & \underline{46.74} & 32.08 & 19.80 & \underline{40.76} & \underline{1.41} & 3.88 & \textcolor{red}{2.40} & 3.35 & 2.83 & \underline{2.87} & 2.52 & \underline{7.99} & 0.40 & \underline{1.09} & \textcolor{red}{0.00} \\
Concerto(O)          & \textcolor{red}{11.44} & \textcolor{red}{4.44} & \textcolor{red}{1.73} & \textcolor{red}{1.03} & \textcolor{red}{13.00} & \textcolor{red}{7.57} & \textcolor{red}{11.12} & \textcolor{red}{1.08} & 15.87 & 0.75 & \textcolor{red}{2.23} & 4.70 & \textcolor{red}{0.94} & \underline{11.22} & 0.30 & \textcolor{red}{0.07} & 5.56 & \textcolor{red}{0.02} & \textcolor{red}{0.00} & \textcolor{red}{0.00} \\
Concerto(L) \cite{zhang2025concerto} & 23.74 & 15.93 & 9.29 & 44.96 & 56.66 & 44.05 & 32.46 & \underline{27.91} & 32.92 & 0.44 & 11.64 & \underline{7.32} & 3.02 & \textcolor{red}{2.65} & \textcolor{red}{0.00} & \underline{4.06} & \textcolor{red}{0.54} & \underline{2.22} & \textcolor{red}{0.00} & \textcolor{red}{0.00} \\
Utonia(L) \cite{zhang2026utonia} & \underline{25.89} & \underline{17.88} & \underline{10.46} & \underline{53.71} & \underline{63.78} & 46.03 & \underline{34.35} & 27.74 & 37.59 & 0.55 & \underline{13.09} & 3.76 & \underline{5.17} & 7.61 & \textcolor{red}{0.00} & 3.41 & 6.71 & 0.49 & \textcolor{red}{0.00} & \underline{0.07} \\
\hline
\multicolumn{21}{l}{\textbf{Decoder Probe (9ch: XYZ + Partial RGB + normals)}} \\
\hline
Sonata(L)            & 36.79 & 27.40 & 19.01 & 63.20 & 67.26 & 59.58 & 46.04 & 44.61 & 27.43 & 3.39 & 33.66 & 23.68 & 5.33 & 42.46 & \cellcolor{bestgreen}\underline{\textbf{7.80}} & 12.64 & 21.81 & 5.65 & \underline{1.21} & \textcolor{red}{0.00} \\
Concerto(O)          & 34.58 & 27.12 & 18.50 & 64.43 & 64.56 & 60.60 & 44.65 & 40.82 & 34.31 & 2.94 & 32.66 & \underline{29.16} & \underline{13.15} & 36.02 & 1.74 & 9.33 & 15.30 & \underline{6.58} & 0.31 & \underline{4.40} \\
Concerto(L)          & 35.40 & 28.19 & 20.59 & 65.51 & 71.69 & 61.92 & 47.64 & 45.71 & 35.18 & 1.98 & 34.22 & 22.66 & 11.15 & 43.15 & 4.22 & 12.10 & 18.47 & 1.32 & 1.01 & 1.31 \\
Utonia(L)            & \underline{40.13} & \underline{32.67} & \underline{24.14} & \underline{67.64} & \underline{78.48} & \underline{65.72} & \underline{51.32} & \underline{51.75} & \underline{54.35} & \underline{19.39} & \underline{35.24} & 27.87 & 10.52 & \underline{46.49} & 1.79 & \underline{14.84} & \underline{28.16} & 1.04 & 0.33 & 0.50 \\
\hline
\multicolumn{21}{l}{\textbf{Fine-Tuning (9ch: XYZ + Partial RGB + normals)}} \\
\hline
Sonata(L)            & 40.28 & 34.41 & 26.21 & 71.14 & 83.24 & 66.69 & 56.64 & 50.36 & 38.33 & 7.75 & \cellcolor{bestgreen}\underline{\textbf{53.35}} & 23.82 & 21.73 & 49.24 & 3.32 & 19.67 & 28.43 & 7.92 & \cellcolor{bestgreen}\underline{\textbf{2.60}} & 0.81 \\
Concerto(O)          & 40.97 & 35.47 & 26.89 & 72.10 & 81.33 & 72.22 & 57.42 & 50.27 & 43.06 & 10.50 & 42.43 & 31.10 & 25.00 & 49.82 & 0.36 & 21.84 & \cellcolor{bestgreen}\underline{\textbf{31.41}} & 8.17 & 2.41 & 3.46 \\
Concerto(L)          & 42.13 & 36.71 & 28.43 & 71.20 & \cellcolor{bestgreen}\underline{\textbf{89.46}} & 71.11 & 61.24 & 52.85 & 40.73 & 11.12 & 50.41 & \underline{33.60} & 24.53 & 52.83 & 2.53 & 22.99 & 25.74 & \underline{8.31} & 1.51 & 3.94 \\
Utonia(L)            & \cellcolor{bestgreen}\underline{\textbf{46.52}} & \cellcolor{bestgreen}\underline{\textbf{39.83}} & \cellcolor{bestgreen}\underline{\textbf{30.63}} & \cellcolor{bestgreen}\underline{\textbf{73.23}} & 83.07 & \cellcolor{bestgreen}\underline{\textbf{73.57}} & \cellcolor{bestgreen}\underline{\textbf{62.55}} & \underline{53.56} & \cellcolor{bestgreen}\underline{\textbf{62.44}} & \cellcolor{bestgreen}\underline{\textbf{31.11}} & 43.79 & 30.36 & \cellcolor{bestgreen}\underline{\textbf{31.38}} & \cellcolor{bestgreen}\underline{\textbf{55.76}} & \underline{7.52} & \cellcolor{bestgreen}\underline{\textbf{24.43}} & 31.38 & 5.84 & 0.97 & \cellcolor{bestgreen}\underline{\textbf{6.12}} \\
\hline
\end{tabular}%
}%
\end{table*}

\subsection{Qualitative Instance Segmentation Results}

\begin{figure*}[!t]
    \centering
    \includegraphics[width=0.95\textwidth, height=0.85\textheight, keepaspectratio]{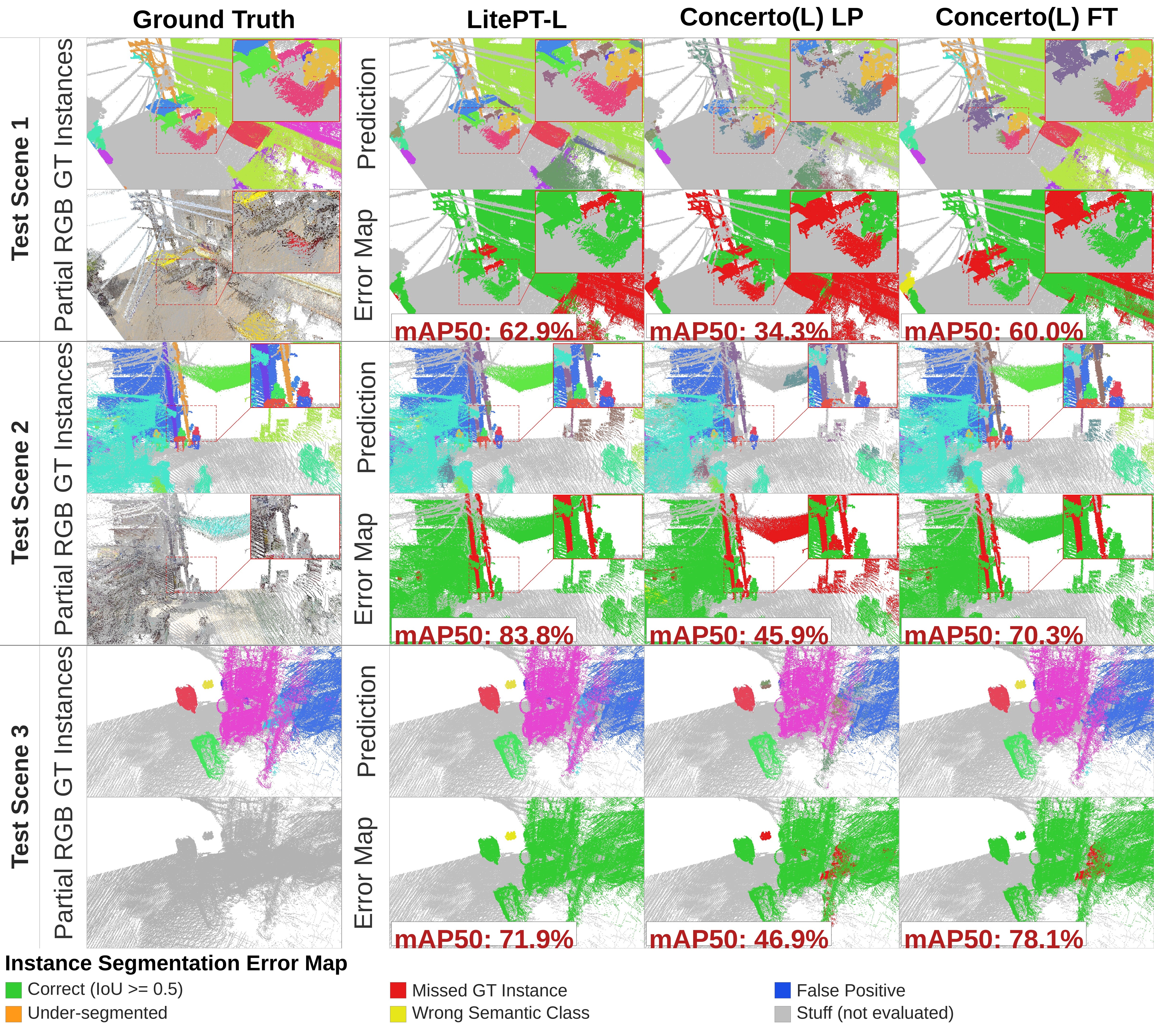}
    \caption{\textbf{Qualitative instance segmentation results (12-class)} on three BhopalMLS test scenes (same scenes as Fig.~\ref{S-fig:seg-predictions}). Left column: ground truth instances and partial RGB (gray = no RGB coverage). Right columns: predicted instances and error maps for LitePT-L, Concerto(L)~LP, and Concerto(L)~FT, with per-scene mAP50 annotated on each error map cell. Instance colors are matched across GT and predictions via IoU ($\geq$~0.5). Error map legend: green = correct (IoU~$\geq$~0.5), red = missed GT, blue = false positive, orange = under-segmented, yellow = wrong class, gray = stuff.}
    \label{S-fig:instance-predictions-12c}
\end{figure*}

Figure~\ref{S-fig:instance-predictions-12c} presents qualitative instance segmentation results under the 12-class taxonomy for the same three representative test scenes used in the semantic segmentation evaluation (Fig.~\ref{S-fig:seg-predictions}). For each scene, we show: (i)~ground truth instance labels and partial RGB point cloud (left column, separated by a vertical divider); (ii)~predicted instance masks from LitePT-L (trained from scratch), Concerto(L)~LP, and Concerto(L)~FT (top row); and (iii)~corresponding error maps with per-scene mAP50 annotations (bottom row). The inclusion of Concerto(L)~LP alongside FT directly visualizes the LP$\rightarrow$FT improvement reported in Table~\ref{S-tab:instance_seg_12c}. Instance colors are consistent across ground truth and predictions: matched instances (IoU~$\geq$~0.5) share the same color, enabling direct visual comparison of segmentation quality. Stuff classes (Ground, Vegetation, Road, Wire) are rendered in gray and excluded from instance evaluation.

The error maps use the following color coding: \textcolor{green!70!black}{green} = correctly matched instance (IoU~$\geq$~0.5), \textcolor{red}{red} = missed GT instance (no prediction match), \textcolor{blue}{blue} = false positive prediction (no GT match), \textcolor{orange}{orange} = under-segmented (multiple GT instances merged into one prediction), \textcolor{yellow!80!black}{yellow} = wrong semantic class, and gray = stuff classes (not evaluated). Note that this instance-level error map differs from the binary semantic segmentation error map in Fig.~\ref{S-fig:seg-predictions}, which uses only two categories (gray = correct, red = misclassified). The richer six-category coding here reflects the additional failure modes specific to instance segmentation, such as under-segmentation and false positive detections. All three models successfully segment large, geometrically distinct objects such as vehicles and buildings, while thin structures (poles, wire-structure connectors, fencing) and small objects (boards, road side objects) remain challenging, consistent with the quantitative trends in Table~\ref{S-tab:instance_seg_12c}. The LP$\rightarrow$FT improvement is particularly visible in the error maps, where LP shows substantially more missed instances (red) and false positives (blue).

\begin{figure*}[!t]
    \centering
    \includegraphics[width=0.95\textwidth, height=0.85\textheight, keepaspectratio]{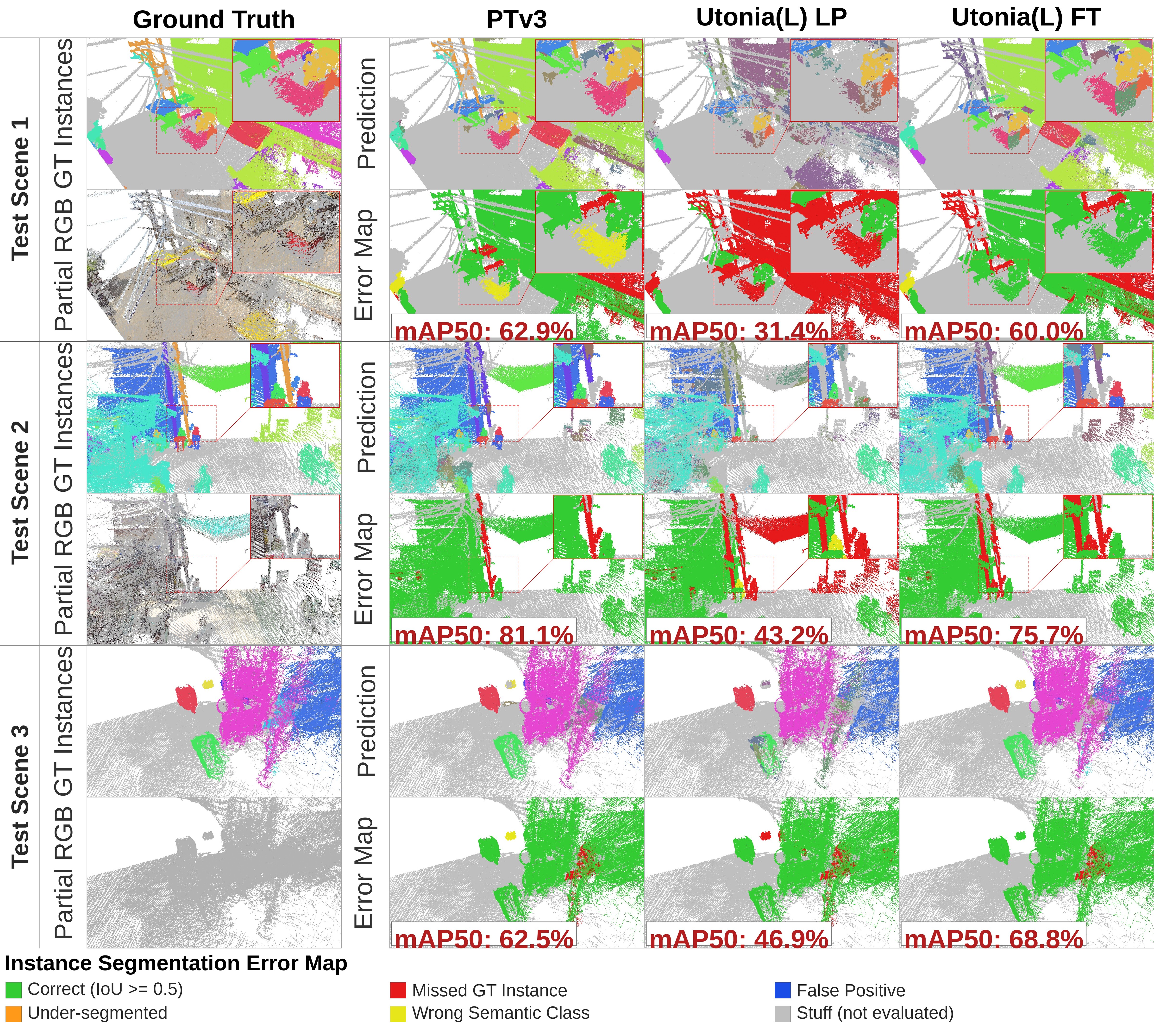}
    \caption{\textbf{Qualitative instance segmentation results (21-class)} on three BhopalMLS test scenes. Same 2-row layout as Fig.~\ref{S-fig:instance-predictions-12c}, but for the 21-class fine-grained taxonomy (17 thing classes) with the supervised PTv3 baseline and Utonia(L)~LP / Utonia(L)~FT (per-scene mAP50 annotated on each error map). The finer granularity introduces additional wrong-class errors (yellow) at vehicle and building boundaries where sub-type distinction is required.}
    \label{S-fig:instance-predictions-21c}
\end{figure*}

Figure~\ref{S-fig:instance-predictions-21c} presents the same three test scenes evaluated under the 21-class fine-grained taxonomy, which expands the 8 thing classes to 17 fine-grained categories including India-specific classes such as 2-Wheeler, 3-Wheeler, Hand Cart, Shop Shade, and Mannequin. The finer class granularity introduces additional challenges: vehicles that were grouped as a single class in the 12-class setting must now be distinguished into 2-Wheeler, 3-Wheeler, 4-Wheeler, and Hand Cart, leading to increased wrong-class errors (yellow) at vehicle boundaries. Similarly, the Building class is split into Commercial Building and Building (other), requiring models to distinguish structural subtypes. Despite these challenges, Utonia(L)~FT maintains the strongest instance delineation (39.83\% mAP50, Table~\ref{S-tab:instance_seg_21c}), while the supervised PTv3 baseline remains competitive on well-separated objects and Utonia(L)~LP exhibits more missed and merged instances, consistent with the per-class AP50 trends in Table~\ref{S-tab:instance_seg_21c}.

\section{Feature Visualization and Prediction Analysis}

We examine the learned representations and prediction behavior of the fine-tuned foundation models: first through PCA feature visualizations at the instance and scene levels, then through a confusion-matrix analysis of their systematic misclassifications.

\subsection{Instance-Level Feature Visualization}

\begin{figure*}[!tp]
    \centering
    \includegraphics[width=\textwidth, height=0.70\textheight, keepaspectratio]{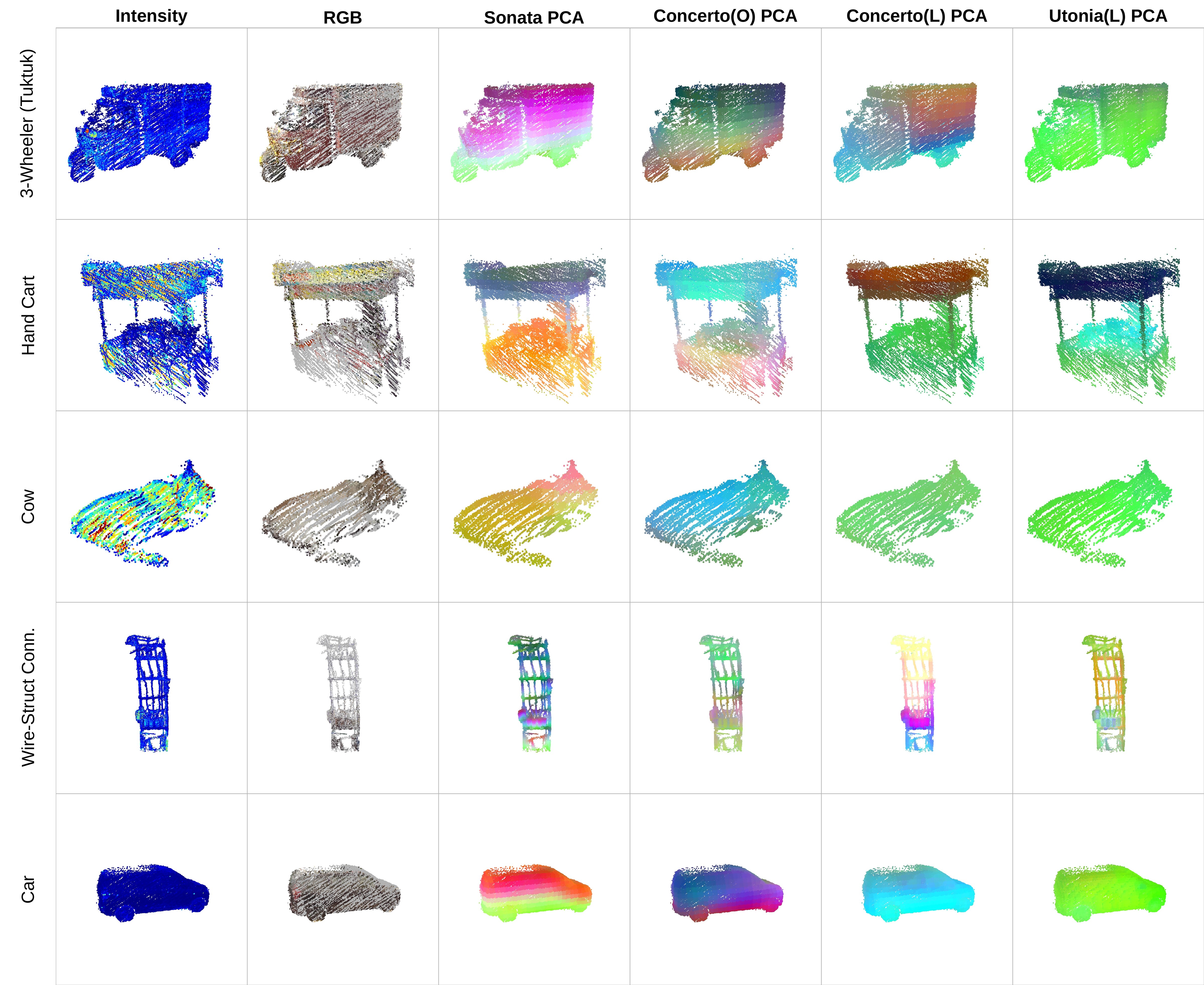}
    \caption{\textbf{Instance-level PCA feature visualization.} Rows: 3-Wheeler (Tuktuk), Hand Cart, Cow, Wire-Structure Connector (India-specific), and Car (universal). Columns: raw intensity, raw RGB (gray = missing coverage), and PCA projections from Sonata(L), Concerto(O), Concerto(L), and Utonia(L). Similar colors indicate similar learned features.}
    \label{S-fig:pca-instances}
\end{figure*}

To qualitatively assess the learned representations of pretrained foundation models, we visualize their intermediate features on individual object instances using PCA. Figure~\ref{S-fig:pca-instances} shows five selected classes: 3-Wheeler (Tuktuk), Hand Cart, Cow, Wire-Structure Connector (India-specific), and Car (a universal class present in most LiDAR datasets), rendered with raw intensity, raw RGB, and PCA projections from Sonata(L)~\cite{wu2025sonata} (Large), Concerto(O)~\cite{zhang2025concerto} (outdoor-pretrained), Concerto(L) (general-purpose pretrained), and Utonia(L) (general-purpose pretrained). Features are extracted from the frozen encoder and projected from high-dimensional space (1232D for Sonata, 1728D for Concerto, 1224D for Utonia) to RGB via PCA (q=6, q=9, or q=12). Points with similar colors share similar learned features, revealing how each model groups geometric and semantic structure. The inclusion of Car enables direct comparison of foundation model representations on familiar vs.\ novel object categories.

\subsection{Scene-Level Predictions and Features}

\begin{figure*}[!tp]
    \centering
    \includegraphics[width=\textwidth, height=0.85\textheight, keepaspectratio]{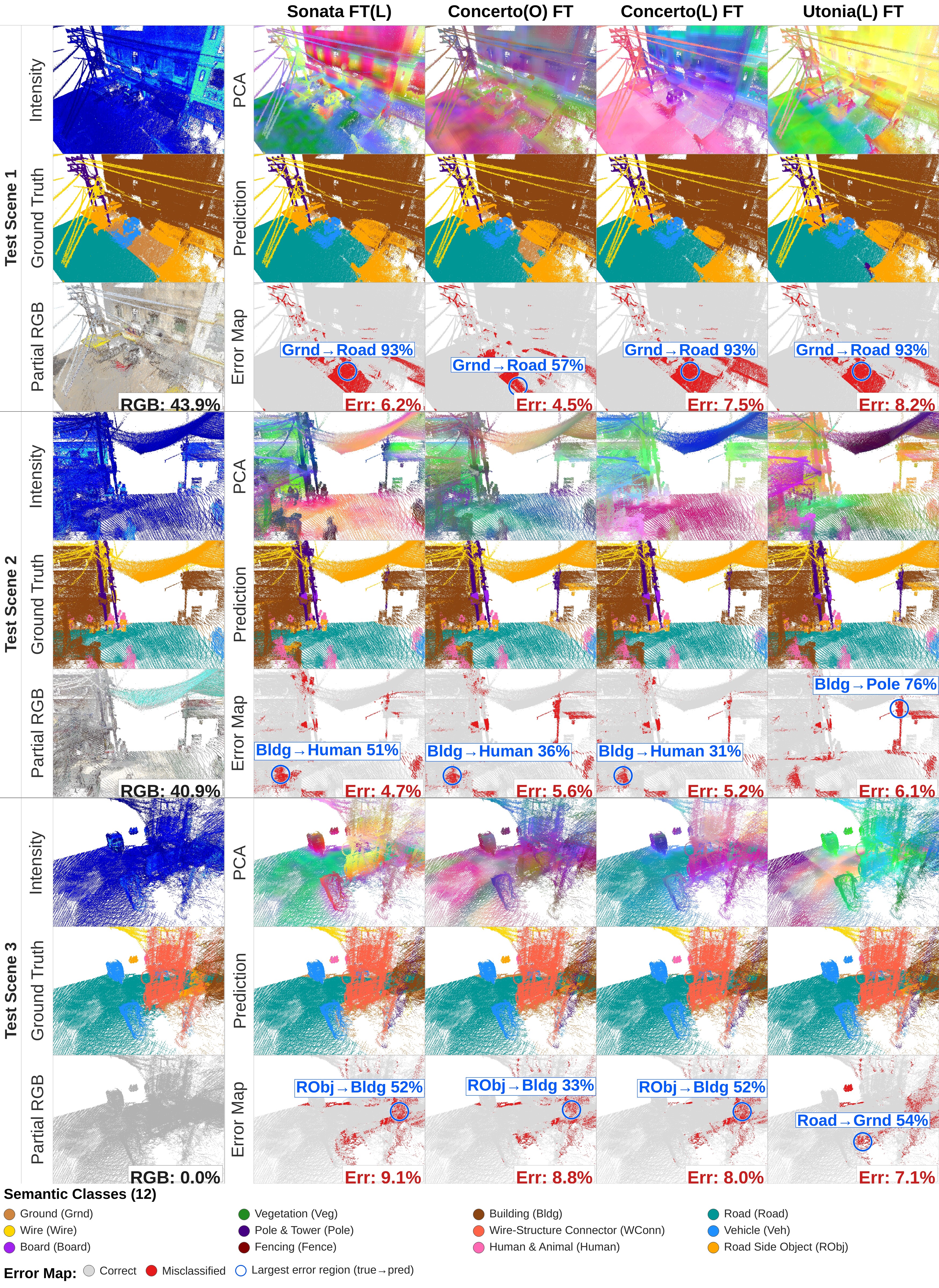}
    \caption{\textbf{Scene-level predictions, PCA features, and error maps} on three BhopalMLS test scenes. Left column: inputs (intensity, ground truth, partial RGB). Right columns: per-model PCA features, predictions, and error maps (gray = correct, red = misclassified). On each error map a blue circle marks the largest error region (most misclassified points), labeled with that region's majority true$\rightarrow$predicted class and local error rate; here the percentage is the fraction of that region's points that are misclassified (\eg, ``Veh$\rightarrow$Road 93\%'' = 93\% of the region's points are truly Vehicle but predicted as Road; class abbreviations in the legend). The overall per-model error rate (\%) is shown at the bottom-right of each cell.}
    \label{S-fig:seg-predictions}
\end{figure*}

Figure~\ref{S-fig:seg-predictions} presents qualitative results for three representative test scenes from the BhopalMLS dataset. For each scene, we show: (i)~the raw LiDAR intensity, ground truth semantic labels, and partial RGB point cloud where available (left column); (ii)~PCA projections of frozen encoder features from Sonata(L)~\cite{wu2025sonata}, Concerto(O)~\cite{zhang2025concerto}, Concerto(L), and Utonia(L) (top row); (iii)~the corresponding 12-class semantic segmentation predictions from the fine-tuned (FT) models (middle row); and (iv)~binary error maps highlighting misclassified points in red (bottom row).

PCA features are extracted from the frozen foundation model encoders and projected to RGB as above; points with similar colors share similar learned representations, revealing how each model groups geometric and semantic structure at the scene level.

The three test scenes are: \textbf{Test Scene~1} (2.56M points, 9 of 12 classes present, 44\% RGB coverage, oblique aerial view), \textbf{Test Scene~2} (1.39M points, 8 of 12 classes present, 41\% RGB coverage, street-level view), and \textbf{Test Scene~3} (1.06M points, 10 of 12 classes present, no RGB coverage). Among the four fine-tuned models, Concerto(O)~FT achieves the lowest error rate on Test Scene~1 (4.5\%), Sonata(L)~FT on Test Scene~2 (4.7\%), and Utonia(L)~FT on Test Scene~3 (7.1\%). Test Scene~3 (no RGB coverage) proves the most challenging, with error rates of 7--9\% across models.

\subsection{Confusion Matrix Analysis}

\begin{figure*}[!tp]
    \centering
    \includegraphics[width=\linewidth]{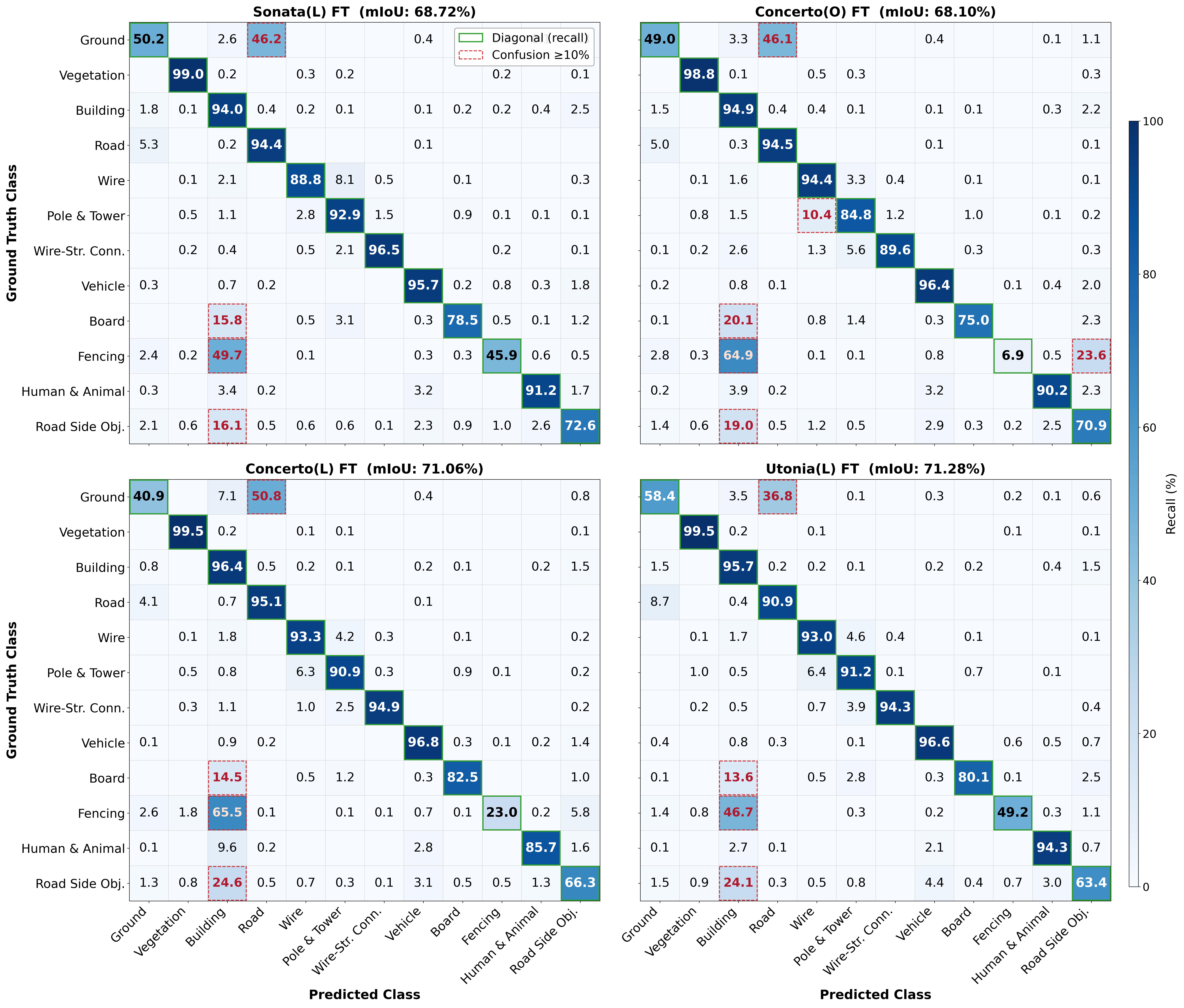}
    \caption{\textbf{Row-normalized confusion matrices (12-class)} for four fine-tuned foundation models on BhopalMLS test set (486 scenes, 201M+ points). Each row sums to 100\%. Diagonal entries (green border) show per-class recall (\%); off-diagonal entries $\geq$10\% (red dashed border) highlight significant confusions. mIoU computed over all 12 classes.}
    \label{S-fig:confusion-matrices}
\end{figure*}

Figure~\ref{S-fig:confusion-matrices} presents row-normalized confusion matrices for the four fine-tuned foundation models on the BhopalMLS test set (Path~07--08, 486 scenes, 201M+ points). Each cell shows the percentage of ground truth points (row) predicted as a given class (column), revealing systematic misclassification patterns.

Across all four models, the dominant confusions occur at semantically and geometrically ambiguous boundaries: \textbf{Fencing$\rightarrow$Building} (47--65\% of Fencing points misclassified as Building), \textbf{Ground$\rightarrow$Road} (37--51\%), and \textbf{Road Side Object$\rightarrow$Building} (16--25\%). These patterns are consistent with the qualitative error maps in Fig.~\ref{S-fig:seg-predictions} and confirm that boundary confusion between adjacent structures is the primary failure mode.

Notably, Utonia(L)~FT achieves the highest Fencing recall (49.2\%), followed by Sonata(L)~FT (45.9\%), Concerto(L)~FT (23.0\%), and Concerto(O)~FT (6.9\%). Concerto(O)~FT additionally exhibits a notable Fencing$\rightarrow$Road Side Object confusion (23.6\%), far exceeding Concerto(L) (5.8\%), Utonia(L) (1.1\%), and Sonata(L) (0.5\%). Across all models, Ground recall (41--58\%) is limited by its dominant misclassification as Road (37--51\%), highlighting the geometric similarity between unpaved ground surfaces and road boundaries in dense Indian urban settings.

\section{Cross-Region Generalization: Per-Class Breakdowns}

This section reports the full per-class IoU breakdown behind the cross-region summary metrics in Table~\ref{tab:cross_region_1cm}, across SK, BM, KM, TR, and PL. It also adds the outdoor-pretrained Concerto(O) variant on the SK-involved pairs and BM$\rightarrow$PL, so the Concerto(L) vs.\ Concerto(O) contrast isolates the effect of pretraining strategy on the structured$\leftrightarrow$unstructured domain gap.

\definecolor{bmsrcgreen}{rgb}{0.85,0.95,0.85}
\definecolor{tgtsrcred}{rgb}{1.00,0.88,0.88}

Table~\ref{S-tab:cross_city_full} reports all values at native resolution (FT-9ch) for Concerto(L)~\cite{zhang2025concerto} (Large, general-purpose pretrained), Concerto(O) (outdoor-pretrained), and Utonia(L)~\cite{zhang2026utonia} across five MLS datasets: \textbf{SemanticKITTI (SK)}~\cite{behley2019semantickitti}, \textbf{BhopalMLS (BM)}, \textbf{Ke-MLS (KM)}~\cite{kumar2026ke}, \textbf{Toronto-3D (TR)}~\cite{tan2020toronto}, and \textbf{Paris-Lille-3D (PL)}~\cite{roynard2018paris}. Each model is fine-tuned on a source dataset (FT-9ch) and evaluated on the target at native resolution, with all labels mapped to the unified 12-class BhopalMLS taxonomy. Values marked \textbf{0.00*} indicate classes absent in the source or target dataset and are excluded from metric computation. Concerto(L) and Utonia(L) are reported on all 13 pairs (5 within-domain, 2 intra-Indian, 6 inter-region); Concerto(O) on SK$\rightarrow$SK / SK$\rightarrow$BM / BM$\rightarrow$SK / BM$\rightarrow$PL.

\begin{table*}[!t]
\centering
\footnotesize
\setlength{\tabcolsep}{4pt}
\renewcommand{\arraystretch}{1.1}
\caption{\textbf{Cross-region generalization} (FT-9ch, native resolution) of Concerto(L)/(O) and Utonia(L) across SK~\cite{behley2019semantickitti}, BM, KM~\cite{kumar2026ke}, TR~\cite{tan2020toronto}, PL~\cite{roynard2018paris}. \textbf{0.00*}: class absent in source or target (excluded from metrics). mIoU/mAcc are over the \emph{common-class} intersection of the source and target BM-12 taxonomies, with $n_c$ classes per target dataset: \textbf{BM:~12}, \textbf{KM:~12}, \textbf{PL:~11} (drops Wire-Structure Connector), \textbf{SK:~9} (drops Wire, Wire-Structure Connector, Road Side Object), \textbf{TR:~7}. When the source has fewer BM-12 classes than the target, $n_c$ is bounded by the source (e.g.\ SK$\to$BM uses $n_c{=}9$). Per-class IoU (\%) is reported in BM-12 taxonomy for all rows. BM$\to$BM rows match Table~\ref{tab:bhopal_iou_benchmark_style_rotated_vline}.}
\label{S-tab:cross_city_full}
\resizebox{\textwidth}{!}{%
\begin{tabular}{l|ccc|cccccccccccc}
\hline
\textbf{Model (train~$\rightarrow$~test)} &
\rotatebox{90}{\textbf{OA (\%)}} &
\rotatebox{90}{\textbf{mAcc (\%)}} &
\rotatebox{90}{\textbf{mIoU (\%)}} &
\rotatebox{90}{\textbf{Board}} &
\rotatebox{90}{\textbf{Building}} &
\rotatebox{90}{\textbf{Fencing}} &
\rotatebox{90}{\textbf{Ground}} &
\rotatebox{90}{\textbf{Human\&Animal}} &
\rotatebox{90}{\textbf{Pole\&Tower}} &
\rotatebox{90}{\textbf{Road}} &
\rotatebox{90}{\textbf{Road Side Obj.}} &
\rotatebox{90}{\textbf{Vegetation}} &
\rotatebox{90}{\textbf{Vehicle}} &
\rotatebox{90}{\textbf{Wire}} &
\rotatebox{90}{\textbf{Wire-Structure Conn.}} \\
\hline
\multicolumn{16}{l}{\textbf{Within-domain (source $=$ target)}} \\
\hline
Concerto(L) (BM~$\rightarrow$~BM) & 93.42 & 80.44 & 71.06 & 67.72 & 93.71 & 20.14 & 25.41 & 69.88 & 72.48 & 87.11 & 53.07 & 97.67 & 92.34 & 87.91 & 85.26 \\
Utonia(L)   (BM~$\rightarrow$~BM) & 92.69 & 83.89 & 71.28 & 67.36 & 93.38 & 38.79 & 26.27 & 64.24 & 68.61 & 85.66 & 51.51 & 97.41 & 91.54 & 87.53 & 83.03 \\
Concerto(L) (KM~$\rightarrow$~KM) & 92.40 & 74.79 & 67.58 & 13.83 & 83.47 & 83.66 & 68.43 & 76.87 & 76.15 & 86.04 & 0.00 & 96.59 & 90.15 & 86.69 & 49.05 \\
Utonia(L)   (KM~$\rightarrow$~KM) & 92.82 & 77.05 & 66.66 & 13.54 & 84.27 & 86.85 & 69.26 & 83.28 & 72.43 & 87.33 & 0.00 & 96.48 & 94.47 & 87.55 & 24.43 \\
Concerto(L) (TR~$\rightarrow$~TR) & 99.03 & 93.25 & 86.96 & \textbf{0.00*} & 93.71 & 50.18 & \textbf{0.00*} & \textbf{0.00*} & 85.41 & 99.41 & \textbf{0.00*} & 97.45 & 95.61 & 86.94 & \textbf{0.00*} \\
Utonia(L)   (TR~$\rightarrow$~TR) & 99.12 & 93.25 & 88.38 & \textbf{0.00*} & 95.09 & 59.90 & \textbf{0.00*} & \textbf{0.00*} & 84.56 & 99.43 & \textbf{0.00*} & 97.83 & 95.10 & 86.78 & \textbf{0.00*} \\
Concerto(L) (PL~$\rightarrow$~PL) & 94.14 & 92.03 & 83.43 & 67.14 & 99.07 & 55.12 & 71.69 & 96.79 & 90.37 & 89.71 & 77.50 & 82.32 & 98.94 & 89.11 & \textbf{0.00*} \\
Utonia(L)   (PL~$\rightarrow$~PL) & 94.89 & 92.11 & 84.77 & 67.70 & 98.86 & 66.38 & 74.90 & 90.60 & 90.45 & 91.32 & 79.03 & 80.85 & 99.48 & 92.89 & \textbf{0.00*} \\
Concerto(L) (SK~$\rightarrow$~SK) & 92.47 & 83.67 & 75.09 & 39.19 & 78.67 & 52.14 & 86.43 & 83.35 & 59.85 & 92.98 & \textbf{0.00*} & 86.88 & 96.33 & \textbf{0.00*} & \textbf{0.00*} \\
Concerto(O) (SK~$\rightarrow$~SK) & 91.79 & 82.62 & 74.65 & 41.83 & 74.83 & 51.95 & 85.28 & 83.72 & 59.90 & 92.76 & \textbf{0.00*} & 85.54 & 96.04 & \textbf{0.00*} & \textbf{0.00*} \\
Utonia(L)   (SK~$\rightarrow$~SK) & 92.69 & 84.86 & 77.14 & 43.96 & 81.52 & 61.83 & 85.00 & 83.63 & 62.10 & 93.26 & \textbf{0.00*} & 86.13 & 96.85 & \textbf{0.00*} & \textbf{0.00*} \\
\hline
\multicolumn{16}{l}{\textbf{Intra-Indian (BM $\leftrightarrow$ KM)}} \\
\hline
\rowcolor{bmsrcgreen} Concerto(L) (BM~$\rightarrow$~KM) & 75.40 & 66.43 & 49.23 & 7.69 & 65.86 & 24.86 & 12.07 & 76.03 & 60.92 & 71.75 & 3.83 & 75.22 & 85.16 & 78.61 & 28.72 \\
\rowcolor{bmsrcgreen} Utonia(L)   (BM~$\rightarrow$~KM) & 80.99 & 74.94 & 56.78 & 13.44 & 76.63 & 48.29 & 33.49 & 80.09 & 61.05 & 82.09 & 6.68 & 75.18 & 84.23 & 79.12 & 41.12 \\
\rowcolor{tgtsrcred} Concerto(L) (KM~$\rightarrow$~BM) & 76.11 & 51.29 & 32.70 & 5.23 & 84.39 & 23.50 & 11.83 & 20.33 & 28.38 & 34.86 & 0.14 & 56.96 & 61.96 & 47.06 & 17.79 \\
\rowcolor{tgtsrcred} Utonia(L)   (KM~$\rightarrow$~BM) & 78.77 & 53.90 & 33.96 & 3.44 & 87.21 & 21.18 & 12.81 & 21.91 & 18.75 & 40.59 & 1.08 & 58.72 & 73.11 & 55.23 & 13.51 \\
\hline
\multicolumn{16}{l}{\textbf{Inter-region (BM $\leftrightarrow$ \{TR, PL, SK\})}} \\
\hline
\rowcolor{bmsrcgreen} Concerto(L) (BM~$\rightarrow$~TR) & 94.99 & 84.67 & 75.20 & \textbf{0.00*} & 89.62 & 16.06 & \textbf{0.00*} & \textbf{0.00*} & 66.05 & 95.39 & \textbf{0.00*} & 93.98 & 85.86 & 79.44 & \textbf{0.00*} \\
\rowcolor{bmsrcgreen} Utonia(L)   (BM~$\rightarrow$~TR) & 92.95 & 90.36 & 77.27 & \textbf{0.00*} & 91.98 & 11.23 & \textbf{0.00*} & \textbf{0.00*} & 74.16 & 92.25 & \textbf{0.00*} & 94.42 & 94.17 & 82.69 & \textbf{0.00*} \\
\rowcolor{tgtsrcred} Concerto(L) (TR~$\rightarrow$~BM) & 77.04 & 49.52 & 39.11 & \textbf{0.00*} & 76.22 & 0.81 & \textbf{0.00*} & \textbf{0.00*} & 18.83 & 52.34 & \textbf{0.00*} & 61.73 & 17.77 & 46.04 & \textbf{0.00*} \\
\rowcolor{tgtsrcred} Utonia(L)   (TR~$\rightarrow$~BM) & 83.72 & 60.66 & 51.65 & \textbf{0.00*} & 82.94 & 0.00 & \textbf{0.00*} & \textbf{0.00*} & 22.48 & 78.10 & \textbf{0.00*} & 80.94 & 30.38 & 66.71 & \textbf{0.00*} \\
\rowcolor{bmsrcgreen} Concerto(L) (BM~$\rightarrow$~PL) & 82.61 & 67.55 & 48.84 & 20.30 & 95.82 & 21.37 & 12.39 & 19.86 & 25.23 & 75.02 & 59.76 & 60.33 & 96.39 & 50.72 & \textbf{0.00*} \\
\rowcolor{bmsrcgreen} Concerto(O) (BM~$\rightarrow$~PL) & 84.87 & 73.22 & 56.35 & 31.50 & 97.06 & 35.93 & 26.01 & 75.28 & 33.15 & 77.49 & 40.02 & 64.14 & 98.25 & 41.03 & \textbf{0.00*} \\
\rowcolor{bmsrcgreen} Utonia(L)   (BM~$\rightarrow$~PL) & 85.54 & 76.95 & 55.68 & 24.36 & 98.04 & 30.82 & 28.07 & 52.87 & 41.00 & 78.26 & 47.28 & 65.48 & 97.76 & 48.53 & \textbf{0.00*} \\
\rowcolor{tgtsrcred} Concerto(L) (PL~$\rightarrow$~BM) & 73.52 & 45.85 & 31.07 & 0.38 & 84.74 & 20.70 & 9.06 & 38.23 & 14.48 & 17.57 & 12.49 & 58.15 & 67.25 & 18.76 & \textbf{0.00*} \\
\rowcolor{tgtsrcred} Utonia(L)   (PL~$\rightarrow$~BM) & 72.38 & 40.42 & 27.06 & 1.34 & 80.95 & 15.90 & 9.11 & 31.24 & 10.67 & 23.22 & 8.00 & 65.21 & 46.33 & 5.66 & \textbf{0.00*} \\
\rowcolor{bmsrcgreen} Concerto(L) (BM~$\rightarrow$~SK) & 53.61 & 47.23 & 32.07 & 0.69 & 37.10 & 14.12 & 45.07 & 27.96 & 30.82 & 35.11 & \textbf{0.00*} & 25.58 & 72.20 & \textbf{0.00*} & \textbf{0.00*} \\
\rowcolor{bmsrcgreen} Concerto(O) (BM~$\rightarrow$~SK) & 62.56 & 50.99 & 38.65 & 1.87 & 52.02 & 8.68 & 53.32 & 38.05 & 35.56 & 32.56 & \textbf{0.00*} & 44.53 & 81.22 & \textbf{0.00*} & \textbf{0.00*} \\
\rowcolor{bmsrcgreen} Utonia(L)   (BM~$\rightarrow$~SK) & 64.01 & 57.88 & 40.05 & 16.54 & 48.73 & 12.68 & 60.91 & 40.93 & 21.11 & 50.41 & \textbf{0.00*} & 37.49 & 71.63 & \textbf{0.00*} & \textbf{0.00*} \\
\rowcolor{tgtsrcred} Concerto(L) (SK~$\rightarrow$~BM) & 50.18 & 27.89 & 12.29 & 1.07 & 64.29 & 4.12 & 4.44 & 0.80 & 2.27 & 14.49 & \textbf{0.00*} & 9.99 & 9.15 & \textbf{0.00*} & \textbf{0.00*} \\
\rowcolor{tgtsrcred} Concerto(O) (SK~$\rightarrow$~BM) & 16.08 & 19.57 & \phantom{0}3.89 & 0.15 & 17.17 & 0.28 & 5.99 & 0.00 & 4.20 & 0.14 & \textbf{0.00*} & 6.83 & 0.24 & \textbf{0.00*} & \textbf{0.00*} \\
\rowcolor{tgtsrcred} Utonia(L)   (SK~$\rightarrow$~BM) & 77.04 & 55.28 & 35.54 & 0.25 & 81.51 & 23.87 & 8.73 & 22.44 & 1.81 & 59.33 & \textbf{0.00*} & 51.37 & 70.58 & \textbf{0.00*} & \textbf{0.00*} \\
\hline
\end{tabular}%
}%
\end{table*}

This table adds the outdoor-pretrained Concerto(O) variant (absent from Table~\ref{tab:cross_region_1cm}, which reports only Concerto(L) and Utonia(L)), enabling a direct Concerto(L) vs.\ Concerto(O) comparison:
\begin{enumerate}
    \item \textbf{Within-domain (SK$\rightarrow$SK)}: Utonia(L) achieves the strongest within-domain SK transfer (77.14\% mIoU), ahead of Concerto(L) (75.09\%) and Concerto(O) (74.65\%); the Concerto(L)~$>$~Concerto(O) ordering is consistent with the within-dataset advantage of general-purpose pretraining observed in the main paper (Table~4).
    \item \textbf{SK$\rightarrow$BM}: this is the pair on which the pretraining corpus separates the two Concerto variants most sharply, Concerto(L) reaching 12.29\% mIoU against Concerto(O)'s 3.89\%. Both nonetheless collapse from above 74\% on their own domain, so the domain gap between structured and unstructured urban environments is not resolved by pretraining strategy alone. Architecture matters more still: Utonia(L) reaches 35.54\% mIoU, $2.9\times$ Concerto(L) and $9.1\times$ Concerto(O), driven by far stronger \textit{Building} (81.51\% vs.\ 64.29\%), \textit{Vehicle} (70.58\% vs.\ 9.15\%), and \textit{Vegetation} (51.37\% vs.\ 9.99\%) transfer, indicating that the structured$\rightarrow$unstructured gap is narrowed more by the backbone than by the pretraining corpus.
    \item \textbf{BM$\rightarrow$SK}: Here the trend reverses: Concerto(O) (38.65\% mIoU) substantially outperforms Concerto(L) (32.07\%). Outdoor-specific pretraining provides a stronger inductive bias for transferring from complex BhopalMLS scenes to the structured SemanticKITTI domain, particularly for \textit{Vehicle} (81.22\% vs.\ 72.20\%), \textit{Building} (52.02\% vs.\ 37.10\%), and \textit{Vegetation} (44.53\% vs.\ 25.58\%), suggesting that outdoor-specific geometric priors transfer more effectively to structured driving scenes.
    \item \textbf{Within-domain (BM$\rightarrow$BM)}: Concerto(L) achieves 71.06\% mIoU over all 12 BhopalMLS classes ($n_c{=}12$; Table~\ref{tab:bhopal_iou_benchmark_style_rotated_vline}), versus 75.09\% mIoU on SemanticKITTI ($n_c{=}9$). The lower within-domain score on BhopalMLS reflects the greater scene complexity and class diversity of dense, unstructured urban environments.
\end{enumerate}

\section{Factor Attribution for the Transfer Asymmetry}
\label{S-sec:asym_attribution}

Table~\ref{S-tab:cross_city_full} establishes \emph{that} BhopalMLS-trained models
transfer outward better than the reverse. This section isolates \emph{which}
property of the dataset produces that advantage, by retraining under controlled
ablations rather than by correlating with per-class outcomes.

\subsection{Design}

We build four training subsets of the BhopalMLS training split. All four carry
an \emph{identical} labelled-point budget of approximately 135M points (25\% of
the split's 540M labelled points, comparable in scale to Paris-Lille-3D), so they
differ only in \emph{what kind} of data is retained
(Table~\ref{S-tab:asym_design}):

\begin{itemize}
    \item \textbf{(B1) size.} A uniformly random scene subsample. This is the
    control: it removes data volume and nothing else. B2 and B3 are read against
    B1, not against the full model, so that the budget reduction they share is
    already accounted for.
    \item \textbf{(B2) variety.} Each of the 12 parent classes is collapsed to
    its single dominant fine subclass by point count; the remaining subclasses
    become unlabelled and are ignored by the loss. Scene geometry is untouched,
    and scenes are added until the dominant-subclass labelled budget matches B1.
    This removes intra-class variety at constant supervision volume.
    \item \textbf{(B3) geometry.} Scenes are ranked by corridor regularity (wide
    carriageway, low trajectory curvature) and only the most regular are kept.
    This removes irregular street geometry.
    \item \textbf{(B4) subtype.} Identical to B2 except that \emph{Vehicle}
    retains \emph{Car} instead of its dominant subtype. B2 alone cannot
    distinguish "intra-class variety transfers" from the much weaker "we happened
    to retain the one subtype the targets lack", because the dominant
    Vehicle subtype in the training split is the two-wheeler (52\% of its
    Vehicle points) while every target benchmark is car-dominated. B4 retains
    Car (23\% of the same) and changes nothing
    else, walking B2's scene ordering so the two differ only in which subtype's
    points count toward the budget.
\end{itemize}

Each subset is fine-tuned with the recipe, schedule and batch configuration used
for the full model in Table~\ref{tab:bhopal_iou_benchmark_style_rotated_vline}, then evaluated on KM, TR, PL and SK
with the protocol used for Table~\ref{S-tab:cross_city_full}. A factor is judged to
matter if destroying it costs markedly more transfer than removing an equal
quantity of data does.

The full-model row of Table~\ref{S-tab:asym_ablation} is carried over from
Table~\ref{S-tab:cross_city_full} rather than recomputed, so to establish that the
two are directly comparable we also re-evaluated the unablated model on the
ablation hardware through the same wrapper. It reproduces
Table~\ref{S-tab:cross_city_full} to within \textbf{0.08} mIoU points on all three
of KM, TR and PL (56.78/56.71, 77.27/77.25, 55.68/55.63; largest single-class
deviation 0.80). Differences between the full model and the ablated arms are
therefore not an artifact of when or where each model was evaluated.

\begin{table}[!tb]
\centering
\footnotesize
\setlength{\tabcolsep}{4pt}
\renewcommand{\arraystretch}{1.15}
\caption{\textbf{Factor-controlled training subsets.} All four arms hold the
same labelled-point budget, so they are distinguished only by what kind of data
survives. B2 holds \emph{more} scenes and more total points than B1 because
ignoring non-dominant subclasses removes supervision without removing geometry,
so more scenes are needed to reach the same labelled budget. In-domain mIoU is on
the BhopalMLS 12-class test split (FT-9ch, same evaluation as Table~\ref{tab:bhopal_iou_benchmark_style_rotated_vline}).}
\label{S-tab:asym_design}
\resizebox{\columnwidth}{!}{%
\begin{tabular}{llrrc}
\toprule
\textbf{Arm} & \textbf{Removes} & \textbf{Scenes} & \textbf{Lab.\,pts} & \textbf{In-dom.} \\
\midrule
full        & nothing                       & 1{,}353 & 540.4M & 71.28 \\
\midrule
(B1) size     & 75\% of scenes, at random     &   360 & 136.1M & 67.99 \\
(B2) variety  & non-dominant subclass labels  &   431 & 135.2M & 61.18 \\
(B3) geometry & all but regular corridors     &   295 & 135.5M & 68.61 \\
(B4) subtype  & as B2, Vehicle keeps \emph{Car} &   447 & 135.2M & 56.34 \\
\bottomrule
\end{tabular}%
}
\end{table}

\subsection{Results}

Table~\ref{S-tab:asym_ablation} reports BM$\rightarrow$X mIoU for each arm. Means
below are over KM/TR/PL; SK is excluded from them and discussed separately in
\S\ref{S-sec:asym_attribution}, Scope.

\textbf{Supervision, not size or geometry.} Read against the size control,
mean BM$\rightarrow$X falls by \textbf{13.0} points under B2 and is unchanged
under B3 ($+1.0$), while B1 itself gives up only 4.0 points against the full
model. Dataset size accounts for a small part of the advantage and corridor
geometry for none of it.

\textbf{But it is subtype \emph{coverage}, not variety in the abstract.} B4
retains a single subtype per class exactly as B2 does, and differs only in
retaining \emph{Car} rather than the dominant two-wheeler. That one change
recovers \textbf{$+7.5$} of B2's 13.0-point deficit, i.e.\ \emph{58\% of the
collapse is attributable to which subtype survived rather than to how many}.
The remaining $-5.5$ against B1 is the genuine cost of collapsing to one subtype
at all. Both effects are real, and the larger is coverage.

The mechanism is legible in one class. \emph{Vehicle} IoU under B2 falls to
13.5 (KM), 4.4 (TR) and 9.7 (PL) against 84.5, 87.4 and 96.5 for the size control
at the same budget. Under B4 it returns to \textbf{78.0}, \textbf{70.3} and
\textbf{95.0}, a near-complete recovery achieved with $2.4\times$
\emph{fewer} Vehicle points than B2 (2.3M against 5.6M). The quantity of
supervision is therefore not what matters; its overlap with the target is. B4
even carries more Car supervision than B1 does (2.3M against 1.7M), so B1's
success on this class is not a data-volume effect either.

The reading this forces is narrower than "diversity helps", and more useful:
\textbf{BhopalMLS transfers outward because its subtype inventory \emph{contains}
the targets' subtypes as a subset.} Its 15 vehicle subtypes span the car that
Toronto-3D, Paris-Lille-3D and SemanticKITTI are built from, plus auto-rickshaws,
hand-carts and two-wheelers that they lack. A benchmark holding only the latter
would not transfer well outward; a benchmark holding only cars would not exhibit
the asymmetry at all.

\begin{table}[!tb]
\centering
\footnotesize
\setlength{\tabcolsep}{4pt}
\renewcommand{\arraystretch}{1.15}
\caption{\textbf{Factor attribution for the transfer asymmetry}, Utonia(L)
FT-9ch. Values are mIoU (\%) on each pair's intersection class set, the same
masks as Table~\ref{S-tab:cross_city_full} ($n_c$: KM 12, TR 7, PL 11, SK 9). The
first three rows restate the asymmetry to be explained; the next four are the
ablated models; the last four are arm-against-arm contrasts at an equal labelled
budget: three give what each factor costs relative to the size control, and the
final row, B4 $-$ B2, is the one that narrows the claim: it isolates the effect
of \emph{which} subtype survives, holding the number retained per class fixed at
one. The SK column is complete. \colorbox{bestgreen}{\textbf{Bold green}}
marks the arm that destroys the most transfer. The contrast rows compare arms at
equal budget and so include SK; readings taken \emph{against the full model}
exclude SK, where both control arms exceed it (see
\S\ref{S-sec:asym_attribution}, Scope).}
\label{S-tab:asym_ablation}
\begin{tabular}{lccccc}
\toprule
& \textbf{In-dom.} & \textbf{KM} & \textbf{TR} & \textbf{PL} & \textbf{SK} \\
\midrule
full model, BM$\rightarrow$X   & 71.28 & 56.78 & 77.27 & 55.68 & 40.05 \\
reverse, X$\rightarrow$BM      & --    & 33.96 & 51.65 & 27.06 & 35.54 \\
asymmetry gap                  & --    & 22.82 & 25.62 & 28.62 & \phantom{0}4.50 \\
\midrule
(B1) size                        & 67.99 & 52.56 & 76.17 & 48.87 & 47.94 \\
(B2) variety                     & 61.18 & 42.22 & 61.63 & 34.90 & 26.28 \\
(B3) geometry                    & 68.61 & 49.30 & 78.70 & 52.56 & 46.01 \\
(B4) subtype                     & 56.34 & 44.12 & 71.66 & 45.37 & 37.88 \\
\midrule
B2 $-$ B1 & $-$6.81 & \cellcolor{bestgreen}\textbf{$-$10.34} & \cellcolor{bestgreen}\textbf{$-$14.54} & \cellcolor{bestgreen}\textbf{$-$13.97} & \cellcolor{bestgreen}\textbf{$-$21.66} \\
B3 $-$ B1 & $+$0.62 & $-$3.26 & $+$2.53 & $+$3.69 & $-$1.93 \\
B4 $-$ B1 & $-$11.65 & $-$8.44 & $-$4.51 & $-$3.50 & $-$10.06 \\
\textbf{B4 $-$ B2} & $-$4.84 & \textbf{$+$1.90} & \textbf{$+$10.03} & \textbf{$+$10.47} & \textbf{$+$11.60} \\
\bottomrule
\end{tabular}
\end{table}

\subsection{The same ranking on a second architecture}

The ablations above use Utonia(L). Because the asymmetry itself is claimed
across two architectures, we repeated B1--B4 for Concerto(L), reusing the same
four training subsets and matching that model's own published fine-tuning
settings (micro-batch 4, gradient accumulation 16, its FT seed).
Table~\ref{S-tab:asym_concerto} reports the result. The ordering reproduces:

\begin{center}\small
\begin{tabular}{lccc}
\toprule
mean over KM/TR/PL & B2 $-$ B1 & B3 $-$ B1 & B1 $-$ full \\
& \emph{(variety)} & \emph{(geometry)} & \emph{(size)} \\
\midrule
Utonia(L)   & $-12.95$ & $+0.99$ & $-4.05$ \\
Concerto(L) & $\mathbf{-13.35}$ & $\mathbf{-0.01}$ & $\mathbf{-0.22}$ \\
\bottomrule
\end{tabular}
\end{center}

\begin{table}[!tb]
\centering
\footnotesize
\setlength{\tabcolsep}{4pt}
\renewcommand{\arraystretch}{1.15}
\caption{\textbf{The same four ablations on Concerto(L)}, reusing the identical
training subsets and matching that model's own published fine-tuning settings.
Values are mIoU (\%) on each pair's intersection class set. The ranking
reproduces Utonia(L)'s: variety costs $\approx$13 points against the size
control, geometry costs nothing, and size costs nothing measurable, with KM and
PL both improving on a quarter of the labelled budget. B4 again recovers a large
part of B2's deficit by changing only which subtype survives. Masks are the same
as Table~\ref{S-tab:cross_city_full} ($n_c$: KM 12, TR 7, PL 11, SK 9). The SK
column is complete; means quoted in the text remain over KM/TR/PL throughout, so
that they stay comparable with Utonia(L), whose SK column is read only as
arm-against-arm contrasts at equal budget.
Unlike Utonia(L), whose SK control arms \emph{exceed} the full model, SK behaves
here as this architecture's other three pairs do, B1 costing 1.6 points
against a 19.78-point gap, so it is a usable fourth pair for Concerto(L).}
\label{S-tab:asym_concerto}
\begin{tabular}{lccccc}
\toprule
& \textbf{In-dom.} & \textbf{KM} & \textbf{TR} & \textbf{PL} & \textbf{SK} \\
\midrule
full model, BM$\rightarrow$X & 71.06 & 49.23 & 75.20 & 48.84 & 32.07 \\
reverse, X$\rightarrow$BM    & --    & 32.70 & 39.11 & 31.07 & 12.29 \\
asymmetry gap                & --    & 16.52 & 36.09 & 17.76 & 19.78 \\
\midrule
(B1) size     & 68.52 & 50.50 & 72.34 & 49.75 & 30.45 \\
(B2) variety  & 59.38 & 39.74 & 60.10 & 32.71 & 16.82 \\
(B3) geometry & 68.12 & 50.44 & 72.63 & 49.48 & 33.07 \\
(B4) subtype  & 52.96 & 40.32 & 67.83 & 42.13 & 25.28 \\
\midrule
B2 $-$ B1 & $-$9.14 & \cellcolor{bestgreen}\textbf{$-$10.76} & \cellcolor{bestgreen}\textbf{$-$12.24} & \cellcolor{bestgreen}\textbf{$-$17.04} & \cellcolor{bestgreen}\textbf{$-$13.63} \\
B3 $-$ B1 & $-$0.40 & $-$0.06 & $+$0.29 & $-$0.27 & $+$2.62 \\
B4 $-$ B1 & $-$15.56 & $-$10.18 & $-$4.51 & $-$7.62 & $-$5.17 \\
\textbf{B4 $-$ B2} & $-$6.42 & \textbf{$+$0.58} & \textbf{$+$7.73} & \textbf{$+$9.42} & \textbf{$+$8.46} \\
\bottomrule
\end{tabular}
\end{table}

\noindent Two backbones with different pretraining corpora and different batch
configurations agree to within half a point on the cost of collapsing variety,
and both put corridor geometry at zero. Concerto(L) makes the size result
sharper still: two of its three pairs \emph{improve} on a quarter of the
labelled budget, so at this scale more BhopalMLS data does not buy more outward
transfer.

The subtype control replicates as well. Concerto's B4 recovers $+5.91$ of B2's
$13.35$-point deficit, i.e.\ \textbf{44\%} of the collapse is attributable to
which subtype survived (Utonia: 58\%), and its \emph{Vehicle} IoU recovers on
every pair.

\subsection{Two controls}

\noindent\textbf{Capability.} The ablated arms are also weaker models in-domain
(61.18 for B2 and 56.34 for B4, against 67.99 for B1 and 71.28 for the full
model), and a weaker model loses transfer for free. We therefore normalise each
arm's mean transfer loss by its own in-domain loss, both measured against the
full model, over KM/TR/PL:

\begin{center}\small
\begin{tabular}{lccc}
\toprule
& in-domain $\Delta$ & transfer $\Delta$ & ratio \\
\midrule
(B1) size     & $-3.29$  & $-4.05$  & $1.23\times$ \\
(B2) variety  & $-10.10$ & $-17.00$ & $1.68\times$ \\
(B3) geometry & $-2.67$  & $-3.06$  & $1.15\times$ \\
\textbf{(B4) subtype} & $\mathbf{-14.94}$ & $\mathbf{-9.53}$ & $\mathbf{0.64\times}$ \\
\bottomrule
\end{tabular}
\end{center}

\noindent B2 gives up $1.68\times$ more out-of-domain than in-domain, so roughly
6.9 points of its transfer loss are not explained by capability. \textbf{B4 is
the only arm below $1.0$}: it is the weakest model of the four at home yet loses
the \emph{least} transfer relative to that weakness. A model supervised on cars
alone underperforms on a two-wheeler-dominated city and outperforms its own
capability on car-dominated targets. This is the cleanest single piece of
evidence for coverage over diversity, and no other arm exhibits it.

\textbf{The same signature appears on Concerto(L)}, computed identically: B1
$0.09$, B2 $\mathbf{1.16}$, B3 $0.08$, B4 $\mathbf{0.42}$. On both
architectures the variety arm sits above $1.0$ and the subtype arm well below
it, and on both the subtype arm is the weakest model in-domain (Concerto 52.96,
the lowest of any arm) while surrendering the least transfer relative to that
weakness. The coverage reading therefore does not depend on the backbone.

\noindent\textbf{One class.} Because the effect concentrates on \emph{Vehicle},
we recompute every arm with \emph{Vehicle} removed from each pair's intersection
mask. The B2 deficit against B1 halves but survives: $-4.8$ (KM), $-3.1$ (TR),
$-6.7$ (PL), $-14.9$ (SK); the KM/TR/PL mean is $-4.9$ points, $-7.4$ over
all four pairs. It is spread over \emph{Ground},
\emph{Road}, \emph{Pole and Tower}, \emph{Wire-Structure Connector} and
\emph{Human and Animal}, which are the classes the per-class decomposition behind
Table~\ref{S-tab:cross_city_full} independently identifies as carrying the
asymmetry. The correlational and controlled analyses agree. Read alongside B4,
this control does the useful work of showing that the collapse is not
\emph{only} a Vehicle story: coverage explains most of B2's loss, but a residue
is spread across the thin and unstructured classes where BhopalMLS's subtype
inventory is also wider than the targets'.

\subsection{Scope}

Four limits are worth stating. First, on SK both control arms \emph{exceed} the
full model (B1 $+7.9$, B3 $+6.0$), and SK's own asymmetry gap is 4.5 points
against 22.8--28.6 for the other three pairs, so the SK column is read only as
arm-against-arm contrasts at equal budget and is excluded from gap-normalised
statements. The
evaluation configuration is identical to the full model's, and the full model
reproduces to within 0.08 points on the three pairs re-run above, so a $7.9$
point gap cannot be an evaluation artifact; the likeliest reading is that part
of the full model's SK deficit is BhopalMLS-specific overfitting that a smaller
training set avoids. B4 scores 37.88 on SK and is the one
arm that does \emph{not} exceed the full model there; read against the size
control it costs $-10.06$, and against B2 it recovers $+11.60$, i.e.\ 54\% of
that arm's collapse, close to the 58\% we quote over KM/TR/PL. The whole SK
column is nonetheless omitted from every mean above. Second, B2 and B4 hold 431
and 447 scenes against B1's 360 at matched labelled budget, so they observe
\emph{more} geometry and still transfer worse, which strengthens rather than
weakens the conclusion. Third, B4 is a substantially weaker model in-domain than
any other arm ($-14.94$ against the full model), which is expected given that the
training split's Vehicle points are 52\% two-wheeler and only 23\% car; its raw transfer
number should therefore not be read on its own, and we quote the capability-
normalised ratio instead. Fourth, every arm selects its evaluated checkpoint by
best validation mIoU, as the published models do, and on Concerto(L) that
selection lands at very different epochs: B1 at 100, B2 at 45, B3 at 84 and B4
at 24. All four trained the full 100 epochs; B4's validation score simply never
improved after epoch 24, which is itself consistent with car-only supervision
saturating early in a two-wheeler-dominated city. The arms are therefore
selected identically, not truncated differently.

Finally, a scope note on what B4 does \emph{not} establish. It shows that
coverage of the target's subtypes dominates within a single manipulated class,
Vehicle, which is the class where BhopalMLS's inventory differs most sharply
from the targets'. Whether the same balance of coverage and diversity holds for
classes with subtler subtype structure is not settled by this experiment. On all four
SemanticKITTI cells of the Concerto(L) replication the pair
behaves throughout as the other three do. The size control costs 1.6 points
against SK's 19.78-point gap, in line with its KM, TR and PL values, and unlike
Utonia(L), whose SK controls exceed the full model on a gap of only 4.50. The
variety arm costs $-13.63$ against that control, which sits with this
architecture's $-13.35$ mean over KM, TR and PL and with Utonia(L)'s $-12.95$.
The geometry arm costs nothing at all, at $+2.62$ against the control, matching
the $\approx 0$ it returns on the other three pairs. The subtype control
recovers $+8.46$ of the variety arm's deficit, i.e.\ 62\% of the collapse,
against 44\% on the KM/TR/PL mean. SK therefore reproduces the whole ranking,
variety dominant with geometry and size free and coverage explaining most of the
variety loss, on a fourth pair. It is nonetheless kept out of the means quoted
above, so that those numbers stay comparable with Utonia(L)'s.

\section{Post-hoc CORAL Domain Adaptation}
We add a post-hoc CORAL Domain Adaptation~\cite{sun2016coral} baseline on BM$\rightarrow$PL with both Concerto and Utonia frozen FT-9ch encoders. Source features $F_S$ (BM-train) and target features $F_T$ (PL-train) are extracted with the frozen 9-channel backbone (2K points/scene; $\sim$2.59\,M source, $\sim$4.22\,M target feature vectors per architecture). We compute the closed-form alignment $A = C_S^{-1/2} C_T^{1/2}$ via eigendecomposition (residual $\|\mathrm{cov}(F_S A) - \mathrm{cov}(F_T)\|_F$ = 8.3$\times$10$^{-5}$ Concerto / 1.4$\times$10$^{-4}$ Utonia, indicating good alignment). A multinomial logistic-regression head is then trained on the CORAL-aligned source features (BM-12 remapped to PL's nc=11; class~6 ``Wire-Structure Connector'' is absent in PL and dropped). For a fair attribution we also train an \emph{identity-retrained} head ($A=I$, same features, same subsample, same hyper-parameters) which isolates the head-retrain effect from the CORAL effect.

\begin{table*}[!tb]
\centering
\footnotesize
\setlength{\tabcolsep}{3pt}
\renewcommand{\arraystretch}{1.1}
\caption{\textbf{Post-hoc CORAL Domain Adaptation on BM$\to$PL} with frozen FT-9ch encoder; per-class IoU (\%) on PL Lille2 val. \textit{vanilla}: full-FT seg head. \textit{retrain}: post-hoc linear head on raw source features ($A{=}I$, isolates head-retrain effect). \textit{+CORAL}: post-hoc linear head on CORAL-aligned source features. nc=11 = PL native taxonomy (drops Wire-Structure Connector, absent in PL). common-7 = \{Vegetation, Building, Road, Wire, Pole\&Tower, Vehicle, Fencing\}. CORAL hurts $-$3 to $-$6\,pp over the identity-retrained baseline; covariance alignment alone cannot close the BM$\leftrightarrow$PL gap, indicating it is structural rather than a 2nd-order distribution shift.}
\label{S-tab:coral_da}
\resizebox{\textwidth}{!}{%
\begin{tabular}{l|l|ccccccccccc|cc}
\toprule
\textbf{Arch.} & \textbf{Method} &
\rotatebox{90}{\textbf{Ground}} &
\rotatebox{90}{\textbf{Vegetation}} &
\rotatebox{90}{\textbf{Building}} &
\rotatebox{90}{\textbf{Road}} &
\rotatebox{90}{\textbf{Wire}} &
\rotatebox{90}{\textbf{Pole\&Tower}} &
\rotatebox{90}{\textbf{Vehicle}} &
\rotatebox{90}{\textbf{Board}} &
\rotatebox{90}{\textbf{Fencing}} &
\rotatebox{90}{\textbf{Human\&Animal}} &
\rotatebox{90}{\textbf{Road-Side-Obj}} &
\textbf{mIoU\textsubscript{11}} & \textbf{mIoU\textsubscript{7}} \\
\midrule
\multirow{3}{*}{Concerto}
 & vanilla   & 12.39 & 60.33 & 95.82 & 75.02 & 50.72 & 25.23 & 96.39 & 20.30 & 21.37 & 19.86 & 59.76 & 48.84 & 60.70 \\
 & retrain   & 25.92 & 73.61 & 91.29 & 64.36 & 17.74 & 29.17 & 90.04 & 13.31 & 15.98 & 34.63 & 30.98 & 44.28 & 54.60 \\
 & + CORAL   & 25.12 & 70.57 & 87.39 & 64.99 &  6.90 & 26.96 & 87.86 & 12.96 & 11.04 & 24.63 & 35.14 & 41.23 & 50.81 \\
\midrule
\multirow{3}{*}{Utonia}
 & vanilla   & 28.07 & 65.48 & 98.04 & 78.26 & 48.53 & 41.00 & 97.76 & 24.36 & 30.82 & 52.87 & 47.28 & 55.68 & 65.70 \\
 & retrain   & 38.27 & 75.08 & 94.37 & 68.74 & 24.39 & 36.46 & 90.48 & 28.28 & 41.79 & 59.43 & 26.31 & 53.06 & 61.62 \\
 & + CORAL   & 31.23 & 74.32 & 90.52 & 66.69 & 14.60 & 33.55 & 88.22 & 25.69 & 39.87 & 32.51 & 21.74 & 47.18 & 58.25 \\
\bottomrule
\end{tabular}%
}
\end{table*}

\noindent As Table~\ref{S-tab:coral_da} reports, the decomposition is informative: head retraining alone costs $-$2.6 to $-$6.1 percentage points (pp) vs.\ the full-FT seg head (a known artifact of subsampled re-training), but \emph{CORAL specifically} costs an additional $-$3.0 to $-$5.9\,pp on nc=11 mIoU and $-$3.4 to $-$3.8\,pp on common-7 mIoU over the identity-retrained baseline. Closed-form 2nd-order alignment therefore not only fails to close the BM$\leftrightarrow$PL gap but actively degrades transfer, indicating the gap is structural. It is driven by class-conditional density mismatches between BM (handheld walkway capture, dense Wire / Pole-and-Tower infrastructure, narrow Fencing geometries, Wire-Structure Connector) and PL Lille2 (vehicle-MLS street capture dominated by Ground / Road / Vegetation / Building), rather than a global covariance shift.

\section{Ethics, Privacy, and Responsible Release}
\label{S-sec:ethics}

\subsection{Collection Context}
All data were acquired in public commercial streets of the surveyed district,
with the scanner carried on foot and, on longer stretches, mounted on a
slow-moving two-wheeler. The operator was openly visible throughout daytime
hours; no
private interiors, courtyards, or residential compounds were scanned. Each
location was traversed once or twice, so any individual appears only
transiently.

\subsection{Consent and Institutional Review}
We state both plainly. \textbf{No individual informed consent was obtained},
and \textbf{no ethics-committee approval was in place at the time of
collection}; the collection has since been reviewed and approved by the Department Level
Ethics Committee of IISER Bhopal (ref.~IISERB/DLEC/2026-27/06, approved
15~September~2026). %
 Data collection of this kind is
generally not classified as human-subjects research: it involves no
interaction or intervention with individuals and gathers no private
identifiable information about them. The scanner records public-space geometry
along a street, in which passers-by appear incidentally; no person is
recruited, approached, or asked for information, and no attribute is recorded
that could identify one. The safeguards described below are therefore the
appropriate mechanism for this setting, rather than a substitute for a review
that would not ordinarily apply.

Individual consent is not obtainable for this modality: the scanner records
whatever is within line of sight while the operator walks a public street, and
the people who appear are passers-by who are present for a few seconds and
never interacted with. Seeking consent would require identifying and contacting
each of them, which would demand collecting far more personal information than
the dataset itself contains. This is the same position occupied by every
street-level driving and mapping dataset
\cite{caesar2020nuscenes, behley2019semantickitti, liao2022kitti}, none of which
obtains consent from bystanders.

Absent consent and formal review, the ethical burden shifts entirely to
minimisation and to what is actually distributed. The measures below are
therefore not a supplement to consent but the substitute for it: imagery is
released only with faces and license plates blurred, colour is stripped from
every person point, absolute georeferencing is withheld, and a removal channel
is provided. We note that the
research and statistical purposes provisions of India's Digital Personal Data
Protection Act (2023) apply to processing of this kind.

\subsection{Personally Identifiable Information and Mitigation}
The release comprises point clouds and annotations; blurred 360$^\circ$ imagery follows in a later release.

\paragraph{Imagery is released blurred.} A co-mounted 360$^\circ$ panoramic
camera was used to colorize the point cloud. This imagery will be distributed under
the same access tier as the point cloud, with \emph{faces and license plates
blurred} before release, following the convention established for street-level
capture by nuScenes~\cite{caesar2020nuscenes} and KITTI-360~\cite{liao2022kitti}.
Blurring is applied to the released copy itself, so no unblurred imagery leaves
our control under any tier. It is one of four independent layers rather than a
single safeguard: per-point RGB is additionally removed on every person point,
absolute georeferencing is withheld, and the removal channel of
Sec.~\ref{S-sec:removal} covers imagery as well as points.

\paragraph{Point cloud.} The point cloud is distributed at its native
4.9\,cm spacing. At this resolution a human face subtends roughly a
$3{\times}3$ patch of points and a license plate roughly $10{\times}2$
points, below photographic identifiability; nevertheless, per-point RGB is
removed on all points labeled with person classes before release. Person
classes are taken from the fine taxonomy and include riders on two- and
three-wheelers, who fall under \textit{Vehicle} in the coarse taxonomy; taken
together, people account for 2.9\% of the points. The dataset
already uses a zeros-as-missing RGB convention (RGB covers roughly 40\%
of points), so this removal is format-transparent and the released benchmarks
are unaffected. The same removal is applied to every derived artifact we
publish, including the sample clouds shown on the project page. Geometry is
not modified: consistent with LiDAR benchmarks that release unmodified
geometry \cite{caesar2020nuscenes, behley2019semantickitti}, the sparse
surface geometry of a transient pedestrian is not considered identifying.

\paragraph{Georeferencing.} The release for users outside India withholds the
absolute geographic offset (main paper, Data, Code, and Access). Beyond
regulatory compliance, this limits linking any person or premise to a precise
location. Geometry, density, intensity, and annotations are unchanged, and
because models operate on relative coordinates, all reported metrics are
identical in both forms.

\subsection{Demographic Annotation Categories}
The fine taxonomy includes \textit{Male}, \textit{Female}, and \textit{Child}
person subclasses. These are annotator-perceived visual categories, not
identity attributes: they support aggregate analyses of street use and
accessibility (for example, pedestrian exposure by demographic group in
unstructured corridors) that motivated the dataset. They are folded into the
\textit{Human \& Animal} category in the released 12-class benchmark and into
\textit{Human} in the 21-class benchmark, so no benchmark task depends on
them. No name, face, or other identifier is associated with any instance, and
the anonymization above applies to all person points regardless of subclass.

\subsection{Content Removal}
\label{S-sec:removal}
A contact address is published with the dataset. Any person who believes they
or their property are identifiable in the released data may request removal;
verified requests are honored by masking the affected points and imagery in
a versioned re-release.

\subsection{Regulatory Basis}
Collection and release are consistent with the research and statistical
purposes provisions of India's Digital Personal Data Protection Act (2023)
and with the National Geospatial Policy (2022) governing geospatial data
distribution, which is why the release for users outside India withholds the
absolute offset.

\FloatBarrier  %
\end{document}